\PassOptionsToPackage{table}{xcolor}
\documentclass[11pt, a4paper, logo, copyright, nonumbering]{mll_style}

\usepackage{color}
\usepackage{epsfig}
\usepackage{graphicx}

\usepackage{booktabs}  %
\usepackage{tabularx}               %
\usepackage{ltablex}

\newcolumntype{C}{>{\centering\arraybackslash}X}
\usepackage{multirow}               %
\usepackage{diagbox}                %
\usepackage{hhline}                 %
\usepackage{color}                  %

\usepackage{booktabs}
\usepackage{extarrows}
\usepackage{makecell}
\usepackage{wrapfig}
\usepackage{colortbl}
\usepackage{longtable}
\usepackage[most]{tcolorbox}
\tcbuselibrary{skins} %
\usepackage{tcolorbox}

\usepackage{amssymb}

\usepackage{adjustbox}
\usepackage{array}
\usepackage{floatflt}

\usepackage{bm}
\usepackage{nicefrac}
\usepackage{microtype}

\usepackage{changepage}
\usepackage{extramarks}
\usepackage{fancyhdr}
\usepackage{lastpage}
\usepackage{setspace}
\usepackage{soul}
\usepackage{xspace}

\usepackage{arydshln} %

\usepackage{enumerate}
\usepackage{enumitem}  %

\usepackage{makecell}

\usepackage{pifont} %

\usepackage{algpseudocode}

\usepackage[symbol]{footmisc}

\usepackage{caption}
\usepackage{subcaption}

\usepackage{scalefnt}

\usepackage{fontawesome5}

\usepackage{siunitx}

\usepackage{listings}

\usepackage[ruled,vlined]{algorithm2e} %
\usepackage{wrapfig,float}

\newcolumntype{L}[1]{>{\raggedright\let\newline\\\arraybackslash\hspace{0pt}}m{#1}}
\newcolumntype{R}[1]{>{\raggedleft\let\newline\\\arraybackslash\hspace{0pt}}m{#1}}

\newcommand{\ignore}[1]{}

\makeatletter
\DeclareRobustCommand\onedot{\futurelet\@let@token\@onedot}
\def\@onedot{\ifx\@let@token.\else.\null\fi\xspace}

\makeatother

\definecolor{MyBlue}{rgb}{0.46, 0.50, 0.61}
\definecolor{MyDarkBlue}{rgb}{0,0.08,0.8}
\definecolor{MyDarkGreen}{RGB}{45,155,45}
\definecolor{MyDarkRed}{rgb}{0.8,0.02,0.02}
\definecolor{MyOrange}{rgb}{1.0, 0.4, 0.2}
\definecolor{MyPurple}{RGB}{111,0,255}
\definecolor{MyRed}{rgb}{0.8,0.0,0.0}
\definecolor{MyGold}{rgb}{0.75,0.6,0.12}
\definecolor{MyDarkgray}{rgb}{0.66, 0.66, 0.66}
\definecolor{MyBrown}{rgb}{0.65, 0.16, 0.16}
\definecolor{MyMutedRose}{rgb}{0.58, 0.29, 0.35}
\definecolor{JiayuanColor}{rgb}{0.60,0.43,0.48}
\definecolor{erranColor}{rgb}{24, 40, 113}

\definecolor{citecolor}{HTML}{696FAD}

\DeclareRobustCommand{\name}{\textls[-15]{\textsc{ENACT}}\xspace}

\newif\ifpropositionfirstitem
\propositionfirstitemtrue

\definecolor{bggray}{HTML}{F5F5F5}
\definecolor{pvdblue}{HTML}{DAE8FC}
\definecolor{RoseQuartzBg}{HTML}{F7CAC9}
\definecolor{RoseQuartz}{HTML}{F5A798}
\definecolor{Serenity}{HTML}{92A8D1}
\definecolor{OrangeRed}{rgb}{1.0, 0.27, 0.0}
\definecolor{RoyalBlue}{cmyk}{1, 0.50, 0, 0}
\definecolor{Turquoise}{HTML}{0F4C81}
\definecolor{mint}{rgb}{0.24, 0.71, 0.54}
\definecolor{green}{rgb}{0.0, 0.120, 0.0}

\newdimen\abovecrulesep
\newdimen\belowcrulesep
\makeatletter
\patchcmd{\@@@cmidrule}{\aboverulesep}{\abovecrulesep}{}{}
\patchcmd{\@xcmidrule}{\belowrulesep}{\belowcrulesep}{}{}
\makeatother

\definecolor{mybluetitle}{HTML}{4B527E} %

\definecolor{mygreen}{RGB}{0,150,0}
\definecolor{boxbackground}{HTML}{F0F7FF}  %
\definecolor{boxborder}{HTML}{D0D9E5}      %
\definecolor{accentblue}{HTML}{4A86E8}     %
\definecolor{lightblue}{HTML}{EEF3FF}  %
\definecolor{bordergray}{HTML}{CCCCCC}  %
\definecolor{headerblue}{HTML}{2C5AA0}  %

\definecolor{lavenderframe}{HTML}{E6E6FA}  %
\definecolor{lighterlav}{HTML}{F5F5FF}  %
\definecolor{codegray}{rgb}{0.5,0.5,0.5}  %
\definecolor{codepurple}{HTML}{483D8B}  %
\definecolor{backcolour}{HTML}{F5F5FF}  %
\lstdefinestyle{mystyle}{
    backgroundcolor=\color{backcolour},
    commentstyle=\color{headerblue},
    keywordstyle=\color{codepurple},
    numberstyle=\tiny\color{codegray},
    stringstyle=\color{codepurple},
    basicstyle=\ttfamily\scriptsize,
    breakatwhitespace=false,
    breaklines=true,
    captionpos=b,
    keepspaces=true,
    frame=none,
    numbersep=5pt,
    showspaces=false,
    showstringspaces=false,
    showtabs=false,
    tabsize=2
}

\definecolor{jsonkey}{RGB}{44, 130, 201}     %
\definecolor{jsonstring}{RGB}{255, 140, 0}   %
\definecolor{jsonnumber}{RGB}{34, 139, 34}   %

\lstdefinelanguage{json}{
    basicstyle=\ttfamily\small,
    numbers=left,
    numberstyle=\tiny\color{gray},
    stepnumber=1,
    numbersep=5pt,
    showstringspaces=false,
    breaklines=true,
    frame=none,
    backgroundcolor=\color{gray!5},
    literate=
     *{:}{{{\color{jsonkey}:}}}{1}
      {,}{{{\color{jsonkey},}}}{1}
      {"}{{{\color{jsonstring}"}}}{1}
      {[}{{{\color{jsonkey}[}}}{1}
      {]}{{{\color{jsonkey}]}}}{1}
      {0}{{{\color{jsonnumber}0}}}{1}
      {1}{{{\color{jsonnumber}1}}}{1}
      {2}{{{\color{jsonnumber}2}}}{1}
      {3}{{{\color{jsonnumber}3}}}{1}
      {4}{{{\color{jsonnumber}4}}}{1}
      {5}{{{\color{jsonnumber}5}}}{1}
      {6}{{{\color{jsonnumber}6}}}{1}
      {7}{{{\color{jsonnumber}7}}}{1}
      {8}{{{\color{jsonnumber}8}}}{1}
      {9}{{{\color{jsonnumber}9}}}{1}
}

\newtcblisting{jsonbox}{
  listing engine=listings,
  colback=gray!3!white,
  colframe=gray!75!black,
  boxrule=0.4mm,
  arc=2mm,
  outer arc=2mm,
  breakable,
  enhanced,
  listing only,
  listing options={language=json}
}

\newtcolorbox{promptbox}[2][]{ %
    enhanced,
    breakable,
    boxsep=5pt,
    left=9pt,
    right=7pt,
    top=5pt,
    bottom=5pt,
    colback=boxbackground,
    colframe=boxborder,
    boxrule=0.5pt,
    arc=4pt,
    frame hidden, %
    borderline west={3pt}{0pt}{accentblue},
    shadow={0.5pt}{0.5pt}{1.5pt}{black!10},
    fontupper=\normalsize,
    title=#2, %
    colbacktitle=accentblue, %
    coltitle=white,         %
    fonttitle={\fontsize{9}{11}\selectfont\bfseries}, %
    attach boxed title to top left={yshift=-2.5mm, xshift=3.2mm},
    boxed title style={
        enhanced,
        left=3pt,
        right=3pt,
        top=1pt,    %
        bottom=1pt, %
        boxsep=2pt,
        arc=3pt,
        boxrule=0pt,
        colback=accentblue,
    },
    #1 %
}

\newtcolorbox{notitlepromptbox}[1][]{
    enhanced,
    breakable,
    boxsep=5pt,          %
    left=9pt,            %
    right=7pt,           %
    top=5pt,             %
    bottom=5pt,          %
    colback=boxbackground,
    colframe=boxborder,
    boxrule=0.5pt,
    arc=4pt,             %
    frame hidden,
    borderline west={3pt}{0pt}{accentblue},  %
    shadow={0.5pt}{0.5pt}{1.5pt}{black!10},  %
    notitle,
    fontupper=\normalsize,    %
    #1
}

\newtcolorbox{onebox}[2][]{
    enhanced, 
    center title,
    left*=0pt, right*=0pt,
    boxsep=2pt, left=5pt, right=5pt,
    skin first=enhanced,
    skin middle=enhanced,
    skin last=enhanced,
    colframe = mybluetitle!90,
  colback  = mybluetitle!10,
    fonttitle=\bfseries\rmfamily\fontfamily{phv}\selectfont,
    title={\footnotesize\strut{#2}  \refstepcounter{subsubsection} \addcontentsline{toc}{subsubsection}{\string\numberline{\thesubsubsection}#2}
    },
    #1
    }

\colorlet{osfirst}{teal!50}
\colorlet{ossecond}{teal!30}
\colorlet{osthird}{teal!10}
\colorlet{lavenderfirst}{violet!50}
\colorlet{lavendersecond}{violet!30}
\colorlet{lavenderthird}{violet!10}

\usepackage{natbib}
\usepackage{graphicx}
\usepackage{xspace}
\usepackage{adjustbox}
\usepackage{array}
\usepackage{float}
\usepackage{geometry}
\usepackage{enumitem}
\usepackage{setspace}
\usepackage{booktabs}
\usepackage{multirow}
\usepackage{longtable}
\usepackage{tabularx}
\usepackage{xltabular}
\usepackage{threeparttable}
\usepackage{siunitx}

\usepackage{amsmath}
\usepackage{amssymb}
\usepackage{amsfonts}
\usepackage{mathtools}
\usepackage{bm}
\usepackage{dsfont}

\usepackage{subcaption}
\usepackage{tikz}
\usepackage{pgfplots}
\pgfplotsset{compat=1.16}
\usepackage{listings}

\usepackage{cleveref}
\hypersetup{colorlinks=true}

\usepackage{enumitem}
\usepackage{fontawesome5}
\usepackage[most]{tcolorbox}
\usepackage{CJKutf8}
\usepackage{lipsum}

\usepackage[toc,page,header]{appendix}
\usepackage{minitoc}

\usepackage{pifont}
\usepackage{tcolorbox}
\usepackage{listings} %
\usepackage{caption} 
\usepackage{lipsum} %
\usepackage{fontawesome5}

\renewcommand \partname{}

\definecolor{tableblue}{RGB}{201,226,239}

\makeatletter
\def\@BTrule[#1]{%
  \ifx\longtable\undefined
    \let\@BTswitch\@BTnormal
  \else\ifx\hline\LT@hline
    \nobreak
    \let\@BTswitch\@BLTrule
  \else
     \let\@BTswitch\@BTnormal
  \fi\fi
  \global\@thisrulewidth=#1\relax
  \ifnum\@thisruleclass=\tw@\vskip\@aboverulesep\else
  \ifnum\@lastruleclass=\z@\vskip\@aboverulesep\else
  \ifnum\@lastruleclass=\@ne\vskip\doublerulesep\fi\fi\fi
  \@BTswitch}
\makeatother

\addto\extrasenglish{
}

 {\begin{list}{}%
         {\setlength{\leftmargin}{#1}}%
         \item[]%
 }
 {\end{list}}

\reportnumber{001} %

\definecolor{firstgrey}{gray}{0.66} %
\definecolor{secondgrey}{gray}{0.88} %

\title{\centering ENACT: Evaluating Embodied Cognition\\with World Modeling of Egocentric Interaction}

\date{}

\author{
Qineng Wang*$^{1}$, Wenlong Huang*$^{2}$, Yu Zhou$^{3}$, Hang Yin$^{2}$, Tianwei Bao$^{1}$, Jianwen Lyu$^{1}$, \quad
Weiyu Liu$^{2}$, Ruohan Zhang$^{\dag2}$, Jiajun Wu$^{\dag2}$, Li Fei-Fei$^{\dag2}$, Manling Li$^{\dag1}$
\\
\small $^1$Northwestern University~~~$^2$Stanford University~~~$^3$UCLA
\\
\footnotesize \emph{* Equal contribution; $\dag$ Equal advising.}
\\
{
    \footnotesize 
    \href{https://enact-embodied-cognition.github.io/}{\faGlobe~Website}
    \quad 
    \href{https://github.com/mll-lab-nu/ENACT}{\faGithub~Code}
    \quad
    \href{https://huggingface.co/datasets/MLL-Lab/ENACT}{\protect\includegraphics[height=1.05em]{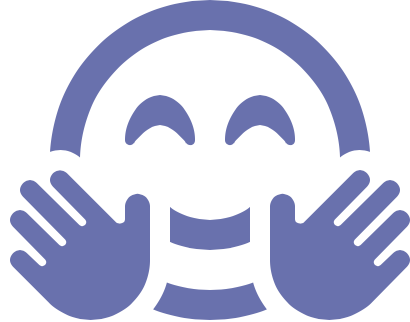}~Dataset}
}
\vspace{-10pt}
}

\begin{document}
\doparttoc %
\faketableofcontents %
\begin{abstract}
Embodied cognition argues that intelligence arises from sensorimotor interaction rather than passive observation.
It raises an intriguing question: do modern vision-language models (VLMs), trained largely in a disembodied manner, exhibit signs of embodied cognition?
We introduce \textbf{\name}, a benchmark that casts evaluation of embodied cognition as \textbf{world modeling from egocentric interaction} in a visual question answering (VQA) format.
Framed as a partially observable Markov decision process (POMDP) whose actions are scene graph changes, \name comprises two complementary sequence reordering tasks: forward world modeling (reorder shuffled observations given actions) and inverse world modeling (reorder shuffled actions given observations).
While conceptually simple, solving these tasks implicitly demands capabilities central to embodied cognition—affordance recognition, action–effect reasoning, embodied awareness, and interactive, long-horizon memory from partially observable egocentric input—while avoiding low-level image synthesis that could confound the evaluation.
We provide a scalable pipeline that synthesizes QA pairs from robotics simulation (BEHAVIOR) and evaluate models on 8,972 QA pairs spanning long-horizon home-scale activities.
Experiments reveal a performance gap between frontier VLMs and humans that widens with interaction horizon.
Models consistently perform better on inverse task than forward one and exhibit anthropocentric biases, including a preference for right-handed actions and degradation when camera intrinsics or viewpoints deviate from human vision.
Website at \href{https://enact-embodied-cognition.github.io/}{enact-embodied-cognition.github.io}.

\end{abstract}

\maketitle
\vspace{-0.4em}
\enlargethispage{3\baselineskip}
{
    \captionsetup{type=figure,hypcap=false,skip=3pt}
    \begin{center}
        \includegraphics[width=\linewidth]{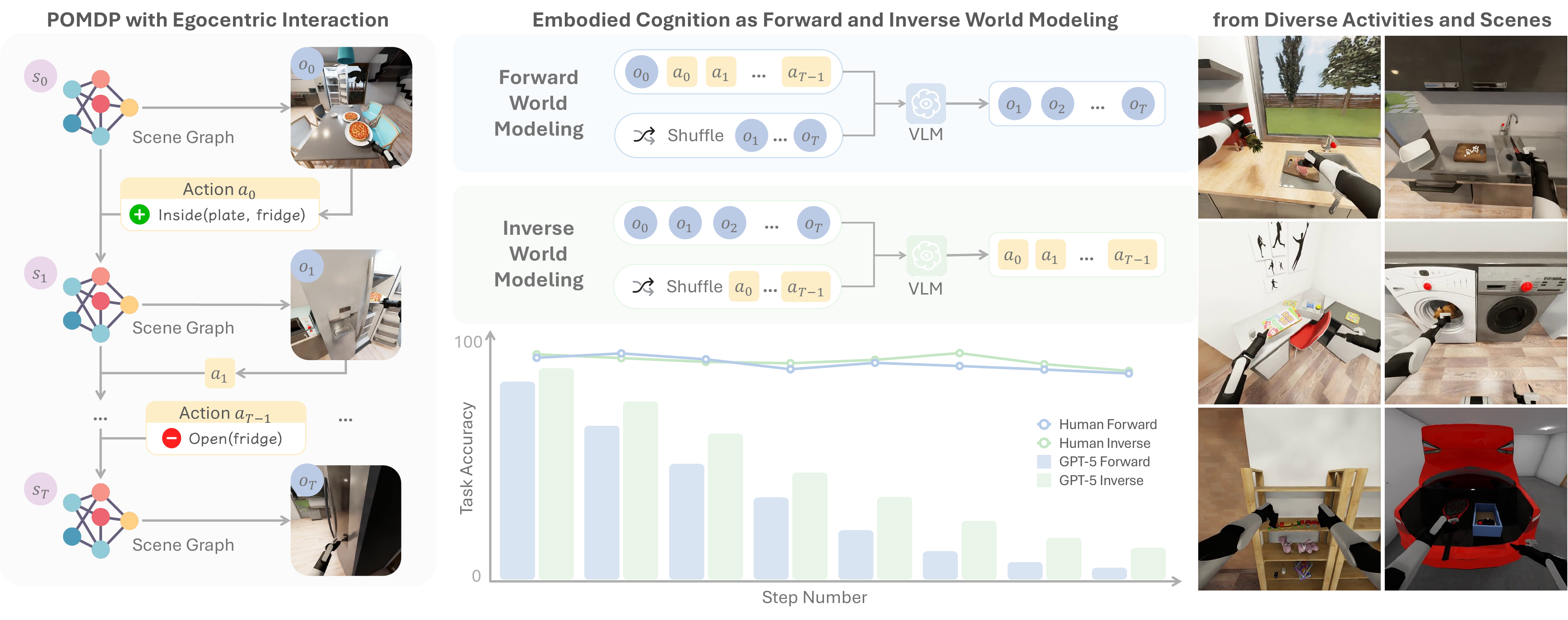}
        \vspace{-1em}
        \captionof{figure}{\small
        \textbf{\name casts embodied cognition evaluation as world modeling through egocentric interaction}.
        Grounded in POMDP framework, \name considers two tasks from diverse activities and scenes: forward world modeling (ordering observations given actions) and inverse world modeling (ordering actions given observations). Evaluation shows that performance of VLMs drops as interaction horizons lengthen, performs better on inverse task, and lags behind humans.}
        \label{fig:teaser}
    \end{center}
}
\vspace{-1.5em}

\section{Introduction}

Intelligent behavior in the physical world requires grounding abstract knowledge in interaction with the environment. Embodied cognition argues that intelligence is not passively acquired but \textit{enacted} through continuous sensorimotor interaction with the world~\citep{smith2005development}. Recent advances in large foundation models such as Vision–Language Models (VLMs)~\citep{openai2025gpt5systemcard, google2025gemini25pro}, although predominantly trained in a disembodied fashion, have exhibited promising signs of interactive intelligence. Yet it remains unclear how to measure the extent to which embodied cognition emerges in these models.

Prior work examines complementary slices of embodied cognition: spatial perception in static scenes~\citep{ramakrishnan2024does}, linguistic reasoning and planning~\citep{li2024embodied}, and reasoning about interactions between primitive objects~\citep{yi2019clevrer, gao2025vision}.
Taxonomies such as \citet{yang2025embodiedbench} attempt to catalogue embodied capabilities for VLMs but ultimately rely on subjective criteria.
Consequently, a unified objective that tightly couples egocentric perception with \emph{embodied} interactions in everyday activities remains missing~\citep{frick2016matter, thompson2005sensorimotor, clark2006language, barsalou2020challenges}.

To this end, we introduce \name, a benchmark that studies embodied cognition through \textbf{world modeling via egocentric interaction} in a visual question answering (VQA) framework.
Grounded in a partially observable Markov decision process \citep[POMDP,][Figure~\ref{fig:teaser}]{aastrom1965optimal},
we formulate world modeling~\citep{ha2018world} as the evolution of egocentric visual observations of the environment conditioned on an agent's actions, which are represented as scene graph~\citep{johnson2017clevr} changes derived from low-level physics-based simulator state.
Specifically, we focus on two forms, forward world modeling and inverse world modeling, and they are formulated as \textit{sequence reordering} for evaluating embodied cognition for a VLM.
This sequence-reordering VQA view isolates long-horizon interactive visual reasoning from photorealistic video prediction and forces models to reason about how sequences of embodied actions transform the environment from egocentric observations.
In \textbf{forward world modeling}, given a visual observation and a sequence of actions, the model must reorder a shuffled sequence of future visual observations.
In \textbf{inverse world modeling}, given an ordered sequence of visual observations, the model must reorder the corresponding shuffled action sequence.
Though seemingly a narrow lens, answering these queries implicitly involves a broad set of capabilities central to embodied cognition: affordance recognition, action–effect reasoning, and embodied awareness, together with reasoning about contact and other low-level physical consequences encoded in our predicates, all from egocentric input.
Under partial observability, it demands integrating observations and actions across extended horizons, posing challenges for interactive and long-term memory.
Through the same lens, we also examine factors that inform future VLM data design, including existing biases toward human embodiment such as right-handedness and human-like egocentric viewpoints and intrinsics (e.g., FOV, aperture).

Leveraging this unified lens, we additionally demonstrate how such evaluation data can be automatically and scalably generated within a robotics simulator, such as BEHAVIOR~\citep{li2024behavior1k}.
Given a robot manipulation trajectory, we extract symbolic scene graphs leveraging ground-truth, physics-based simulation state (e.g., continuous poses and contacts), which is uniquely accessible in simulation compared to the real world.
We then select the key-frames where the scene graph changes and sample subsequences of desired lengths to assemble the state-action sequences for constructing the question-answer (QA) pairs.
By additionally leveraging a novel QA sampling scheme, our pipeline can generate up to millions of QAs from a single episode, enabling massive scaling across the thousands of trajectories common in robotics datasets.
Although we do not train or finetune VLMs here, the resulting data directly supports future embodied decision-making studies~\citep{azzolini2025cosmos}.

We report two metrics at different granularities: Task Accuracy (exact ordering) and Pairwise Accuracy (fraction of adjacent pairs correctly ordered).
Our experiments reveal that \name is challenging for current VLMs, which lag significantly behind human performance (Figure~\ref{fig:teaser}). This performance gap widens as the task horizon increases, where VLM accuracy degrades sharply with trajectory length while human performance remains high. We also find that all models consistently perform better on inverse than on forward world modeling. Furthermore, we uncover two notable biases: VLMs show a clear preference for understanding right-handed dynamics, and the performance of representative models like GPT-5 mini drops significantly with non-human-eye-like camera intrinsics or viewpoints.

Overall, our contributions are threefold: (1) We introduce \name, a benchmark for evaluating embodied cognition via forward and inverse world modeling from egocentric interaction. (2) We provide an easily scalable data generation pipeline leveraging robotics simulation (BEHAVIOR) and provide a dataset of 8,972 QAs synthesized from diverse, long-horizon interactions in everyday environments. (3) Experiments on state-of-the-art VLMs reveal a widening gap to human performance with horizon length, anthropocentric biases (e.g., right-handed priors, human-like camera intrinsics), and real-world evaluations that mirror simulator trends with only a limited sim-to-real gap.

\vspace*{-1em}

\begin{figure}[tbp]
    \centering
    \includegraphics[width=\linewidth]{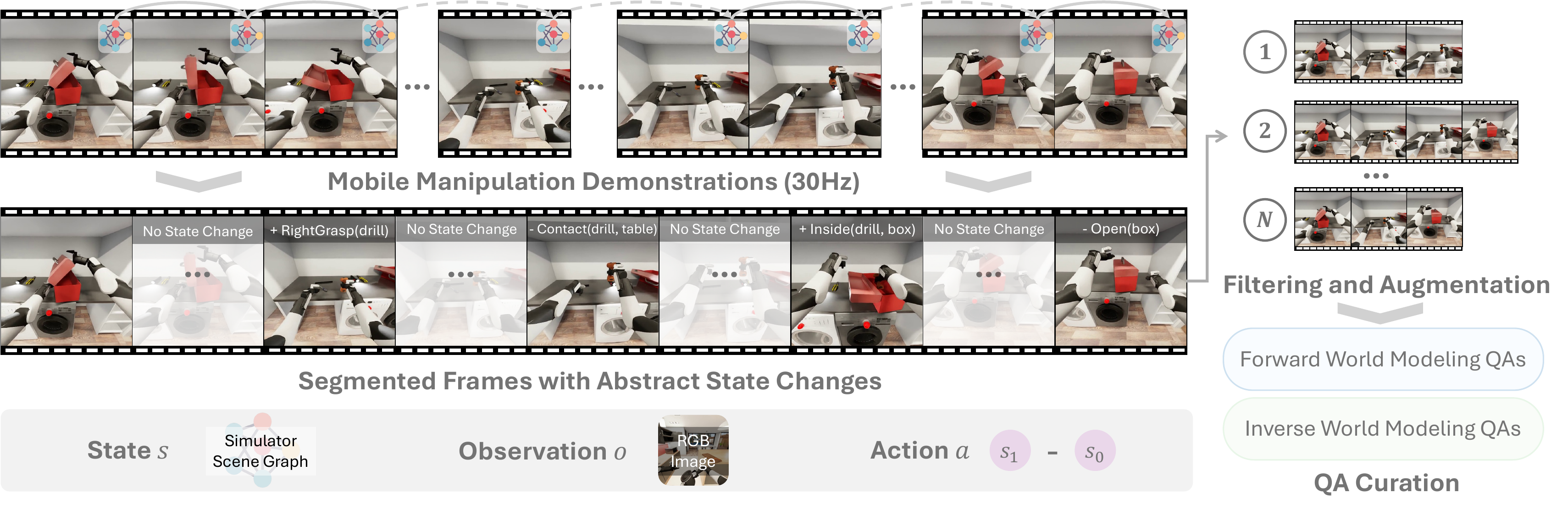}
    \vspace*{-1.5em}
    \caption{\textbf{Overview of \name data curation pipeline.} We first obtain aligned scene graphs (states) and RGB observations from a mobile manipulation dataset in a robotics simulation (BEHAVIOR). The trajectory is then segmented by identifying key-frames where an abstract state change occurs (i.e., the scene graph difference is non-empty). From this set of key-frames, we sample multiple key-frame trajectories, which are used to construct the forward and inverse world modeling VQA questions. Here $N$ refers to the number of all sampled trajectories across all step lengths.}
    \label{fig:method}
    \vspace*{-1em}
\end{figure}
\label{sec:2_1}

\section{\name: Egocentric Interactive Embodied Cognition Test}
\subsection{Problem Formulation}\label{sec:prob_formulate}
We investigate the embodied cognition of VLMs by framing it as a world modeling problem, which we evaluate using egocentric, interactive reasoning tasks.
Formally, the benchmark is defined over a \emph{state space} $\mathcal{S}$, whose elements are symbolic scene graphs derived from low-level simulator states $\mathcal{G}$; an \emph{observation space} $\mathcal{O}\subset\mathbb{R}^{H\times W\times 3}$ of robot's egocentric RGB views of the environment; and an \emph{action space} $\mathcal{A}$ whose elements are scene-graph differences $a_t=\delta(s_t, s_{t-1})$.
A symbolic scene graph is a structured data model that represents the objects in a scene as symbolic nodes (e.g., \texttt{On(fridge)}) and their relationships as edges (e.g., \texttt{OnTop(pen, desk)}).
We view the underlying embodied task as a POMDP.
As shown in Figure~\ref{fig:method}, we first filter this raw data to identify all timestamps where an abstract scene graph state change occurs (i.e., the scene-graph difference $\delta(s_t, s_{t-1}) \neq \varnothing$). This process yields a smaller, chronologically ordered set of segmented frames, which serve as the candidate pool for our benchmark.

From the pool of segmented frames, we sample $R$ trajectories, each with a chronologically ordered tuple $\pi=(i_0, \cdots, i_{L-1})$ of $L$ key frames. This initial abstraction into discrete decision epochs is similar to a semi-MDP~\citep{sutton1999between}. However, we treat each of these final key-frame trajectories as a self-contained POMDP instance with scene graphs $S_\pi$ and observations $O_\pi$. For $k=0,\cdots,L-2$, the action connecting consecutive key frames is the visible scene-graph delta $a_k:=\Delta_\mathrm{Vis}(s_{i_{k+1}}, s_{i_k})$, where $\Delta_\mathrm{Vis}$ returns the subset of differences in $\delta(s_{i_{k+1}}, s_{i_k})$ that are visible in both images. Together, these actions form a discrete symbolic action space $\mathcal{A}$. For notation simplicity, we relabel indices in $\pi$ for each key-frame trajectory to $\pi=(0, \cdots,L-1)$ and $(s_k, o_k) := (s_{i_k}, o_{i_k})$.

Building on these trajectories, we formalize two tasks.
For \textbf{forward world modeling}, given the current image \(o_0\), the correctly ordered action sequence
\((a_0,\ldots,a_{L-2})\),
and a \emph{shuffled list} of observation images \(O'=(o'_1,\ldots,o'_{L-1})\),
the model outputs a permutation \(\sigma\in \mathrm{Sym}([L-1])\) that orders the images to match the actions:
\((o'_{\sigma(1)},\ldots,o'_{\sigma(L-1)})=(o_1,\ldots,o_{L-1})\).
For \textbf{inverse world modeling}, given \(o_0\), the correctly ordered observation images
\((o_1,\ldots,o_{L-1})\), and a \emph{shuffled list} of actions
\(A'=(a'_0,\ldots, a'_{L-2})\), the model outputs a permutation \(\tau\in \mathrm{Sym}([L-1])\) that orders the actions to be consistent with the observation progression:
\((a'_{\tau(1)},\ldots, a'_{\tau(L-1)})=(a_0,\ldots,a_{L-2})\).

\subsection{Key-Frame Trajectories Synthesis for Scalable Data Generation}
\label{sec:2_2}
\textbf{Segmented Frames with Abstract State Changes.}
Raw robot trajectories often contain long stretches with no semantic changes (e.g., gripper motion when opening the toolbox in Figure~\ref{fig:method}).
We mark a timestamp $t$ whenever the simulator state makes a minimal abstract state change, such as transitioning from \emph{not grasping} to \emph{grasping} a drill with the right hand. The BEHAVIOR simulator exposes boolean and relational predicates, where flipping one predicate or updating a relation is our atomic state change. A time $t$ enters the candidate pool if the scene-graph difference $\delta(s_t,s_{t-1})$ is nonempty. To avoid near-duplicate frames, we compare each new change with the last accepted segmented frame: we form a predicate-level \emph{change signature} $c_j$ and keep $t$ only if its cosine similarity with the previous signature $c_{j-1}$is below a threshold. This yields a chronological set of segmented frames $\mathcal{K}=\{t_1<\cdots<t_M\}$ with $(s_{t_i},o_{t_i})$. Thresholds and further details are in the Appendix~\ref{app_1_1:seg_frame}.

\textbf{Key-Frame Trajectories Synthesis.}
From the segmented $M$ frames, we sample length-$L$ key-frame trajectories $\pi=(i_0,\ldots,i_{L-1})$ with $1\!\le\! i_0<\cdots<i_{L-1}\!\le\! M$, so indices do not need to be adjacent. Each candidate is strictly validated: for every $k$, the visible state change $\Delta_{\mathrm{Vis}}(s_{i_{k+1}},s_{i_k})$ is nonempty, and the edited objects are visible in both images, except for object transitioning events (e.g., pineapple being diced), where transient occlusion is permitted. We then treat each valid key-frame trajectory as an individual POMDP instance, with $S_\pi$ and $A_\pi$ as defined in the problem formulation. To further scale data generation, we exploit that typically $L<M$ (in practice $L\!\le\!10$ while $M\!\gtrsim\!30$), and we use skipping to convert trajectory construction into a \textit{“seat selection”} combinatorics problem, choosing $L$ seats out of $M$, which yields at most $\binom{M}{L}$ distinct candidates from a single trajectory. The detailed algorithm is in the Appendix~\ref{app_1_2:key_frame}.

\textbf{World Modeling QA Generation.}
After obtaining the sequence of key-frame trajectories, we formulate the forward and inverse world modeling as \textit{sequence reordering} visual-question answering (VQA) tasks.
This formulation offers two advantages.
First, it yields a clean evaluation signal for long-horizon interactive visual reasoning without conflating performance with photorealistic video prediction.
Second, it requires models to reason about how a \emph{sequence of embodied actions} causally transforms the environment over multiple steps, while maintaining long-horizon spatial memory from purely egocentric observations in large home-scale scenes.

\begin{figure}[tbp]
    \centering
    \includegraphics[width=\linewidth]{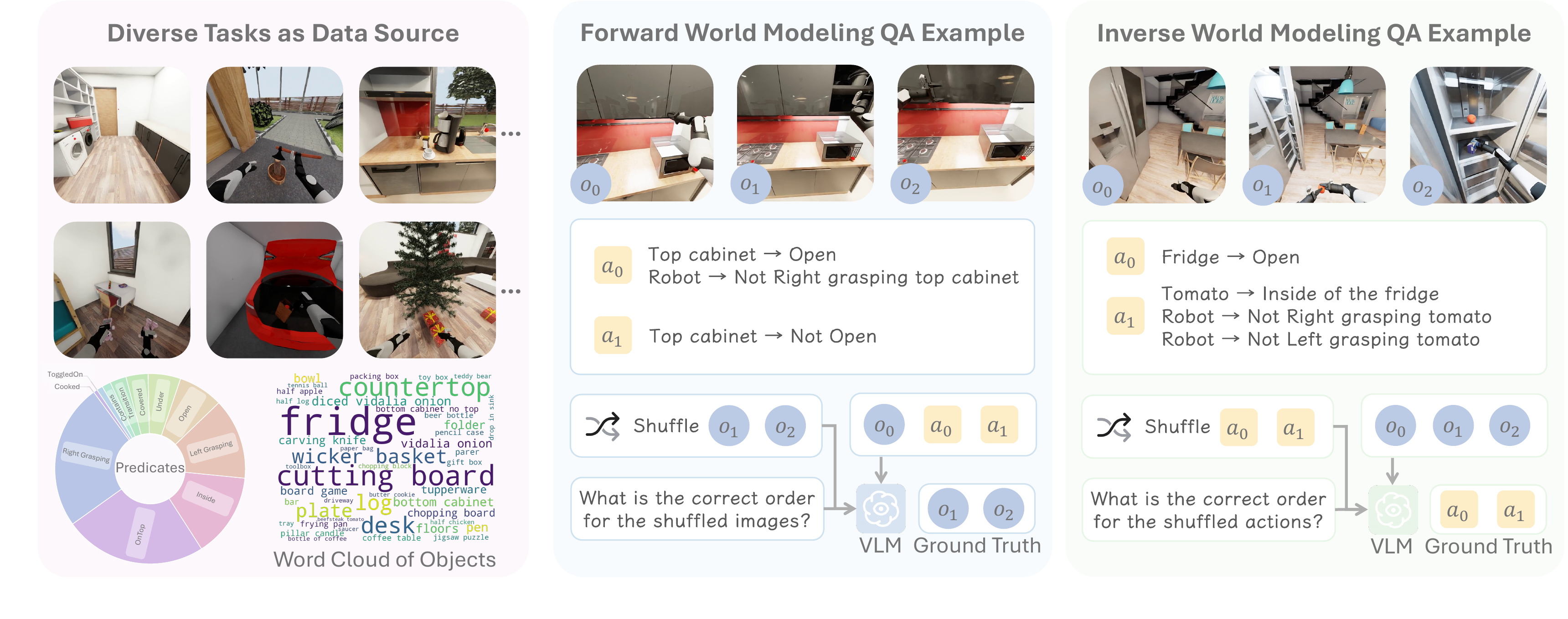}
    \vspace*{-2.5em}
    \caption{\textbf{Data sources and QA examples.} \name is built from diverse, long-horizon activities performed by real robots (left). We provide examples for (mid) forward world modeling and (right) inverse world modeling. More QA examples and prompts are available in the Appendix~\ref{app_2_3:bench_examples}.}
    \vspace*{-1em}
    \label{fig:example}
\end{figure}
\subsection{Dataset Overview and Evaluation Design}\label{sec:2_3}
\textbf{Dataset Overview.}
We construct the benchmark from the BEHAVIOR simulator and challenge~\citep{li2024behavior1k}. BEHAVIOR Challenge provides diverse long-horizon home-scale activities, and we select \(29\) of them, using one trajectory per activity to recover aligned pairs \(\{(s_t,o_t)\}\). Each trajectory is segmented into \emph{segmented frames} \(\mathcal{K}\), then converted into key-frame trajectories and finally into two QA types: forward and inverse world modeling (examples in Figure~\ref{fig:example}). Across step lengths \(L\in\{3,\dots,10\}\) we sample about \(560\) items per \(L\) for each QA type, yielding \(8{,}972\) total questions. The data uses common predicate classes (e.g., \texttt{Open}, \texttt{Cooked}, \texttt{Grasping}), with distributions shown in Figure~\ref{fig:example}; full statistics appear in Appendix~\ref{app_2_1:data_stats}.

\textbf{Evaluation Design.}
Multiple valid answers can exist for a given question, so we use an \textit{online verifier} that accepts any predicted permutation, $\sigma$ or $\tau$, consistent with the input constraints. We report two metrics: \textit{Task accuracy} captures exact ordering, while \textit{Pairwise accuracy} grants partial credit for near-correct sequences. Specifically, (1) \textit{Task accuracy} measures success at the question level: a question receives score \(1\) if the verifier accepts the full prediction and \(0\) otherwise, and the dataset score is \( \mathrm{TA}=(1/|\mathcal{D}|)\sum_{x\in\mathcal{D}}\mathbf{1}\{\text{accepted}(x)\}\). (2) \textit{Pairwise accuracy} measures stepwise consistency: for a question with length \(L\), we count adjacent pairs passing the verifier’s local check (state–action for forward; action–state for inverse) and divide by \(L\), then report the micro-average \( \mathrm{PA}=\big(\sum_x \#\text{correct pairs in }x\big)\big/\big(\sum_x L_x\big) \), equivalent to averaging per-item pairwise scores when \(L\) is fixed. Implementation details are in Appendix~\ref{app_2_2:eval_metrics}.

\definecolor{forwardbg}{HTML}{F0F6FD}
\definecolor{inversebg}{HTML}{F0F6F0}
\definecolor{proprietary}{HTML}{FFF2E6}
\definecolor{openweight}{HTML}{FFFDED}

\begin{table*}[tbp]
\centering
\setlength{\tabcolsep}{2pt}
\vspace*{-0.5em}
\resizebox{\textwidth}{!}{
\begin{tabular}{@{}p{3mm} l cccccccc cccccccc@{}}
\toprule
& \multirow{3}{*}{\textbf{Model}} & \multicolumn{8}{>{\columncolor{forwardbg}}c}{\textbf{Forward World Modeling}} & \multicolumn{8}{>{\columncolor{inversebg}}c}{\textbf{Inverse World Modeling}} \\
\addlinespace[0.5ex]
\cmidrule(l){3-10}\cmidrule(r){11-18}
\addlinespace[0.5ex]
& & \bf 3 & \bf 4 & \bf 5 & \bf 6 & \bf 7 & \bf 8 & \bf 9 & \bf 10 & \bf 3 & \bf 4 & \bf 5 & \bf 6 & \bf 7 & \bf 8 & \bf 9 & \bf 10 \\
\midrule
\multicolumn{18}{>{\columncolor{proprietary}}l}{\textit{\footnotesize Proprietary Models}}\\
& GPT-5 & 84.62 & 75.26 & \colorbox{secondgrey}{69.96} & \colorbox{firstgrey}{64.18} & \colorbox{secondgrey}{57.48} & \colorbox{secondgrey}{52.16} & \colorbox{firstgrey}{49.45} & \colorbox{firstgrey}{46.93} & \colorbox{secondgrey}{86.28} & \colorbox{secondgrey}{80.37} & \colorbox{firstgrey}{76.09} & \colorbox{secondgrey}{68.78} & \colorbox{secondgrey}{65.71} & \colorbox{secondgrey}{62.13} & \colorbox{secondgrey}{57.12} & \colorbox{secondgrey}{55.33} \\
& GPT-5 mini & \colorbox{firstgrey}{87.50} & \colorbox{secondgrey}{76.25} & \colorbox{firstgrey}{70.65} & \colorbox{secondgrey}{63.41} & \colorbox{firstgrey}{58.14} & \colorbox{firstgrey}{52.38} & \colorbox{secondgrey}{46.65} & \colorbox{secondgrey}{44.11} & 85.05 & 76.77 & \colorbox{secondgrey}{75.43} & 67.67 & 63.79 & 57.04 & 55.04 & 50.02 \\
& GPT-5 nano & 67.83 & 50.29 & 38.61 & 30.35 & 25.97 & 21.90 & 17.59 & 16.84 & 72.81 & 53.95 & 42.48 & 36.45 & 31.68 & 28.20 & 24.11 & 20.33 \\
& Gemini 2.5 Pro & \colorbox{secondgrey}{86.10} & \colorbox{firstgrey}{76.42} & 69.83 & 60.80 & 53.26 & 48.12 & 40.12 & 36.98 & \colorbox{firstgrey}{87.94} & \colorbox{firstgrey}{81.18} & 75.39 & \colorbox{firstgrey}{70.03} & \colorbox{firstgrey}{66.03} & \colorbox{firstgrey}{62.91} & \colorbox{firstgrey}{57.78} & \colorbox{firstgrey}{56.62} \\
& Gemini 2.5 Flash & 81.64 & 67.94 & 54.17 & 43.38 & 37.43 & 32.73 & 29.88 & 28.07 & 82.78 & 72.18 & 60.83 & 58.19 & 53.14 & 51.78 & 47.99 & 44.98 \\
& Gemini 2.5 Flash-Lite & 64.34 & 49.07 & 38.70 & 33.87 & 27.81 & 25.44 & 23.31 & 20.31 & 69.58 & 57.55 & 46.04 & 39.09 & 34.06 & 30.18 & 27.51 & 23.16 \\
& Claude Sonnet 4 & 65.65 & 45.82 & 36.65 & 30.52 & 26.61 & 22.78 & 21.49 & 20.16 & 73.25 & 56.85 & 48.87 & 43.07 & 37.00 & 32.71 & 30.50 & 28.49 \\
\midrule
\multicolumn{18}{>{\columncolor{openweight}}l}{\textit{\footnotesize Open-Weight Models}}\\
& GLM-4.5V & 74.30 & 59.99 & 47.65 & 38.78 & 30.83 & 25.69 & 21.60 & 19.67 & \colorbox{secondgrey}{80.59} & \colorbox{secondgrey}{69.28} & \colorbox{secondgrey}{57.04} & \colorbox{secondgrey}{51.53} & \colorbox{secondgrey}{46.95} & \colorbox{secondgrey}{41.68} & \colorbox{secondgrey}{37.36} & \colorbox{secondgrey}{37.93} \\
& Llama-4-Mav-17B-128E-Ins & 72.47 & 52.09 & 43.87 & 35.30 & 29.90 & 25.89 & 22.79 & 20.49 & 72.55 & 62.60 & 50.52 & 43.10 & 35.17 & 31.68 & 28.10 & 25.80 \\
& InternVL3.5-241B-A28B & \colorbox{secondgrey}{75.79} & \colorbox{firstgrey}{62.25} & \colorbox{firstgrey}{50.83} & \colorbox{firstgrey}{45.85} & \colorbox{firstgrey}{37.84} & \colorbox{firstgrey}{32.88} & \colorbox{secondgrey}{27.85} & \colorbox{firstgrey}{25.24} & \colorbox{firstgrey}{82.26} & \colorbox{firstgrey}{70.09} & \colorbox{firstgrey}{60.61} & \colorbox{firstgrey}{53.38} & 45.90 & 39.35 & 34.12 & 30.56 \\
& Gemma-3-27b-it & 63.29 & 44.66 & 32.04 & 25.82 & 22.11 & 19.50 & 16.74 & 16.29 & 64.95 & 48.37 & 40.04 & 33.87 & 28.53 & 23.63 & 21.74 & 19.36 \\
& QVQ-72B-Preview & 69.14 & 52.96 & 40.83 & 36.27 & 33.16 & 30.63 & 26.30 & 24.76 & 71.33 & 58.77 & 48.43 & 44.36 & 40.26 & 39.30 & 36.66 & 36.58 \\
& Qwen2.5-VL-72B-Ins & \colorbox{firstgrey}{78.15} & \colorbox{secondgrey}{60.05} & \colorbox{secondgrey}{49.87} & \colorbox{secondgrey}{41.92} & \colorbox{secondgrey}{36.77} & \colorbox{secondgrey}{31.73} & \colorbox{firstgrey}{28.03} & \colorbox{secondgrey}{25.07} & 77.80 & 65.85 & 53.30 & 48.19 & 44.07 & 37.57 & 33.76 & 36.27 \\
& Qwen2.5-VL-32B-Ins & 67.83 & 55.46 & 44.35 & 35.75 & 27.52 & 26.42 & 22.01 & 18.07 & 63.55 & 59.70 & 54.57 & 51.01 & \colorbox{firstgrey}{49.36} & \colorbox{firstgrey}{47.17} & \colorbox{firstgrey}{41.47} & \colorbox{firstgrey}{40.16} \\
& Ovis2.5-9B & 58.39 & 42.51 & 34.96 & 31.08 & 24.61 & 20.78 & 18.11 & 16.96 & 64.86 & 51.74 & 41.65 & 35.47 & 30.95 & 26.64 & 23.70 & 23.25 \\
& MiniCPM-V-4.5 & 60.75 & 38.73 & 33.65 & 25.47 & 24.81 & 21.40 & 21.56 & 18.33 & 69.23 & 53.08 & 47.35 & 39.55 & 34.87 & 30.63 & 27.05 & 25.71 \\
& Idefics3-8B-Llama3 & 60.23 & 36.99 & 31.83 & 24.25 & 21.29 & 20.80 & 20.46 & 17.71 & 47.38 & 33.86 & 27.26 & 23.48 & 19.87 & 18.50 & 17.04 & 15.16 \\
& Cosmos-Reason1 & 56.28 & 41.86 & 34.75 & 28.40 & 26.46 & 26.49 & 25.41 & 24.88 & 58.30 & 45.93 & 44.25 & 38.50 & 35.72 & 34.56 & 31.50 & 28.64 \\
& BAGEL & 30.24 & 40.19 & 29.65 & 25.37 & 22.75 & 19.45 & 17.84 & 15.87 & 56.73 & 52.85 & 40.09 & 35.44 & 29.67 & 24.39 & 28.70 & 18.91 \\
\midrule
& \textbf{Human Performance} & \textbf{93.62} & \textbf{95.30} & \textbf{95.04} & \textbf{93.87} & \textbf{95.43} & \textbf{95.41} & \textbf{94.75} & \textbf{95.13} & \textbf{92.05} & \textbf{93.56} & \textbf{94.35} & \textbf{94.25} & \textbf{95.96} & \textbf{97.74} & \textbf{96.30} & \textbf{96.29} \\
\bottomrule
\end{tabular}
}
\vspace*{-0.5em}
\caption{\textbf{Evaluation on \name (Pairwise Accuracy).}
{\setlength{\fboxsep}{2pt}\protect\colorbox{gray!60}{\rule{0pt}{1.1ex}Dark gray}}
indicates the best result within each category (Proprietary or Open-Weight Models), and
{\setlength{\fboxsep}{2pt}\protect\colorbox{gray!20}{\rule{0pt}{1.1ex}Light gray}}
denotes the second-best result within the category.
Complete results are in Table~\ref{tab:app_task_acc} (Task Accuracy) and Table~\ref{tab:app_pair_acc} (Pairwise Accuracy).%
}
\label{tab:main_pair_acc}
\vspace*{-1em}
\end{table*}

\section{Experiments and Analysis}
\subsection{World Modeling as a Proxy for Evaluating Embodied Cognition}\label{sec_3:main_exp}
\textbf{Experimental Setup.}
(1) \textit{VLM evaluation setup.}
We evaluate \name with 7 proprietary VLMs from 3 families~\citep{openai2025gpt5systemcard, google2025gemini25pro, anthropic2025claudesonnet4} and 23 open-weight models from 11 families~\citep{wang2025internvl3, bai2025qwen2, hong2025glm, meta2025llama4, team2025gemma, lu2025ovis25, yao2024minicpm, azzolini2025cosmos, qvq-72b-preview, deng2025emerging}. For input, all images are resized to \(512\times512\), and we use a unified prompt template per QA type. Models are instructed to return a parsable Python list encoding a permutation of indices. We apply the online verifier in Section~\ref{sec:2_3} and report Task Accuracy and Pairwise Accuracy.
(2) \textit{Human evaluation setup.}
We also recruit trained annotators to answer the benchmark under the same instructions as the models. For inter-annotator agreement (IAA), we uniformly stratify \(240\) items over QA type and step length and collect independent labels from three annotators. Krippendorff’s \(\alpha = 0.83\) indicates strong agreement. Full details are in Appendix~\ref{app_3_2_1:main_setup} and~\ref{app_3_1_1:human_setup}.

We visualize \textit{Task Accuracy} for GPT\mbox{-}5 and human annotators in Figure~\ref{fig:teaser}. Since many models collapse at long horizons (\(L\!=\!8\)–\(10\), near-zero task success), we focus on the more informative \textit{Pairwise Accuracy}. The main results appear in Table~\ref{tab:main_pair_acc}.

\textbf{Is inverse world modeling easier than forward?}
Across families and step lengths, inverse consistently outperforms forward, with the margin widening as \(L\) grows. For example, GPT\mbox{-}5 and Gemini~2.5~Pro maintain gaps at \(L\ge6\), and open\mbox{-}weight models such as GLM\mbox{-}4.5V and Qwen2.5\mbox{-}VL also show higher inverse scores than forward for most \(L\) (see Table~\ref{tab:main_pair_acc}). This asymmetry suggests that models handle retrospective textual reasoning better than the prospective visual simulation required for forward planning. 

\textbf{How does performance change with step length?}
Accuracy decreases monotonically with \(L\) for nearly every model, proprietary and open\mbox{-}weight alike. Shorter tasks (\(L\le4\)) are manageable for several VLMs, while longer tasks (\(L\ge8\)) are challenging even for the strongest models. Pairwise Accuracy softens but does not alter this downward trend. This sharp performance decay reveals that VLMs struggle to track evolving physical states.

\textbf{Can VLMs achieve near-human performance?} Human performance is far better than any evaluated VLM. SOTA VLMs such as GPT-5 and Gemini-2.5 Pro are comparable to humans only at step length \(L=3\); their performance drops sharply as the horizon grows. This vast gap confirms that, compared with humans, VLMs still struggle with interactive embodied world modeling tasks.

\textbf{What is the performance comparison among VLMs?}
GPT\mbox{-}5 and Gemini~2.5~Pro are the strongest overall in both forward and inverse settings. Several open\mbox{-}weight VLMs are competitive: InternVL3.5\mbox{-}241B\mbox{-}A28B, GLM\mbox{-}4.5V, and Qwen2.5\mbox{-}VL often close much of the gap and even surpass Claude~4~Sonnet in multiple settings. GPT\mbox{-}5~mini is also highly competitive, achieving the best score at several short and mid horizons (e.g., forward at \(L=3,7,8\)). 

\textbf{Does Cosmos-Reason1, trained on embodied data, outperform other similar-sized models?} We compare Cosmos-Reason1-7B and other similar-sized VLMs in Table~\ref{tab:app_pair_acc} and Figure~\ref{fig_app:cosmos}. For similar-sized models, Cosmos-Reason1-7B exhibits more stable and generally better performance than other VLMs when the interaction horizon exceeds 5 steps.

\begin{promptbox}{\faLightbulb\ Key Takeaways: World Modeling as a Proxy for Evaluating Embodied Cognition}
\it
\begin{itemize}[leftmargin=*, after=\vspace{-0.5em}, itemsep=1pt]
\item Higher performance on the inverse task than the forward one highlights stronger language-based retrospection than action-conditioned visual reasoning.
\item Long-horizon degradation reveals limited interactive, spatial memory under partial observability.
\item The human–model gap shows that current VLMs are still far from robust embodied world models in mobile manipulation settings.
\end{itemize}
\end{promptbox}
Beyond these trends, we also confirm that augmenting key-frame selection with contact-based predicates derived from low-level physics yields qualitatively similar behavior (inverse $>$ forward, strong long-horizon degradation); detailed results are provided in Appendix~\ref{app:02_experiments}, Table~\ref{tab:contact_ablation}.

\subsection{Sensitivity to Image Realism}\label{sec:3_2_image}
Since \name is generated in the BEHAVIOR simulator, we ask whether VLMs are sensitive to image realism and whether a sim-to-real gap appears.
To investigate this, we collect additional real-world videos and manually annotate them.
Furthermore, leveraging the automated \name pipeline with simulation, we conduct an in-depth and larger-scale study of how different rendering configurations in the simulator affect performance.

\begin{figure}[t]
    \centering
    \includegraphics[width=\linewidth]{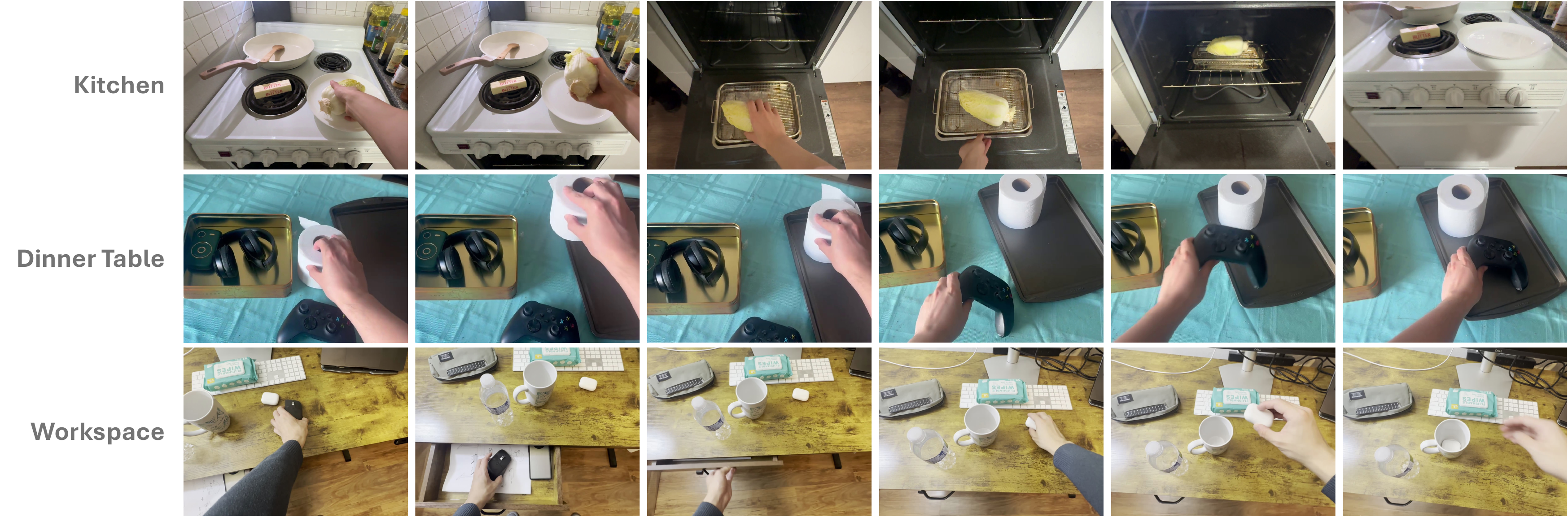}
    \vspace*{-1.5em}
    \caption{\small
    \textbf{Real-World Evaluations.} Key frames from the three real-world scenes used in our evaluation: kitchen, dinner table, and workspace. Together, these scenes contain diverse rigid, deformable, and articulated objects in diverse environments with varying lighting conditions.
    }
    \label{fig:real_world}
    \vspace*{-1em}
\end{figure}

\textbf{Real-World Experimental Setup.} 
To assess whether the simulator findings transfer to real imagery, we construct a real-world benchmark over three everyday scenes (kitchen, dinner table, and workspace). From videos recorded in each scene, we manually select segmented frames and annotate a scene graph for every frame. Applying the \name pipeline to these annotations yields 960 real-world QA pairs, on which we evaluate InternVL3.5-241B-A28B. Examples from the three scenes are shown in Figure~\ref{fig:real_world}.

\textbf{Simulation Experimental Setup.}
(1) \textit{Configuration.}
We use GPT-5~mini as the base model for SOTA VLMs due to its strong cost-performance balance shown in Table~\ref{tab:main_pair_acc}. For diversity, we also evaluate InternVL3.5-241B and report its performance across all settings in Figure~\ref{fig_app:intern_behavior_analysis}. We evaluate step lengths \(L\in\{3,6,9\}\). For each \(L\) and each QA type (forward, inverse), we sample \(50\) items, yielding \(300\) total QAs. We report, for each setting, the Pairwise Accuracy difference
\(\Delta=\mathrm{PA}_{\text{baseline}}-\mathrm{PA}_{\text{variant}}\) and two\mbox{-}tailed unpaired p\mbox{-}values versus the baseline. $|\Delta| < 0.05$ will be considered as a small change.
(2) \textit{Image realism implementation.}
BEHAVIOR uses Isaac Sim~\citep{NVIDIA_Isaac_Sim}, our \emph{baseline} uses Ray Tracing~\citep{nvidia_rtx} with default global effects. We evaluate three alternatives on a realism spectrum: \emph{Realistic} (segmented frames translated to a real\mbox{-}world style using GPT\mbox{-}image\mbox{-}1~\cite{openai2025gptimage1}),
\emph{Path Tracing} (higher-fidelity rendering, \citet{kajiya1986rendering}), and
\emph{Ray Tracing Only} (Ray Tracing with global effects such as reflections and stage lights disabled).
 Detailed setup, prompts, and examples are in the Appendix~\ref{app_3_3:image_realism}. Results are summarized in Figure~\ref{fig:behavior_analysis} (panel A).

\textbf{Does real-world data change performance?}
As summarized in Table~\ref{tab:app_realworld_eval}, InternVL3.5-241B-A28B exhibits similar absolute accuracy on real-world data as in the simulator. On real videos, inverse queries consistently outperform forward ones, and accuracy drops sharply as the temporal horizon increases. Overall, real-world evaluation does not reveal a significant sim-to-real gap, and the empirical trends are fully consistent with those observed in the simulator. This consistency validates the simulation as a faithful proxy for evaluating real-world embodied cognition.

\begin{table*}[!htbp]
\centering
\setlength{\tabcolsep}{2pt}
\small
\resizebox{\textwidth}{!}{
\begin{tabular}{@{}p{3mm} l cccccccc cccccccc@{}}
\toprule
& \multirow{3}{*}{\textbf{Metric}} &
\multicolumn{8}{>{\columncolor{forwardbg}}c}{\textbf{Forward (Real-World)}} &
\multicolumn{8}{>{\columncolor{inversebg}}c}{\textbf{Inverse (Real-World)}} \\
\addlinespace[0.5ex]
\cmidrule(l){3-10}\cmidrule(r){11-18}
\addlinespace[0.5ex]
& & \bf 3 & \bf 4 & \bf 5 & \bf 6 & \bf 7 & \bf 8 & \bf 9 & \bf 10
  & \bf 3 & \bf 4 & \bf 5 & \bf 6 & \bf 7 & \bf 8 & \bf 9 & \bf 10 \\
\midrule
& Task Accuracy     & 73.33 & 50.00 & 33.33 & 13.33 &  8.33 &  3.33 & 0.00 & 0.00
                    & 90.00 & 85.00 & 55.00 & 38.33 & 21.67 & 13.33 & 6.67 & 3.33 \\
& Pairwise Accuracy & 80.00 & 67.78 & 61.25 & 49.00 & 38.89 & 37.14 & 26.88 & 25.74
                    & 90.00 & 88.33 & 72.92 & 57.67 & 44.17 & 43.57 & 31.04 & 26.85 \\
\bottomrule
\end{tabular}
}
\caption{\textbf{Real-world evaluation of InternVL3.5-241B-A28B.}
Task and pairwise accuracies (\%) on 960 QA pairs generated from real-world manipulation videos.
We report performance across different interaction horizons for both forward and inverse world modeling.}
\label{tab:app_realworld_eval}
\end{table*}

\textbf{Does rendering realism change performance?}
We find no statistically significant degradation or improvement across the spectrum. All settings have \(p\!\ge\!0.2\) relative to the baseline, and observed deltas are small across both QA types and all step lengths (Figure~\ref{fig:behavior_analysis}, A; Figure~\ref{fig_app:intern_behavior_analysis}, A), suggesting that the evaluated VLMs are robust to rendering variations in world modeling tasks.

\begin{promptbox}{\faLightbulb\ Key Takeaways: Image Realism}
\it
\begin{itemize}[leftmargin=*, after=\vspace{-0.5em}, itemsep=1pt]
\item Real-world evaluation mirrors our simulation findings with minimal sim-to-real gap, while simulation offers a more controlled and reproducible testbed.
\item Insensitivity to rendering variations indicates bottlenecks in multi-step interaction reasoning rather than low-level image realism.
\end{itemize}
\end{promptbox}

\subsection{Do VLMs Exhibit Anthropocentric Bias on Human Vision?}\label{sec:3_3_camera}

VLMs are mostly trained on RGB images that mirror how humans typically see the world. However, different embodiments may have diverse camera configurations. We therefore test whether VLM performance is sensitive to camera configuration, i.e., if dataset bias is present.

\begin{figure}[tbp]
    \centering
    \includegraphics[width=\linewidth]{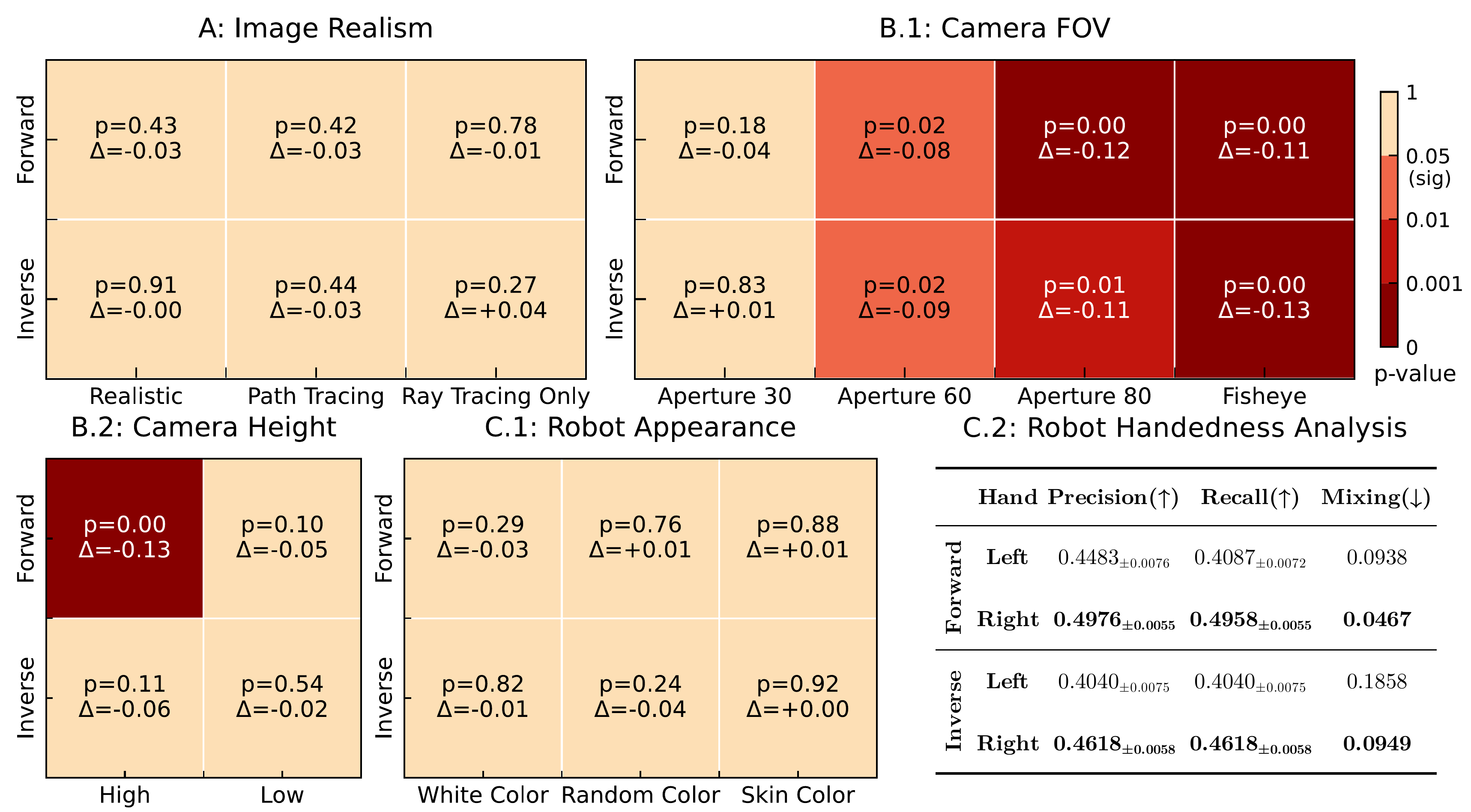}
    \vspace*{-1.5em}
    \caption{\textbf{Evaluations on image realism and anthropocentric bias on human vision through \name.} Heatmaps show two-tailed unpaired t-test results against the baseline, using \textit{Pairwise Accuracy}. $p<0.05$ is considered \textit{significant}. Darker red means more significant. $\Delta$ is the performance change from the baseline. If \textit{significant} and $\Delta<0$, the setting is worse than the baseline. C.2 reports the robot's performance on the left- and right-hand predicates, where \textit{Mixing} is the proportion of ground truth left or right cases that are predicted as the other hand (i.e., mixing one hand into the other). $\pm$ means standard error.}
    \label{fig:behavior_analysis}
    \vspace*{-1.0em}
\end{figure}

\textbf{Experimental Setup.}
(1) \textit{Configuration.}
We reuse the setup from Section~\ref{sec:3_2_image}. We use GPT\mbox{-}5~mini as the base VLM, and report InternVL3.5-241B in the Appendix~\ref{app_3_4:camera_config}.
(2) \textit{Camera FOV.}
The baseline is Aperture 40. We examine Aperture 30, 60, 80, and Fisheye. Rendering and all other parameters are held fixed.
(3) \textit{Camera Height.}
The baseline is \((1.75\mathrm{\,m})\) high for eye\mbox{-}level view used in Behavior replays. We test High \((+0.5\mathrm{\,m})\) and Low \((-0.25\mathrm{\,m})\). We choose \((-0.25\mathrm{\,m})\) since a lower height will consistently make relevant objects invisible. Examples are in the Appendix~\ref{app_3_4:camera_config}. Results are summarized in Figure~\ref{fig:behavior_analysis} (panels B.1 and B.2).

\textbf{Does field of view matter?}
Figure~\ref{fig:behavior_analysis} (B.1) shows that a small change to \(\text{Aperture }30\) shows no significant difference from baseline \((p>0.1)\).
Larger deviations hurt performance: \(\text{Aperture }60, 80\), and Fisheye are consistently and significantly worse than baseline across QA types and step lengths \((p\le0.01)\), suggesting that the model performs better with human-like intrinsics.

\textbf{Does camera height matter?}
As shown in Figure~\ref{fig:behavior_analysis} (B.2), increasing the camera height (\textit{High}) significantly reduces GPT-5 mini's accuracy in the forward setting with $\Delta=-0.13$. By contrast, the \textit{High} inverse setting shows no statistically significant change, though with a performance drop $\Delta=-0.06$. For the \textit{Low} camera, both forward and inverse are not significantly different from the baseline, likely because the $-0.25\ \mathrm{m}$ shift remains within normal human height variation.

\begin{promptbox}{\faLightbulb\ Key Takeaways: Anthropocentric Bias on Human Vision}
\it
\begin{itemize}[leftmargin=*, after=\vspace{-0.5em}, itemsep=1pt]
\item Performance degradation on non-standard views implies VLMs are biased towards human-like egocentric viewpoints and intrinsics.
\item Reliance on human-centric visual priors limits generalization to diverse robotic embodiments with non-standard optics.
\end{itemize}
\end{promptbox}

\subsection{Do VLMs Exhibit Anthropocentric Bias on Embodiment?}
\label{sec:3_4}
To further understand VLM embodiment, we investigate two potential biases: \textbf{self-awareness} regarding the robot's own body and \textbf{handedness asymmetry}, a common trait in humans.

\textbf{Experimental Setup.}
We study these two aspects using distinct experimental setups. (1) \textit{Robot Appearance}. To test for self-awareness, we assess whether VLMs can recognize their embodiment regardless of its appearance. We reuse the experimental configuration from Section~\ref{sec:3_2_image}, with GPT-5~mini as the base model. The baseline is the default black-and-white robot appearance. We test three variants: White Color, Random Color (robot color is randomized at each frame), and Skin Color (robot is rendered with a human-like skin tone). (2) \textit{Handedness Asymmetry.} Inspired by human motor control, where approximately 89\% of the population is right-handed~\citep{papadatou2020human}, we investigate if VLMs exhibit a similar ``dominant hand''. We analyze this configuration with a predicate-level error analysis of all tested VLMs and report GPT-5 mini in Figure~\ref{fig:behavior_analysis}. We isolate all errors related to the \texttt{LeftGrasping} and \texttt{RightGrasping} predicates. Using the framework described in Section~\ref{sec:3_5_error}, we frame our metrics in terms of \textit{Precision} and \textit{Recall}. We also report \textit{Mixing Rate}, which measures the proportion of ground-truth state differences for one hand that the model incorrectly attributes to the other. Higher precision and recall with lower mixing indicate greater proficiency. Appearance examples and handedness analysis are in the Appendix~\ref{app_3_5_1:robot_appearance} and~\ref{app_3_6_2:hand}.

\textbf{Are VLMs aware of their own embodiment, and is this awareness robust to changes in their visual appearance?}
As shown in Figure~\ref{fig:behavior_analysis} and Figure~\ref{fig_app:intern_behavior_analysis} (panel C.1), altering the robot's appearance has no statistically significant impact on performance for both GPT-5 mini and InternVL3.5-241B. For all variants (White, Random, Skin Color), the performance deltas are small ($|\Delta|<0.05$) and non-significant (all $p>0.10$), suggesting that the model's understanding of its interaction with the world is not tied to a specific visual representation of its body.

\textbf{Do VLMs exhibit a handedness asymmetry in their interactions with the world?}
Our analysis of hand-related errors, summarized in Figure~\ref{fig:behavior_analysis} (panel C.2), reveals a consistent and strong asymmetry (complete error results are shown in Figure~\ref{fig:full_mixing_forward} and~\ref{fig:full_mixing_inv}). For both forward and inverse tasks, the right hand consistently outperforms the left hand across all metrics. Precision and recall are substantially higher for the right hand, while the mixing rate is significantly lower. For instance, in the forward task, 9.38\% of true left-hand changes were incorrectly identified as right-hand changes, whereas only 4.67\% of right-hand changes were misattributed to the left. Full analysis is in Appendix~\ref{app_4_1_3:handedness}.

\begin{promptbox}{\faLightbulb\ Key Takeaways: Anthropocentric bias on Embodiment}
\it
\begin{itemize}[leftmargin=*, after=\vspace{-0.5em}, itemsep=1pt]
\item VLMs are robust to the embodiment appearance variations.
\item VLMs exhibit strong right-handed bias, which is consistent with human handedness distribution.
\end{itemize}
\end{promptbox}

\subsection{Error Analysis} \label{sec:3_5_error}
\subsubsection{Experimental Setup}
To gain a deeper insight into the reasoning failures of VLMs, we design a systematic error analysis framework. Evaluating errors directly from output permutations (e.g., comparing predicted order [3, 2, 1] to ground truth [2, 3, 1]) is difficult and often uninformative about the underlying cognitive mistakes.
Exploiting the fact ground-truth symbolic scene graphs are readily accessible in simulation, we instead convert the model's output into a format that allows for a structured, fine-grained comparison with the ground truth.
For the \textbf{forward world modeling} task, we take the model's predicted permutation of images \((o'_{\sigma(1)},\ldots,o'_{\sigma(L-1)})=(o_1,\ldots,o_{L-1})\) and compute the corresponding sequence of actions (i.e., visible state differences) that this ordering implies: \(\hat{a}_k:=\Delta_\mathrm{Vis}(s'_{\sigma(k+1)}, s'_{\sigma(k)})\). This yields a predicted action sequence \((\hat{a}_0,\cdots,\hat{a}_{L-2})\). For the \textbf{inverse world modeling} task, the model directly outputs a predicted action sequence.

With both a predicted and a ground-truth action sequence, we can perform a pairwise comparison at each step $k$. Each action $a_k$ is a \textit{set} of atomic state differences (e.g., \texttt{\{Add Open(fridge), Remove Inside(basket, cabinet)\}}). By comparing the predicted set $\hat{a}_k$ with the grounded-truth set $a_k$, we can categorize each atomic state difference. This comparison, similar to analyzing a Venn diagram, yields three primary outcomes for each ground-truth state difference: (1) \textit{Correct}: The state difference is present in both the ground-truth and predicted sets. (2) \textit{Omission}: The state difference is in the ground-truth set but missing from the prediction. (3) \textit{Hallucination}: The state difference is in the predicted set but not in the ground truth. Detailed setup is in Appendix~\ref{app:03_error_analysis}.
We assume each state difference is an \textit{independent event} and aggregate these counts across all actions and all questions in the dataset. Based on this framework, we classify errors into five main categories:
\vspace*{-0.5em}
\begin{enumerate}[leftmargin=2em, after=\vspace{-0.5em}, itemsep=1pt]
    \item \textbf{Entity Substitution.} The model correctly identifies the state change predicate but applies it to the wrong object(s).
    \item \textbf{Polarity Inversion.} The model correctly identifies both the object(s) and the predicate, but reverses the polarity of the change (e.g., `remove' instead of `add').
    \item \textbf{Predicate Substitution.} The model correctly identifies the object(s) involved but describes the state change with an incorrect predicate.
    \item \textbf{Hallucination.} The model predicts a state change that did not occur in the ground truth.
    \item \textbf{Omission.} The model fails to predict a ground-truth state change that occurred.
\end{enumerate}

\subsubsection{Error Distribution Analysis}

\begin{wrapfigure}{r}{0.5\linewidth}
    \vspace{-15pt} %
    \centering
    \includegraphics[width=\linewidth]{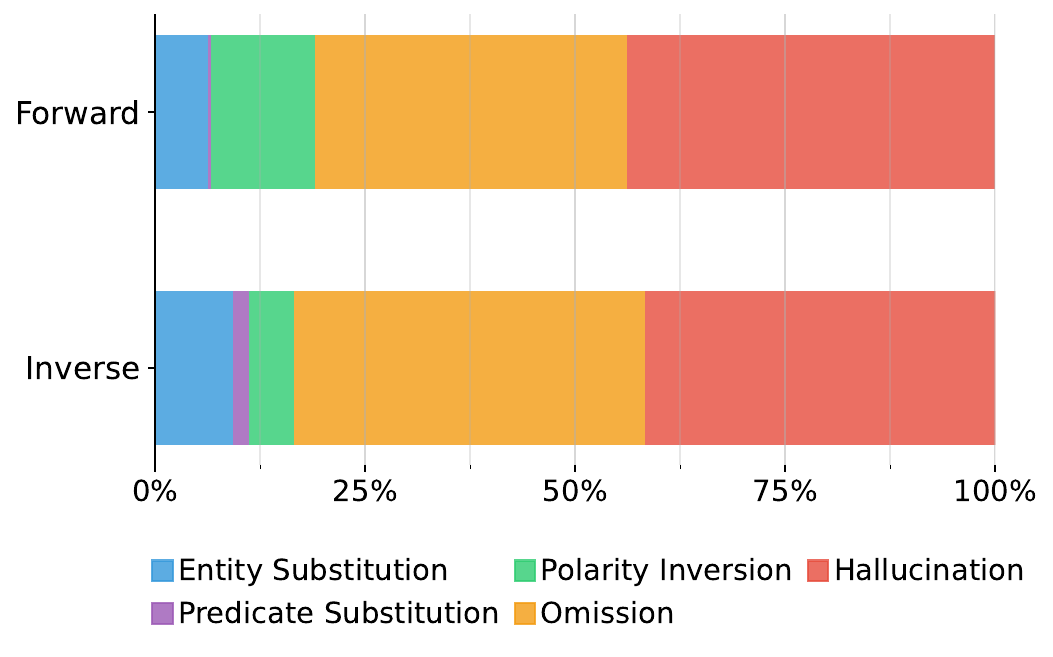}
    \vspace*{-2em}
    \caption{\textbf{Error Distribution}, broken down by forward and inverse tasks, evaluated on GPT-5.}
    \label{fig:error_analysis_gpt5}
    \vspace{-0.7em}
\end{wrapfigure}
Our error analysis for GPT-5, shown in Figure~\ref{fig:error_analysis_gpt5}, reveals that the vast majority of errors fall into two main categories: \textbf{Omission} and \textbf{Hallucination}. For the forward task, these two error types account for a combined 81\% of all failures. This figure is even higher for the inverse task, where they make up nearly 84\% of errors. This indicates that the model's primary challenge is not misinterpreting the specifics of a known state change, but correctly identifying which changes occurred and which did not.

While Omission and Hallucination errors are dominant in both settings, their distribution shifts between tasks. In forward modeling, \textbf{Hallucination} is the most common error at 43.9\%, followed by \textbf{Omission} at 37.1\%. Remarkably, in the inverse task, these two errors are perfectly balanced, each accounting for exactly 41.8\% of all failures.
Other error types are far less frequent. \textbf{Polarity Inversion} is more common in the forward setting (12.4\%) than the inverse (9.2\%). \textbf{Entity Substitution} is also slightly more prevalent in the forward task (6.3\% vs. 5.4\%). Finally, \textbf{Predicate Substitution} remains the rarest error type, though it is more pronounced in the inverse setting (1.9\%) compared to the forward task (0.3\%). Detailed analysis is in Appendix~\ref{app:03_error_analysis}.

\begin{promptbox}{\faLightbulb\ Key Takeaways: Structural Error Analysis}
\it
\begin{itemize}[leftmargin=*, after=\vspace{-0.5em}, itemsep=1pt]
\item The dominance of hallucinations suggests models rely on learned textual priors rather than faithful visual grounding.
\item High omission rates reveal a fundamental deficit in tracking object persistence across complex state transitions under egocentric partial observation.
\end{itemize}
\end{promptbox}
\vspace*{-1em}

\section{Related Work}
\textbf{Embodied Cognition.} The theory of \emph{Embodied Cognition} argues that intelligence arises from an agent's sensorimotor interaction within its environment, grounding abstract knowledge in perception and action \citep{gibson2014ecological, varela2017embodied, clark1998being, brooks1991intelligence, o2001sensorimotor, barsalou1999perceptual, lakoff2008metaphors}.
Grounded in the POMDP framework for decision making~\citep{kaelbling1998planning}, our work focuses on how spatial perception, physical interaction, and linguistic abstraction jointly support embodied cognition \citep{frick2016matter, thompson2005sensorimotor, clark2006language, barsalou2020challenges}.
Rather than enumerating individual capabilities, we adopt a unified lens of world modeling through egocentric interaction, using controlled, reproducible, and scalable simulation to contrast the behavior of current VLMs with humans.

\textbf{World Modeling.}
World models learn action-conditioned dynamics for imagination and planning \citep{ha2018world,hafner2019learning}, achieving scalable gains from counterfactual rollouts \citep{hafner2023mastering,bruce2024genie,agarwal2025cosmos,janner2022planning, wan2025worldagen}. Many recent systems emphasize generative fidelity and long-horizon prediction in video or latent space \citep{bruce2024genie,agarwal2025cosmos,finn2017deep,ebert2018visual}. Complementary benchmarks evaluate control and prediction quality in visually rich settings \citep{tian2023control,chi2024eva,yue2025ewmbench}, or study physical and scene understanding from non-interactive visual data \citep{bakhtin2019phyre,yi2019clevrer,girdhar2020cater, bear2021physion,tung2023physion++,li2024behavior1k,dang2025ecbench,yang2025embodiedbench}. Others focus on sequence-level temporal coherence and ordering \citep{qin2024worldsimbench,chen2025worldprediction}. Aurora-Bench~\citep{qiu2025bootstrapping} focuses on short-horizon and general-purpose video forward and inverse world modeling. As argued by \citet{xing2025critiques}, a world model should serve as a sandbox for reasoning. Our benchmark is therefore designed to study forward and inverse ordering with an explicitly defined action space and scalable construction.

\textbf{VLMs in Embodied AI.}
VLMs are central to embodied agents, acting as high-level planners \citep{huang2022language,ahn2022can,huang2023voxposer,huang2022inner,liang2022code,huang2023instruct2act,huang2024rekep,jiang2024roboexp, wang2025vagen} or end-to-end policies \citep{zitkovich2023rt,kim2024openvla,team2024octo,driess2023palm}.
However, current applications are often confined to settings where involved reasoning is not required~\citep{lynch2023interactive}.
Correspondingly, many benchmarks emphasize instruction-following and goal-conditioned control \citep{das2018embodied,padmakumar2022teach,mees2022calvin,fan2022minedojo,sermanet2024robovqa, li2024embodied,yang2025embodiedbench,gao2025vision}, with less focus on the multi-step, consequence-aware reasoning essential for long-horizon interaction.
We address this gap with a benchmark that uses egocentric interaction to evaluate an agent's forward and inverse world modeling, requiring long-range interactive reasoning.
GVL~\citep{ma2024vision} casts value estimation as a reordering task over observation sequences, whereas \name focuses on transition modeling over interleaved observation-action sequences with an explicitly defined scene graph action space that applies to interactions of any quality.

\section{Conclusions and Limitations}
\paragraph{Conclusions.}
In this work, we introduced \name, a benchmark designed to evaluate the extent to which embodied cognition emerges in VLMs trained on passive datasets. By framing evaluation as forward and inverse world modeling from egocentric interaction, \name assesses a model's understanding of environmental dynamics and the consequences of its actions. Grounded in a POMDP, we cast this as two sequence-reordering tasks: forward world modeling, which predicts an ordered sequence of future states from actions, and inverse world modeling, which infers an ordered sequence of actions from state changes.
Our extensive experiments reveal a significant performance gap between state-of-the-art VLMs and humans, a gap that widens dramatically as the interaction horizon increases. We consistently found that models solve the inverse problem more effectively than the forward one. Furthermore, our analysis uncovered strong embodied biases within these models, including a preference for right-handed actions and a significant performance drop with non-human-like camera perspectives. An in-depth error analysis showed that reasoning failures are primarily driven by the omission and hallucination of state changes. \name provides a scalable and insightful tool for charting a course toward more genuinely embodied artificial intelligence.

\paragraph{Limitations. } 
Our work has limitations primarily related to its scope. First, while we introduce several diagnostic tasks that reveal key model biases, this set is not exhaustive. The experiments on factors like camera configuration and agent appearance serve as foundational examples, but the \name framework is designed to be an extensible tool. It can support future, more complex investigations into a much broader spectrum of different embodied-related settings. Second, due to the significant computational cost of evaluation on VLMs, the in-depth ablation experiments were necessarily focused on a representative subset of models and data. A broader evaluation across more architectures and larger data scales would be beneficial to generalize our findings. Furthermore, we do not explore finetuning VLM in this work, but we expect our automatic and scalable dataset can also bring benefits to improving VLM's embodied world modeling abilities. Additionally, due to the frequent physical inconsistency of generated rollouts and the difficulty of designing fair evaluation metrics, which often require costly human studies, we do not evaluate video generative models on \name (for unified VLMs, we evaluate BAGEL~\citep{deng2025emerging} and the result is shown in Table~\ref{tab:app_task_acc} and~\ref{tab:app_pair_acc}).

\section*{Ethics Statement}

The ENACT benchmark was generated in the BEHAVIOR simulator to avoid the privacy risks associated with real-world human data; it contains no human subjects or personally identifiable information. All human annotators hired for evaluation were compensated at rates significantly exceeding their local minimum wage and were not exposed to any sensitive content.

We acknowledge that the simulator may not fully capture the complexity of real-world environments, which can introduce biases and limit the generalizability of our findings. Furthermore, the large-scale models we evaluate carry a significant computational and environmental cost. While ENACT is intended for academic research, we recognize that the technologies it helps develop could have dual-use applications.

\section*{Reproducibility Statement}

To ensure full reproducibility, our complete codebase is available at \href{https://github.com/mll-lab-nu/ENACT}{our Github Repository}. This repository contains all scripts for data generation using the BEHAVIOR simulator \citep{li2024behavior1k}, evaluation of all Vision-Language Models, and analysis. Our implementation includes the automated verifier, prompt templates, and the code to replicate our main experiments, controlled ablation studies (Sections~\ref{sec:3_2_image}, \ref{sec:3_3_camera}, and~\ref{sec:3_4}), and human baseline evaluation. The full ENACT dataset is also publicly available.

\bibliography{iclr2026_conference}
\bibliographystyle{iclr2026_conference}
\clearpage
\appendix
\renewcommand{\partname}{}
\part{Appendix} %
\parttoc %
\clearpage

\section{\name: Egocentric Interactive Embodied Cognition Test}
\label{app:01_benchmark}

\subsection{Notations}
We list all the notations we used across the entire paper in the following two tables.

\newcolumntype{Y}{>{\raggedright\arraybackslash}X}

\begin{table*}[htbp]
\centering
\footnotesize
\setlength{\tabcolsep}{4pt}
\renewcommand{\arraystretch}{1.08}

\begin{tabularx}{\textwidth}{@{}lY lY@{}}
\toprule
\textbf{Notation} & \textbf{Short description} & \textbf{Notation} & \textbf{Short description} \\
\midrule
$T$ & \# frames in a raw replay & $H,W$ & image height and width \\
$[M]$ & index set $\{1,2,\dots,M\}$ & $M$ & \# segmented key frames \\
$\mathcal{K}$ & segmented timestamps $\{t_1<\cdots<t_M\}$ & $(o_i,s_i)$ & RGB \& scene graph at timestamp $t_i$ \\
$o_t$ & RGB image at time $t$ & $s_t$ & scene graph at time $t$ \\
$\mathcal{G}$ & space of scene graphs & $L$ & target trajectory length (steps) \\
$R$ & \# sampled trajectories & $E$ & adjacency matrix on $[M]$ (DAG) \\
$\mathrm{Adj}(i)$ & successors of node $i$ & $|E|$ & \# edges in the DAG \\
$\delta(\cdot,\cdot)$ & scene-graph difference (see long) & $\mathrm{Vis}(\cdot)$ & visibility predicate (see long) \\
$\Delta_{\mathrm{vis}}(\cdot,\cdot)$ & visible-change extractor (see long) & $\mathcal{A}$ & action space \\
$a_i$ & local action $s_{i+1}-s_i$ & $a_{i\to j}$ & action from $i$ to $j$ \\
$\pi$ & key-frame trajectory (see long) & $S_\pi$ & state sequence along $\pi$ \\
$A_\pi$ & action sequence along $\pi$ & $\Pi$ & set of sampled trajectories \\
$DP[\ell,i]$ & \# paths of length $\ell$ ending at $i$ & $w_i$ & end-node weight $DP[L,i]$ \\
$\mathcal{P}$ & predecessor set in backtracking & $\mathrm{Categorical}(w)$ & weighted discrete distribution \\
$\mathbf{1}\{\cdot\}$ & Iverson bracket (true=1, false=0) & $\mathcal{D}$ & datasets \\
$c$ & component in signature $a_i^{sig}$ & $\gamma, \mathrm{transition}$ & The operation key in component \\
$e$ & The entity involved in component & $\rho$ & The predicate in component\\
\bottomrule
\end{tabularx}

\vspace{4pt}

\begin{tabularx}{\textwidth}{@{}lX@{}}
\toprule
\textbf{Notation} & \textbf{Longer description} \\
\midrule
$\mathcal{T}=\{(o_t,s_t)\}_{t=1}^{T}$ & Raw replay trajectory with RGB observations $o_t\in\mathbb{R}^{H\times W\times 3}$ and scene graphs $s_t\in\mathcal{G}$. \\
$\delta(s_i,s_j)$ & A difference operator over scene graphs summarizing semantic changes (objects, relations, attributes) between frames $i$ and $j$. \\
$\mathrm{Vis}(\delta(s_i,s_j))$ & Predicate returning $1$ iff the semantic difference is visually verifiable; induces an edge $i\!\to\!j$ when $i<j$ and the predicate is true (frame skipping allowed). \\
$\Delta_{\mathrm{vis}}(s_i,s_j)\in\mathcal{A}\cup\{\varnothing\}$ & Action-level representation extracted from $\delta(s_i,s_j)$; may be atomic or composite and can be empty when no visible semantic change exists. \\
$\pi=(i_1,\ldots,i_L)$ & Key-frame trajectory: strictly increasing indices with valid edges $E_{i_\ell,i_{\ell+1}}=1$ for all $\ell=1,\dots,L-1$. \\
$S_\pi,\,A_\pi$ & Sequences induced by $\pi$: $S_\pi=(s_{i_1},\ldots,s_{i_L})$, \quad $A_\pi=(a_{i_1\to i_2},\ldots,a_{i_{L-1}\to i_L})$. \\
$DP[\ell,i]$ recurrence & Base: $DP[1,i]=1$. \ \ Recurrence: $DP[\ell,i]=\sum_{j<i} DP[\ell-1,j]\cdot E_{ji}$ for $\ell=2,\ldots,L$. \\
$\mathcal{M}_\pi=\langle \{s_{i_\ell}\},\{a_{i_\ell\to i_{\ell+1}}\},P\rangle$ & Deterministic finite-horizon fragment induced by $\pi$ with transition $P(s_{i_\ell},a_{i_\ell\to i_{\ell+1}})=s_{i_{\ell+1}}$. \\
$a_{i}^{sig}$ & Signature corresponding to an action $a_i$, transformed from natural language to predicate-based structural format.\\

\bottomrule
\end{tabularx}
\vspace*{1em}
\caption{Notation used throughout the paper.}
\label{tab:notation}
\end{table*}

\subsection{Key-Frame Trajectories Synthesis for Scalable Data Generation}

\subsubsection{Segmented Frames with Abstract State Changes}\label{app_1_1:seg_frame}
We provide examples of a scene graph and our scene graph differences for two adjacent segmented frames, shown in Figure~\ref{app_fig:scene_graph_example} and~\ref{app_fig:scene_graph_diff_example}.

Our frame selection process is iterative. For a previously selected key-frame at time $t_{i-1}$, we search for the earliest subsequent frame $t_k$ that satisfies a set of criteria designed to ensure semantic significance and visual clarity.

First, to handle discrepancies where the rule-based simulator updates the scene graph before a change is visually apparent (e.g., registering an object as `\texttt{OnTop}` upon initial contact), we introduce a temporal stability filter: a state change is only considered a candidate if the resulting new state persists for \textbf{at least} 40 frames. At our simulator’s 30Hz rate, this corresponds to $\approx$1.3s, which is consistent with cognitive science findings that humans update attentional sub-events on the order of $\sim$1s~\citep{wyble2009attentional, gavazzi2013time}, and empirically yields reliable keyframe segmentation for our home-scale manipulation tasks. This value is a tunable hyperparameter rather than a hard constraint. It can be adjusted for other environments or model classes within the same automated \name pipeline.

Second, to prevent the recording of minor, oscillatory state changes, such as those that might occur from vibrations when a robot carries an object (e.g., a plate with a pizza), we employ a filtering algorithm to suppress these small fluctuations in the scene graph.

Finally, to ensure that each selected key-frame represents a sufficiently distinct change from the previous one, we implement a similarity check. We convert the scene graph difference between the last selected frame $t_{i-1}$ and a candidate frame $t_k$ into a one-hot vector, which serves as a unique signature for that state change. We then compute the cosine similarity between the signature of the change at $t_k$ and the signature of the previously accepted change at $t_{i-1}$. We aim to find a balance between maximizing the number of segmented frames and ensuring each frame depicts a clearly visible state change. Through empirical evaluation, we determined a cosine similarity threshold of \textbf{0.97}. A candidate frame $t_k$ is accepted only if its change signature's similarity to the previous one is below this threshold. This method effectively filters out near-duplicate frames while retaining a rich, sequential set of key-frames that clearly chronicle the task's progression.

\begin{figure}[htbp]
\centering
\begin{tcolorbox}[colback=black!5!white, colframe=black!75!white,
  title=A Scene Graph Example,
  boxrule=0.5mm, width=\textwidth, arc=2mm, auto outer arc=true]
  \scriptsize
  \begin{verbatim}
{
  'nodes': [
    {'name': 'robot_r1', 'category': 'agent', 'states': []},
    {'name': 'plate_94', 'category': 'plate', 'states': []},
    {'name': 'plate_93', 'category': 'plate', 'states': []},
    {'name': 'bowl_92', 'category': 'bowl', 'states': []},
    {'name': 'bowl_91', 'category': 'bowl', 'states': []},
    {'name': 'pizza_90', 'category': 'pizza', 'states': []},
    {'name': 'pizza_89', 'category': 'pizza', 'states': []},
    {'name': 'floors_zqjkvm_0', 'category': 'floors', 'states': []},
    {'name': 'breakfast_table_xftrki_0', 'category': 'breakfast_table', 'states': []},
    {'name': 'fridge_petcxr_0', 'category': 'fridge', 'states': ['Open']},
    {'name': 'drop_in_sink_lkklqs_0', 'category': 'drop_in_sink', 'states': []},
    {'name': 'straight_chair_uofiqj_0', 'category': 'straight_chair', 'states': []},
    {'name': 'bottom_cabinet_rhdbzv_0', 'category': 'bottom_cabinet', 'states': []}
  ],
  'Edges': [
    {'from': 'robot_r1', 'to': 'plate_93', 'states': ['RightGrasping']},
    {'from': 'plate_94', 'to': 'pizza_90', 'states': ['Under']},
    {'from': 'plate_94', 'to': 'breakfast_table_xftrki_0', 'states': ['OnTop']},
    {'from': 'bowl_92', 'to': 'breakfast_table_xftrki_0', 'states': ['OnTop']},
    {'from': 'bowl_91', 'to': 'breakfast_table_xftrki_0', 'states': ['OnTop']},
    {'from': 'pizza_90', 'to': 'plate_94', 'states': ['OnTop']},
    {'from': 'pizza_89', 'to': 'plate_93', 'states': ['OnTop']},
    {'from': 'breakfast_table_xftrki_0', 'to': 'plate_94', 'states': ['Under']},
    {'from': 'breakfast_table_xftrki_0', 'to': 'floors_zqjkvm_0', 'states': ['OnTop']},
    {'from': 'straight_chair_uofiqj_0', 'to': 'floors_zqjkvm_0', 'states': ['OnTop']}
  ]
}
  \end{verbatim}
\end{tcolorbox}
\caption{A scene graph representation detailing the entities (\textbf{nodes}) and their semantic or physical connections (\textbf{edges}) within the BEHAVIOR~\citep{li2024behavior1k} environment.}
\label{app_fig:scene_graph_example}
\end{figure}
\begin{figure}[htbp]
\centering
\begin{tcolorbox}[colback=black!5!white, colframe=black!75!white,
  title=A Scene Graph Difference Example,
  boxrule=0.5mm, width=\textwidth, arc=2mm, auto outer arc=true]
  \scriptsize
\begin{verbatim}
'2442': 
{
  'type': 'diff',
  'add': {
    'nodes': [],
    'edges': [
      {'from': 'robot_r1', 'to': 'plate_93', 'states': ['RightGrasping']}
    ]
  },
  'remove': {
    'nodes': [],
    'edges': [
      {'from': 'plate_93', 'to': 'pizza_89', 'states': ['Under']},
      {'from': 'plate_93', 'to': 'breakfast_table_xftrki_0', 'states': ['OnTop']}
    ]
  }
}
\end{verbatim}
\end{tcolorbox}
\caption{An example of a scene graph difference, representing a state change by specifying the \textbf{added} and \textbf{removed} edges between objects.}
\label{app_fig:scene_graph_diff_example}
\end{figure}

\subsubsection{Key-Frame Trajectories Synthesis}
\label{app_1_2:key_frame}

\SetKw{KwDownTo}{down to}
\SetKw{KwBreak}{break}

\begin{algorithm}[t]
\caption{KFTS: Key-Frame Trajectory Sampling}
\label{alg:kfts-wrap}
\DontPrintSemicolon
\KwIn{Segmented frames $\{(o_i,s_i)\}_{i=1}^M$, step length $L\ge2$, samples $R$, predicate $\mathrm{Vis}$}
\KwOut{Set of key-frame trajectories $\Pi$}

\textbf{Build DAG}: \For{$1\le i<j\le M$}{ $E_{ij}\leftarrow [\,\mathrm{Vis}(\delta(s_i,s_j))\,]$ }

\textbf{DP counting}: initialize $DP[1,i]\leftarrow 1$; 
\For{$\ell=2..L$}{ \For{$i=1..M$}{ $DP[\ell,i]\leftarrow \sum_{j<i} DP[\ell-1,j]\cdot E_{ji}$ } }

\textbf{Weights}: $w_i\leftarrow DP[L,i]$; \If{$\sum_i w_i=0$}{\Return $\emptyset$}

\textbf{Weighted backtracking sampling}: $\Pi\leftarrow \emptyset$; sample $R$ end-nodes $i_L^{(r)}\sim \mathrm{Categorical}(w)$\;
\For{$r=1..R$}{
  $\pi\leftarrow [\,i_L^{(r)}\,]$, $cur\leftarrow i_L^{(r)}$\;
  \For{$\ell=L..2$}{
    $\mathcal{P}\leftarrow \{\, j<cur \mid E_{j,cur}=1 \land DP[\ell-1,j]>0 \,\}$\;
    \If{$\mathcal{P}=\emptyset$}{\textbf{break}}
    sample $j^\star\in\mathcal{P}$ with prob $\propto DP[\ell-1,j]$; prepend $j^\star$ to $\pi$; $cur\leftarrow j^\star$\;
  }
  \If{$|\pi|=L$}{add $\pi$ to $\Pi$}
}
\Return $\Pi$
\end{algorithm}

Given the set of $M$ segmented frames from the previous stage, the goal of Key-Frame Trajectory Synthesis (KFTS, see Algorithm~\ref{alg:kfts-wrap}) is to efficiently sample a large number of valid trajectories of a fixed length $L$. A trajectory is defined as a sequence of indices $\pi=(i_1, \ldots, i_L)$ such that $1 \le i_1 < i_2 < \cdots < i_L \le M$. The key constraint is that for any two consecutive frames $i_k$ and $i_{k+1}$ in the trajectory, the state change between them must be semantically meaningful and visually verifiable. The KFTS algorithm, detailed in Algorithm~\ref{alg:kfts-wrap}, accomplishes this efficiently by converting the problem into path sampling on a Directed Acyclic Graph (DAG) and using dynamic programming. The process consists of three main stages:

\begin{enumerate}[leftmargin=2em, after=\vspace{-0.5em}, itemsep=1pt]
    \item \textbf{Directed Acyclic Graph (DAG) Construction:} We first model the relationships between all segmented frames. The $M$ frames are treated as nodes in a graph. A directed edge exists from frame $i$ to frame $j$ (where $i < j$) if and only if the state difference $\delta(s_i, s_j)$ constitutes a valid, visible transition. This validity is determined by a predicate $\mathrm{Vis}(\cdot)$, which checks if the objects involved in the state change are clearly visible in both frames, as described in Section~\ref{sec:2_2}. This process results in an adjacency matrix $E$ for a DAG, where $E_{ij}=1$ indicates a valid one-step transition from frame $i$ to $j$.

    \item \textbf{Dynamic Programming Path Counting:} Instead of enumerating all possible $\binom{M}{L}$ combinations, we use dynamic programming (DP) to efficiently count the number of valid trajectories. We build a DP table where $DP[\ell, i]$ stores the total number of valid trajectories of length $\ell$ that terminate at frame $i$. The base case is $DP[1, i] = 1$ for all frames $i$, as any single frame is a valid path of length one. The table is filled using the recurrence:
    $$ DP[\ell, i] = \sum_{j<i} DP[\ell-1, j] \cdot E_{ji} $$
    This equation sums the number of valid paths of length $\ell-1$ ending at any valid predecessor $j$ of frame $i$. After filling the table up to length $L$, the entry $DP[L, i]$ gives the exact number of distinct, valid, length-$L$ trajectories that end at frame $i$.

    \item \textbf{Weighted Backtracking Sampling:} With the DP table computed, we can sample trajectories efficiently without bias. To generate one trajectory, we first sample an end-node $i_L$ from all possible frames $\{1, \dots, M\}$. The sampling is weighted, with the probability of selecting frame $i$ being proportional to its weight $w_i = DP[L, i]$. This ensures that frames that can be part of more trajectories are more likely to be chosen as endpoints.

    Once the end-node $i_L$ is selected, we reconstruct the path backwards. To select the previous node $i_{L-1}$, we consider all valid predecessors $j$ of $i_L$ (i.e., all $j < i_L$ where $E_{j, i_L}=1$). We sample the predecessor $j^\star$ with a probability proportional to $DP[L-1, j^\star]$. This process is repeated iteratively: to find node $i_k$, we sample from the predecessors of $i_{k+1}$ with probabilities proportional to the values in the $DP[k, \cdot]$ row. This weighted backtracking ensures that every valid trajectory of length $L$ has a chance of being sampled, and the likelihood of sampling any specific path is uniform across all valid paths. We repeat this procedure $R$ times to generate the desired number of trajectories.
\end{enumerate}

This DP-based approach is highly scalable as its complexity is polynomial in $M$ and $L$, making it far more efficient than a brute-force combinatorial search, especially when $M$ is large.

\textbf{Motivation of Scene Graph State-Action Spaces.}~ We opt for a symbolic scene graph representation of state and action, and while it may not capture the fine-grained details of low-level dynamics, this abstraction is advantageous for our objectives for two primary reasons. First, our focus is on detecting semantic changes within a scene, a task that naturally aligns with the semantic abstraction of VLMs rather than the continuous motor trajectories.
Second, the symbolic predicates we employ are grounded in practical robotics applications. This is demonstrated through our egocentric real-world experiments, confirming the real-world relevance of our chosen states. The feasibility of this symbolic approach is further substantiated by its use in guiding the data collection for the BEHAVIOR benchmark, where these predicates defined the goal conditions for simulated activities and enabled human annotators to clearly verify whether task states were successfully achieved.

\subsection{Dataset Statistics and Evaluation Design}

\begin{table}[htbp]
    \centering
    \caption{The 11 predicate classes used to define abstract state changes in our benchmark.}
    \begin{tabular}{lll}
        \toprule
        \multicolumn{3}{c}{\textbf{Predicate Classes}} \\
        \midrule
        \texttt{RightGrasping} & \texttt{LeftGrasping} & \texttt{OnTop}     \\
        \texttt{Inside}        & \texttt{Under}        & \texttt{Contains}  \\
        \texttt{Covered}       & \texttt{Open}         & \texttt{ToggledOn} \\
        \texttt{Cooked}        & \texttt{Transition}   &                    \\
        \bottomrule
    \end{tabular}
    \label{tab:predicates}
\end{table}

\begin{figure}[htbp]
    \centering
    \includegraphics[width=\linewidth]{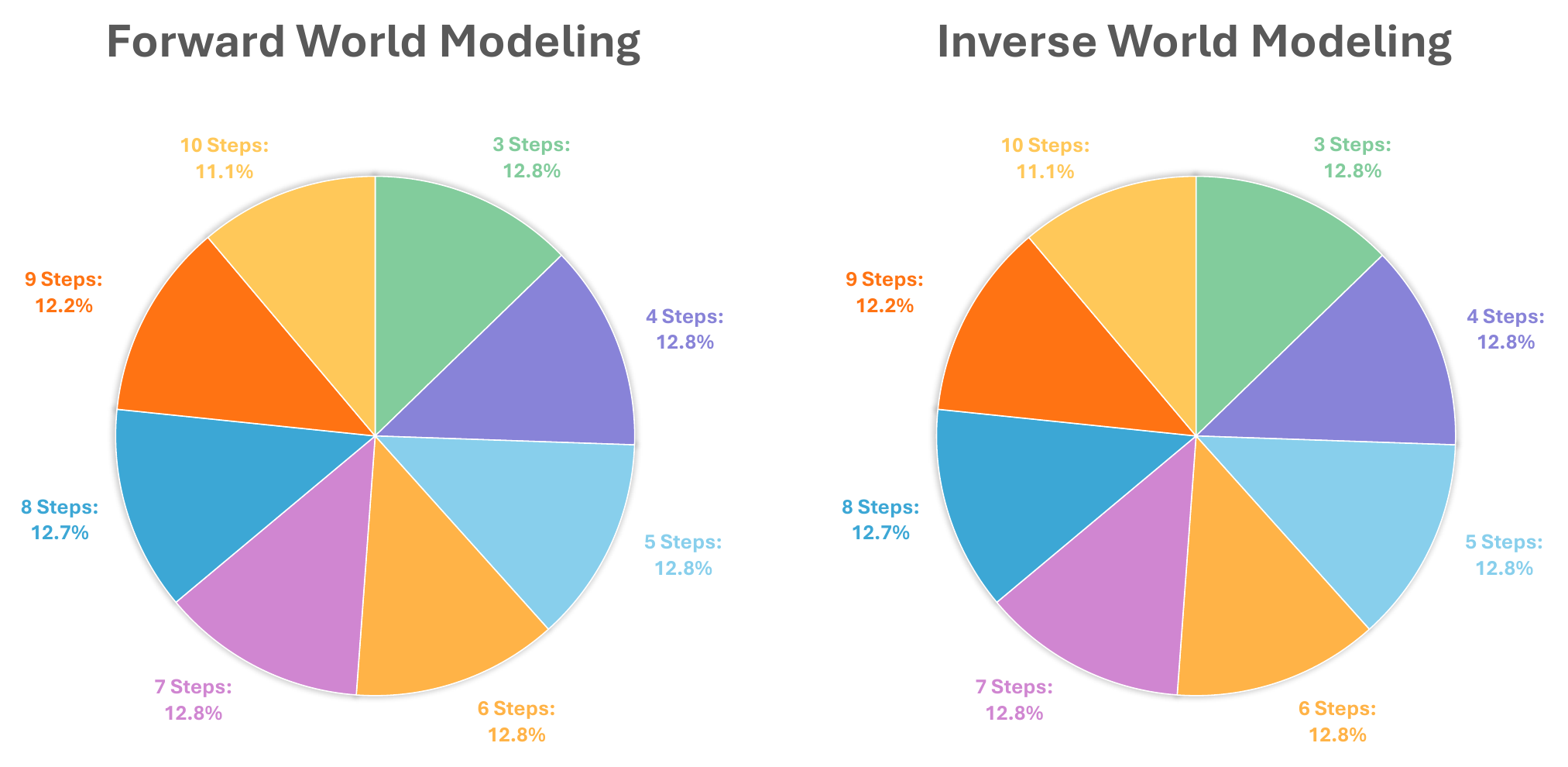}
    \caption{The distribution of problems by the number of steps in our ENACT benchmark dataset is shown for both forward (left) and inverse (right) world modeling tasks. The dataset is balanced, with a nearly uniform distribution of problems ranging from 3 to 10 steps.}
    \label{fig:benchmark_stats}
\end{figure}

\subsubsection{Dataset Statistics} \label{app_2_1:data_stats}
To ensure a comprehensive evaluation of models' reasoning capabilities across different time horizons, the sampled problems feature trajectory lengths varying from 3 to 10 steps. As illustrated in Figure~\ref{fig:benchmark_stats}, this dataset is intentionally balanced, featuring a near-uniform distribution of problems for each step length across both task types. This balance ensures that our evaluation is not biased towards shorter or longer-term reasoning.

The abstract state changes that define the actions in our benchmark are grounded in a set of 11 symbolic predicates. These predicates describe relationships between the agent and objects, as well as changes in object states. The complete list of predicates is detailed in Table~\ref{tab:predicates}.

\subsubsection{Evaluation Design}
\label{app_2_2:eval_metrics}

\textbf{From indices to dynamics.}
We grade \emph{what changes}, not just \emph{which index}. Each adjacent state pair yields an \emph{action signature}
$a^{\text{sig}}(s_{i-1},s_i)=\{c=(\gamma,e,\rho)\}$,
turning scene-graph deltas into compact semantics (operation $\gamma$ on entity $e$ and predicate $\rho$). For the reference sequence, we compute (i) the \emph{visible} subset $C_i$ and (ii) the \emph{full} set $F_i$. For a prediction, we compute $\tilde C_i$ (full diff). This uses state differences as the model’s proxy answer and avoids brittle numeric matching.

\paragraph{Online verifier.}
\emph{Forward dynamics.} After reconstructing the shuffled storyboard, we compare the ground-truth index sequence $\tau$ and the prediction $\sigma$.
Exact acceptance: $\sigma=\tau$.
Semantic acceptance (when lengths match): for all steps $i$,

$$
C_i \subseteq \tilde C_i.
$$

Intuition: the predicted step must \emph{cover} the reference’s visible change. The overall decision is \texttt{match} = (\texttt{exact} OR \texttt{semantic}); length-mismatched predictions are not accepted (but still get pairwise credit below).

\emph{Inverse dynamics.} The model orders actions. Exact acceptance: indices match. Semantic acceptance (equal length): for all $i$,

$$
\tilde C_i \subseteq F_i,
$$

i.e., the predicted action description can be a concise \emph{subset} of the full reference transition at that position. Again, \texttt{match} = (\texttt{exact} OR \texttt{semantic}).

\paragraph{Metrics.}
\textbf{Task accuracy (TA).} Score $1$ iff the verifier accepts the full prediction, else $0$; average over the split:

$$
\mathrm{TA}=\frac{1}{|\mathcal{D}|}\sum_{x\in\mathcal{D}}\mathbf{1}\{\text{accepted}(x)\}.
$$

\textbf{Pairwise accuracy (PA).} Measures stepwise consistency.
If lengths match,

$$
\mathrm{PA}(x)=\frac{1}{L}\sum_{i=1}^{L}\mathbf{1}\{\,C_i\subseteq \tilde C_i\ \text{(forward)}\ \text{or}\ \tilde C_i\subseteq F_i\ \text{(inverse)}\,\}.
$$

Accepted predictions have $\mathrm{PA}(x)=1$.
If lengths differ, we compute PA via a monotone alignment between reference and predicted steps that maximizes the number of subset-satisfying pairs (forward/inverse rule as above). We report the micro-average:

$$
\mathrm{PA}=\frac{\sum_x \#\text{correct pairs in }x}{\sum_x L_x}.
$$

\textbf{Summary.}
Multiple valid answers are allowed via the subset rules: forward requires $\text{reference-visible}\subseteq\text{predicted}$, inverse requires $\text{predicted}\subseteq\text{reference-full}$. TA captures all-or-nothing acceptance; PA gives graded credit for near-correct dynamics.

\subsubsection{\name Examples} \label{app_2_3:bench_examples}

\begin{figure}[!htbp]
    \centering
    \includegraphics[width=\linewidth]{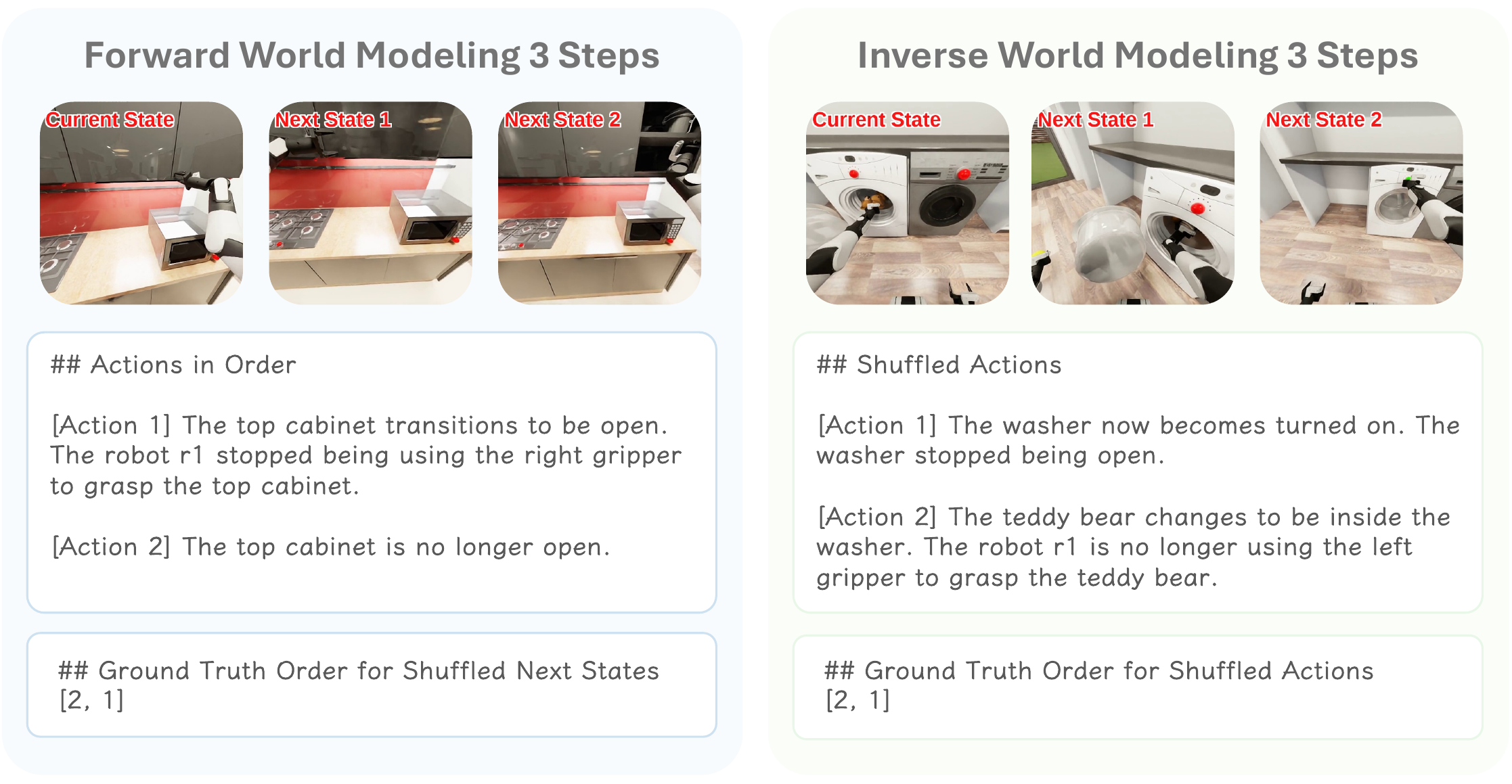}
    \caption{3-Step \textbf{Forward World Modeling} (left) and \textbf{Inverse World Modeling} (right) samples.}
    \label{fig:3_steps_sample}
\end{figure}

\begin{figure}[!htbp]
    \centering
    \includegraphics[width=\linewidth]{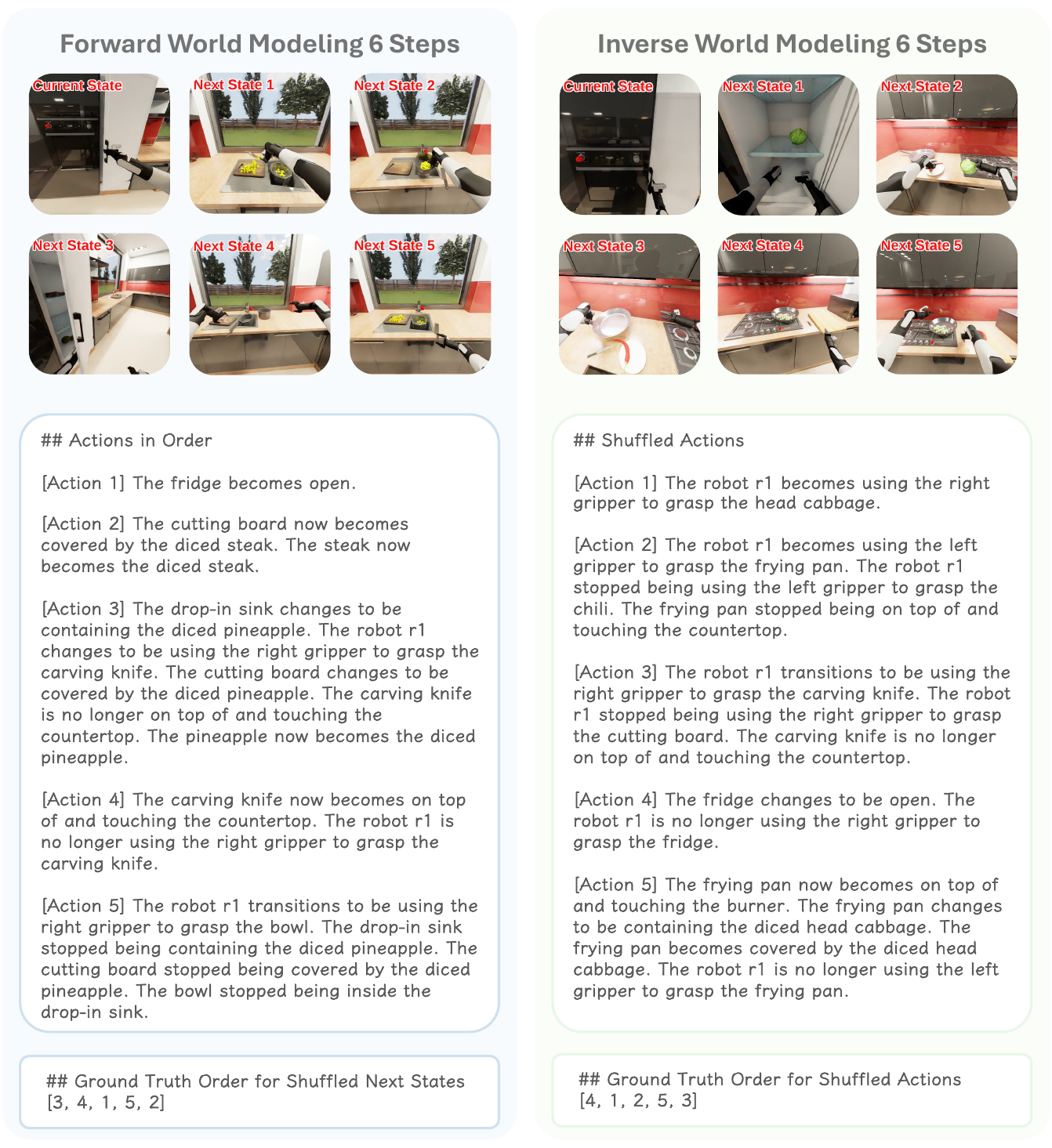}
    \caption{6-Step \textbf{Forward World Modeling} (left) and \textbf{Inverse World Modeling} (right) samples.}
    \label{fig:6_steps_sample}
\end{figure}

\begin{figure}[!htbp]
    \centering
    \includegraphics[width=\linewidth]{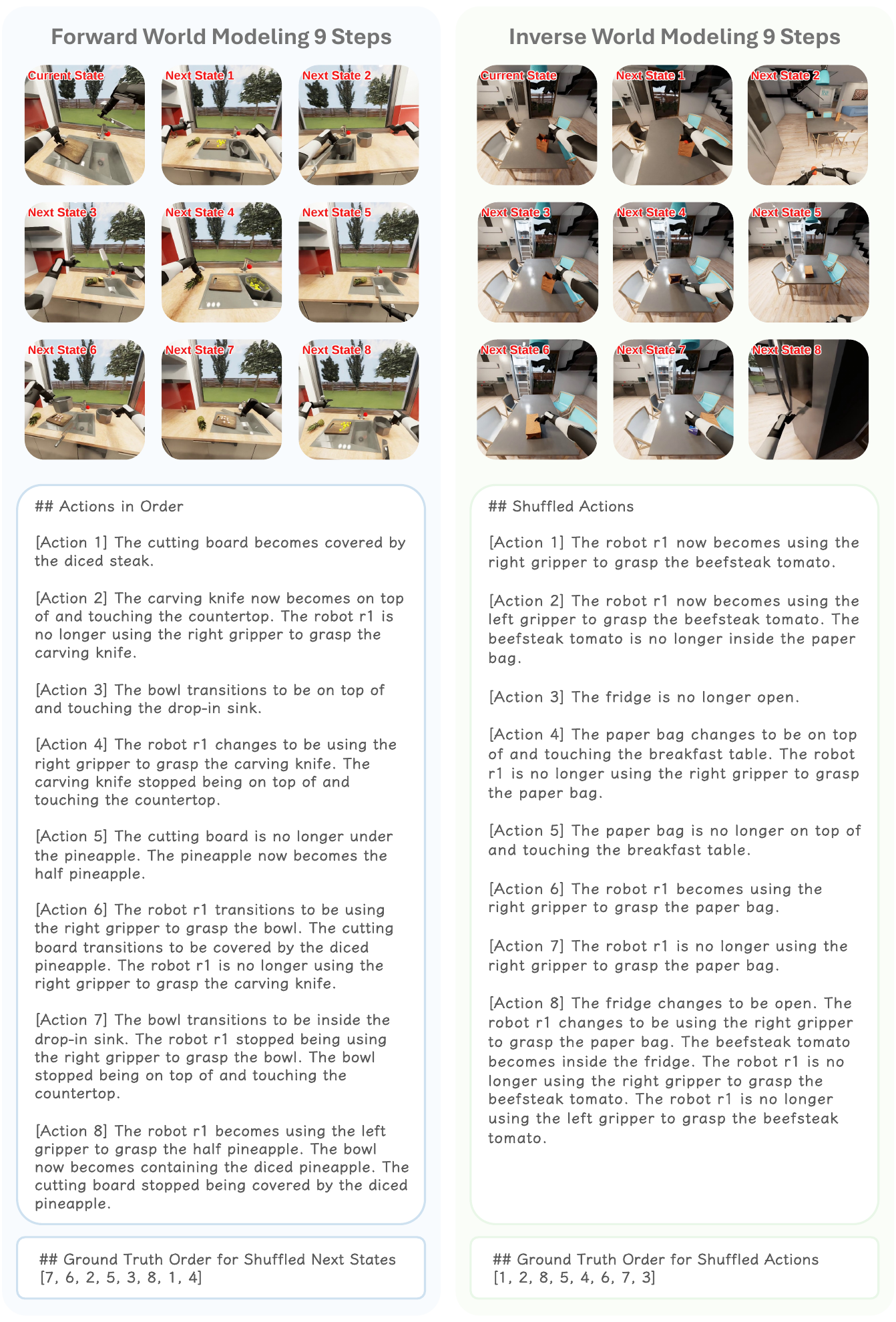}
    \caption{9-Step \textbf{Forward World Modeling} (left) and \textbf{Inverse World Modeling} (right) samples.}
    \label{fig:9_steps_sample}
\end{figure}

\section{Experiments and Analysis}
\label{app:02_experiments}
\subsection{Human Annotation}
\subsubsection{Annotation Interface \& Human Performance Evaluation}\label{app_3_1_1:human_setup}

\begin{figure}[htbp]
    \centering
    \includegraphics[width=\linewidth]{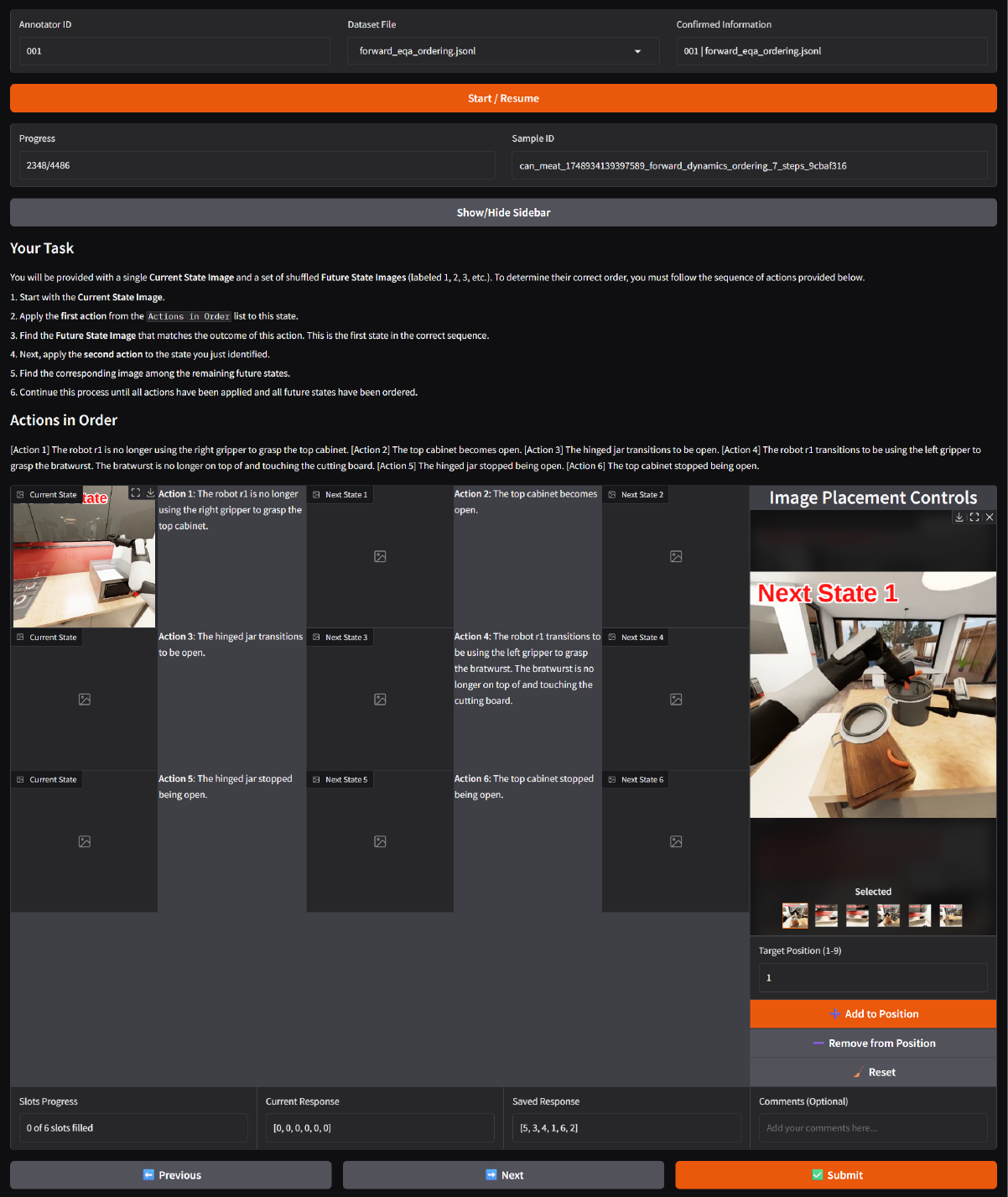}
    \caption{The annotation interface used for evaluating human performance on \textbf{Forward World Modeling} problems. Annotators are presented with a \textit{``Current State''} image (top left) and an ordered list of textual actions. The main task is to fill the \textit{``Next State''} slots by selecting the correct image from the shuffled \textit{Candidate Image Library} on the right. The annotator must follow the sequence of actions, using the result of the previous action as the starting point for the next, to determine the correct chronological order of all future states.}
    \label{fig:annotation_interface_fwd}
\end{figure}

\begin{figure}[htbp]
    \centering
    \includegraphics[width=\linewidth]{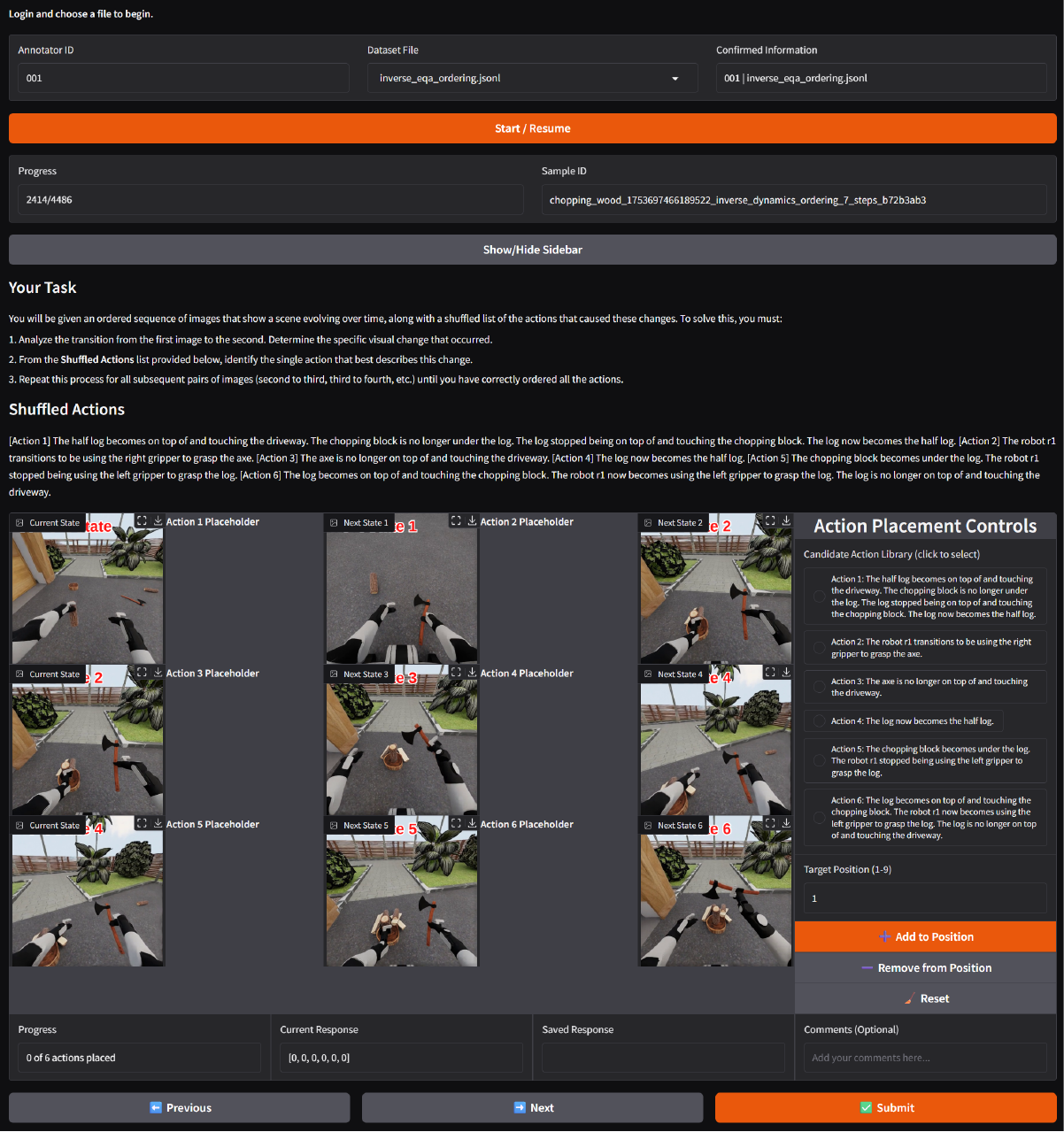}
    \caption{The annotation interface used for evaluating human performance on \textbf{Inverse World Modeling} problems. Annotators are shown an ordered sequence of state transitions, displayed as pairs of \textit{``Current State''} and \textit{``Next State''} images. For each transition, their task is to select the correct action description from a shuffled \textit{Candidate Action Library} that caused the visual change between the two states. }
    \label{fig:annotation_interface_invs}
\end{figure}

To establish an empirical upper bound on performance for the ENACT benchmark, we recruited three trained annotators to complete the same set of tasks assigned to the Vision-Language Models (VLMs). The annotators interacted with our customized human annotation interfaces implemented in Gradio, which are illustrated in Figure~\ref{fig:annotation_interface_fwd} for the \textbf{Forward World Modeling} tasks and in Figure~\ref{fig:annotation_interface_invs} for the \textbf{Inverse World Modeling} tasks. Importantly, annotators followed exactly the same instructions and task prompts as those provided to the VLMs, ensuring a fair and consistent comparison between human and model performance. This setup allows us to quantify the extent to which current VLMs approach human-level competence on the ENACT benchmark.

\subsubsection{Inter-Annotator Agreement Analysis}

To ensure the reliability of our human-generated labels, we conducted a rigorous Inter-Annotator Agreement (IAA) analysis. The initial dataset was annotated by three annotators working on disjoint, non-overlapping subsets, which precluded direct agreement measurement. We therefore implemented a systematic cross-annotation protocol. For each task type (forward and inverse) and for each of the eight step-length categories (from 3 to 10 steps), we randomly sampled five questions from each annotator's original assignment. This process created a balanced IAA evaluation set totaling 240 unique questions. Each of these sampled questions was then reassigned to the two annotators who had not performed the original annotation. For example, the five items sampled from Annotator A's work for a given category were re-annotated independently by Annotator B and Annotator C.

Following this protocol, we assessed the resulting annotations using \textbf{Krippendorff's Alpha ($\alpha$)}~\citep{krippendorff2011computing}, a robust statistical measure that is well-suited for this analysis as it accommodates multiple annotators and is resilient to missing data. Given that our annotation task involved ordering, we configured the analysis for ordinal data. The alpha coefficient is calculated based on the observed and expected disagreement among annotators, according to the formula:

$$\alpha = 1 - \frac{D_o}{D_e}$$

Here, $D_o$ represents the \textbf{observed disagreement}, which is calculated from the pairwise differences between all annotations for each item. $D_e$ is the \textbf{expected disagreement}, which represents the disagreement that would occur by chance, derived from the marginal distribution of all annotations. This balanced design ensured that every question in the IAA set received three independent labels, allowing for robust pairwise agreement calculation across all three pairs of annotators (A vs. B, A vs. C, and B vs. C) for each condition. To assess the stability of our $\alpha$ coefficient, we computed \textbf{95\% confidence intervals (CI)} using the bootstrap percentile method with 1,000 resamples of the 240 evaluation items. 

Our analysis yielded an overall Krippendorff's Alpha of \textbf{$\alpha = 0.8320$}, with a 95\% bootstrap confidence interval of \textbf{[0.7879, 0.8682]}. Given the established standard~\citep{krippendorff1999content, hayes2007answering, de2012calculating}, an alpha value above 0.80 indicates a high level of reliability. This strong result confirms that our annotation guidelines are clear and consistently applied by the annotators.

Pairwise agreement scores were also consistently high, further validating the reliability between individual annotators:
\begin{itemize}
    \item \textbf{Annotator A01 - B02}: $\alpha = 0.8180$
    \item \textbf{Annotator A01 - C03}: $\alpha = 0.8265$
    \item \textbf{Annotator B02 - C03}: $\alpha = 0.8518$
\end{itemize}

In addition to the chance-corrected alpha metric, we found that the annotators were in perfect agreement on 184 of the 240 selected questions, resulting in a \textbf{agreement rate of 76.67\%}. Collectively, these strong agreement metrics validate the reliability of our annotation process and the high quality of the resulting dataset.

\subsection{World Modeling as a Proxy for Evaluating Embodied Cognition}
\begin{table}[t]
\centering
\setlength{\tabcolsep}{4pt}
\small
\resizebox{\textwidth}{!}{
\begin{tabular}{lccccc}
\toprule
\textbf{Task} & \textbf{Action--Effect Reasoning} & \textbf{Causal Inference} & \textbf{Affordance Recognition} & \textbf{Embodied Awareness} & \textbf{Temporal Abstraction} \\
\midrule
Forward  & $\checkmark$ & $\times$ & $\checkmark$ & $\checkmark$ & $\checkmark$ \\
Inverse  & $\times$ & $\checkmark$ & $\checkmark$ & $\checkmark$ & $\checkmark$ \\
\bottomrule
\end{tabular}
}
\caption{
We posit that \name reflects embodied world modeling reasoning rather than simple temporal correlation.
Unlike passive video prediction, our formulation explicitly conditions state transitions on actions (Forward World Modeling) and infers actions from state changes (Inverse World Modeling), thereby evaluating VLMs as transition models.
Based on your suggestion, we include a table that relates ENACT's evaluation tasks to specific cognitive constructs.
}
\label{tab:cognitive_constructs}
\end{table}

\subsubsection{Experimental Setup}~\label{app_3_2_1:main_setup}

To ensure a fair and consistent comparison across all models, we employed a standardized evaluation protocol. For each task type (forward and inverse world modeling), a unified question prompt template was used. All input images were resized to a uniform resolution of $512 \times 512$ pixels before being passed to the models. To ensure deterministic and reproducible outputs, the decoding temperature for all models was set to 0. Models were instructed to return their answers as a parsable Python list representing the permutation of indices, as shown in Figure~\ref{app_fig:fwd_modeling_prompt} and Figure~\ref{app_fig:invs_modeling_prompt}. A comprehensive list of the specific models used in our evaluation is provided in Table~\ref{tab:model_details}. We deliberately choose one prompt template across all experiments because we follow the design choice mentioned in \citet{liang2022holistic}, that the models should adapt to users' input, instead of the reverse case.

\begin{table*}[ht]
\centering
\footnotesize
\resizebox{\textwidth}{!}{%
\begin{tabular}{lllcc}
\toprule
\textbf{Organization} & \textbf{Model Name} & \textbf{Release Date} & \textbf{Full Name} & \textbf{\begin{tabular}[c]{@{}c@{}}Evaluation Pipeline\end{tabular}} \\
\midrule
\rowcolor[HTML]{FFF2E6}\multicolumn{5}{c}{\emph{\textbf{Proprietary Models}}} \\
\midrule
\multirow{3}{*}{OpenAI} 
& GPT-5 & 2025-08 & \texttt{gpt-5-2025-08-07} & OpenAI API \\
& GPT-5-mini & 2025-08 & \texttt{gpt-5-mini-2025-08-07} & OpenAI API \\
& GPT-5-nano & 2025-08 & \texttt{gpt-5-nano-2025-08-07} & OpenAI API \\
\noalign{\vskip 0.5ex}\hdashline\noalign{\vskip 0.5ex}
\multirow{3}{*}{Google} 
& Gemini 2.5 Pro & 2025-06 & \texttt{gemini-2.5-pro} & Gemini API \\
& Gemini 2.5 Flash & 2025-06 & \texttt{gemini-2.5-flash} & Gemini API \\
& Gemini 2.5 Flash-Lite & 2025-06 & \texttt{gemini-2.5-flash-lite} & Gemini API \\
\noalign{\vskip 0.5ex}\hdashline\noalign{\vskip 0.5ex}
\multirow{1}{*}{Anthropic} & Claude Sonnet 4 & 2025-05 & \texttt{claude-sonnet-4-20250514} & Anthropic API \\
\midrule
\rowcolor[HTML]{FFFDED}\multicolumn{5}{c}{\emph{\textbf{Open-Weight Models}}} \\
\midrule
\multirow{2}{*}{Zhipu AI} 
& GLM-4.5V & 2025-08 & \texttt{GLM-4.5V} & Zhipu Foundation Model Open Platform API \\
& GLM-4.1V-Thinking & 2025-07 & \texttt{GLM-4.1V-Thinking-FlashX} & Zhipu Foundation Model Open Platform API\\
\noalign{\vskip 0.5ex}\hdashline\noalign{\vskip 0.5ex}
\multirow{2}{*}{Meta} 
& Llama-4-Scout-17B-16E-Ins & 2025-04 & \texttt{meta-llama/Llama-4-Scout-17B-16E-Instruct} & ModelScope API\\
& Llama-4-Mav-17B-128E-Ins & 2025-04 & \texttt{meta-llama/Llama-4-Mav-17B-128E-Instruct} & ModelScope API\\
\noalign{\vskip 0.5ex}\hdashline\noalign{\vskip 0.5ex}
\multirow{4}{*}{Shanghai AI Lab} 
& InternVL3.5-241B-A28B & 2025-08 & \texttt{OpenGVLab/InternVL3.5-241B-A28B} & Intern API \\
& InternVL3.5-14B & 2025-08 & \texttt{OpenGVLab/InternVL3.5-14B} & Hugging Face Transformers \\
& InternVL3.5-8B & 2025-08 & \texttt{OpenGVLab/InternVL3.5-8B} & Hugging Face Transformers\\
& InternVL3.5-4B & 2025-08 & \texttt{OpenGVLab/InternVL3.5-4B} & Hugging Face Transformers  \\
\noalign{\vskip 0.5ex}\hdashline\noalign{\vskip 0.5ex}
\multirow{3}{*}{Google} 
& Gemma-3-27b-it & 2025-03 & \texttt{google/gemma-3-27b-it} & Gemini API  \\
& Gemma-3-12b-it & 2025-03 & \texttt{google/gemma-3-12b-it} & Gemini API  \\
& Gemma-3-4b-it & 2025-03 & \texttt{google/gemma-3-4b-it} & Gemini API  \\
\noalign{\vskip 0.5ex}\hdashline\noalign{\vskip 0.5ex}
\multirow{5}{*}{Alibaba} 
& QVQ-72B-Preview & 2024-12 & \texttt{Qwen/QVQ-72B-Preview} & ModelScope API\\
& Qwen2.5-VL-72B-Ins & 2025-01 & \texttt{Qwen/Qwen2.5-VL-72B-Instruct} & ModelScope API\\
& Qwen2.5-VL-32B-Ins & 2025-01 & \texttt{Qwen/Qwen2.5-VL-32B-Instruct} & ModelScope API \\
& Qwen2.5-VL-7B-Ins & 2025-01 & \texttt{Qwen/Qwen2.5-VL-7B-Instruct} & Hugging Face Transformers  \\
& Qwen2.5-VL-3B-Ins & 2025-01 & \texttt{Qwen/Qwen2.5-VL-3B-Instruct} & Hugging Face Transformers  \\
\noalign{\vskip 0.5ex}\hdashline\noalign{\vskip 0.5ex}
\multirow{2}{*}{AIDC} 
& Ovis2.5-9B & 2025-08 & \texttt{AIDC-AI/Ovis2.5-9B} & Hugging Face Transformers \\
& Ovis2.5-2B & 2025-08 & \texttt{AIDC-AI/Ovis2.5-2B} & Hugging Face Transformers  \\
\noalign{\vskip 0.5ex}\hdashline\noalign{\vskip 0.5ex}
\multirow{2}{*}{OpenBMB} 
& MiniCPM-V-4.5 & 2025-08 & \texttt{openbmb/MiniCPM-V-4.5} & Hugging Face Transformers \\
& MiniCPM-o-2.6 & 2025-01 & \texttt{openbmb/MiniCPM-o-2.6} & Hugging Face Transformers \\
\noalign{\vskip 0.5ex}\hdashline\noalign{\vskip 0.5ex}
\multirow{1}{*}{Hugging Face} 
& Idefics3-8B-Llama3 & 2024-08 & \texttt{HuggingFaceM4/Idefics3-8B-Llama3} & Hugging Face Transformers\\
\noalign{\vskip 0.5ex}\hdashline\noalign{\vskip 0.5ex}
\multirow{1}{*}{Nvidia} 
& Cosmos-Reason1 & 2025-05 & \texttt{nvidia/Cosmos-Reason1} & Hugging Face Transformers\\
\bottomrule
\end{tabular}
}
\caption{Details of Vision Language Models (VLMs) assessed in this study.}
\label{tab:model_details}
\end{table*}

\begin{tcolorbox}[colback=black!5!white, colframe=black!75!white,
  title=Forward World Modeling Prompt,
  boxrule=0.5mm, width=\textwidth, arc=2mm, auto outer arc=true, breakable,
  ]
\scriptsize
\begin{verbatim}
You are a capable agent designed to infer multi-step forward dynamics transitions in 
embodied decision-making. Your goal is to predict the correct sequence of future 
states that result from applying a given series of actions to an initial state.

## Your Task
You will be provided with a single **Current State Image** and a set of shuffled 
**Future State Images** (labeled 1, 2, 3, etc.). To determine their correct order, 
you must follow the sequence of actions provided below.

1.  Start with the **Current State Image**.
2.  Apply the **first action** from the `Actions in Order` list to this state.
3.  Find the **Future State Image** that matches the outcome of this action. 
    This is the first state in the correct sequence.
4.  Next, apply the **second action** to the state you just identified.
5.  Find the corresponding image among the remaining future states.
6.  Continue this process until all actions have been applied and all future states 
    have been ordered. 

## Output Format
Your response **must be only** a Python list of integers representing the correct 
chronological order of the future state image labels. Do not include any other text, 
reasoning, or explanation.

**Example:** 
If you determine the correct sequence is 
'Next State 1' -> 'Next State 3' -> 'Next State 2', 
Your output must be: `[1, 3, 2]`

## Actions in Order
{STATE_CHANGES}

Now, please provide your answer in the requested format.
\end{verbatim}
\end{tcolorbox}
\captionof{figure}{The prompt used to evaluate VLMs on the multi-step \textbf{Forward World Modeling} task. The model must order shuffled future state images by reasoning over a given action sequence.}
\label{app_fig:fwd_modeling_prompt}

\begin{tcolorbox}[colback=black!5!white, colframe=black!75!white,
  title=Inverse World Modeling Prompt,
  boxrule=0.5mm, width=\textwidth, arc=2mm, auto outer arc=true, breakable]
\scriptsize
\begin{verbatim}
You are a capable agent designed to infer multi-step inverse dynamics transitions in 
embodied decision-making. Your goal is to determine the correct chronological order 
of actions that caused the state transitions shown in a sequence of images.

## Your Task
You will be given an ordered sequence of images that show a scene evolving over time, 
along with a shuffled list of the actions that caused these changes. 

To solve this, you must:
1.  Analyze the transition from the first image to the second. Determine the specific 
    visual change that occurred.
2.  From the **Shuffled Actions** list provided below, identify the single action that 
    best describes this change.
3.  Repeat this process for all subsequent pairs of images (second to third, third to 
    fourth, etc.) until you have correctly ordered all the actions.

## Output Format
Your response **must be only** a Python list of integers representing 
the correct order of the action labels. 
Do not include any other text, reasoning, explanations, or code formatting.

**Example:** 
If the correct sequence is [Action 2] -> [Action 3] -> [Action 1], 
your output must be: `[2, 3, 1]`

## Shuffled Actions
{SHUFFLED_ACTIONS}

Now, please provide your answer in the requested format.
\end{verbatim}
\end{tcolorbox}
\captionof{figure}{The prompt used to evaluate VLMs on the multi-step \textbf{Inverse World Modeling} task. The model must order a set of shuffled actions by reasoning over an ordered sequence of state images.}
\label{app_fig:invs_modeling_prompt}
\subsubsection{Detailed Results}~\label{app_3_2_2:main_analysis}

A detailed examination of the full experimental results are presented in Table~\ref{tab:app_task_acc} (Task Accuracy) and Table~\ref{tab:app_pair_acc} (Pairwise Accuracy).

\begin{table*}[!htbp]
\centering
\setlength{\tabcolsep}{2pt}
\small
\resizebox{\textwidth}{!}{
\begin{tabular}{@{}p{3mm} l cccccccc cccccccc@{}}
\toprule
& \multirow{3}{*}{\textbf{Model}} & \multicolumn{8}{>{\columncolor{forwardbg}}c}{\textbf{Forward World Modeling}} & \multicolumn{8}{>{\columncolor{inversebg}}c}{\textbf{Inverse World Modeling}} \\
\addlinespace[0.5ex]
\cmidrule(l){3-10}\cmidrule(r){11-18}
\addlinespace[0.5ex]
& & \bf 3 & \bf 4 & \bf 5 & \bf 6 & \bf 7 & \bf 8 & \bf 9 & \bf 10 & \bf 3 & \bf 4 & \bf 5 & \bf 6 & \bf 7 & \bf 8 & \bf 9 & \bf 10 \\
\midrule
\multicolumn{18}{>{\columncolor{proprietary}}l}{\textit{\footnotesize Proprietary Models}}\\
& GPT-5 & 80.59 & \colorbox{firstgrey}{62.72} & \colorbox{secondgrey}{47.13} & \colorbox{firstgrey}{33.62} & \colorbox{firstgrey}{20.24} & \colorbox{firstgrey}{11.58} & \colorbox{firstgrey}{7.30} & \colorbox{firstgrey}{5.00} & \colorbox{secondgrey}{86.19} & \colorbox{secondgrey}{72.65} & \colorbox{firstgrey}{59.65} & \colorbox{firstgrey}{43.73} & \colorbox{firstgrey}{33.68} & \colorbox{firstgrey}{24.04} & \colorbox{firstgrey}{17.15} & \colorbox{secondgrey}{13.00} \\
& GPT-5 mini & \colorbox{firstgrey}{83.39} & \colorbox{firstgrey}{62.72} & 45.22 & \colorbox{secondgrey}{31.71} & \colorbox{secondgrey}{19.02} & 9.12 & \colorbox{secondgrey}{5.29} & 2.80 & 84.79 & 67.42 & 58.09 & 41.11 & 29.67 & 18.07 & 13.50 & 8.60 \\
& GPT-5 nano & 58.57 & 30.66 & 9.74 & 3.83 & 1.40 & 0.00 & 0.00 & 0.00 & 72.03 & 39.02 & 17.22 & 8.19 & 3.14 & 1.05 & 0.36 & 0.00 \\
& Gemini 2.5 Pro & \colorbox{secondgrey}{81.99} & \colorbox{firstgrey}{62.72} & \colorbox{firstgrey}{47.30} & 29.79 & 17.80 & \colorbox{secondgrey}{10.00} & 3.28 & \colorbox{secondgrey}{3.60} & \colorbox{firstgrey}{87.76} & \colorbox{firstgrey}{73.52} & \colorbox{secondgrey}{58.61} & \colorbox{secondgrey}{43.38} & \colorbox{secondgrey}{33.51} & \colorbox{secondgrey}{23.68} & \colorbox{secondgrey}{15.88} & \colorbox{firstgrey}{14.40} \\
& Gemini 2.5 Flash & 75.52 & 50.52 & 25.22 & 14.29 & 6.28 & 2.98 & 1.28 & 0.20 & 82.52 & 61.15 & 38.96 & 27.70 & 17.98 & 13.86 & 6.20 & 3.80 \\
& Gemini 2.5 Flash-Lite & 52.97 & 27.18 & 10.09 & 3.83 & 1.40 & 0.18 & 0.18 & 0.00 & 69.06 & 42.33 & 19.83 & 8.54 & 4.71 & 0.88 & 0.73 & 0.00 \\
& Claude Sonnet 4 & 56.29 & 24.91 & 8.52 & 2.96 & 0.70 & 0.00 & 0.00 & 0.00 & 72.73 & 42.16 & 24.17 & 13.59 & 6.98 & 2.63 & 1.46 & 1.00 \\
\midrule
\multicolumn{18}{>{\columncolor{openweight}}l}{\textit{\footnotesize Open-Weight Models}}\\
& GLM-4.5V & 66.08 & \colorbox{secondgrey}{40.77} & 18.09 & \colorbox{secondgrey}{8.54} & 1.57 & 0.35 & \colorbox{secondgrey}{0.18} & 0.00 & \colorbox{secondgrey}{79.55} & \colorbox{secondgrey}{57.32} & \colorbox{secondgrey}{32.52} & \colorbox{secondgrey}{20.38} & \colorbox{secondgrey}{11.69} & \colorbox{secondgrey}{5.44} & \colorbox{secondgrey}{1.64} & 0.40 \\
& GLM-4.1V-Thinking & 57.52 & 28.40 & 11.30 & 2.26 & 0.35 & 0.18 & 0.00 & 0.00 & 73.43 & 39.37 & 12.00 & 4.53 & 0.87 & 0.53 & 0.00 & 0.00 \\
& Llama-4-Scout-17B-16E-Ins & 58.74 & 21.43 & 5.04 & 1.74 & 0.70 & 0.18 & 0.00 & 0.00 & 64.34 & 34.32 & 10.26 & 2.96 & 1.75 & 0.00 & 0.18 & 0.00 \\
& Llama-4-Mav-17B-128E-Ins & 63.99 & 32.58 & 14.78 & 4.36 & 1.57 & 0.35 & 0.00 & 0.00 & 71.50 & 49.30 & 24.35 & 11.85 & 4.19 & 1.58 & 0.55 & 0.00 \\
& InternVL3.5-241B-A28B & \colorbox{secondgrey}{67.83} & \colorbox{firstgrey}{43.38} & \colorbox{firstgrey}{21.22} & \colorbox{firstgrey}{12.02} & \colorbox{firstgrey}{4.71} & \colorbox{secondgrey}{1.05} & \colorbox{firstgrey}{0.36} & 0.00 & \colorbox{firstgrey}{81.99} & \colorbox{firstgrey}{59.76} & \colorbox{firstgrey}{40.35} & \colorbox{firstgrey}{24.22} & \colorbox{firstgrey}{15.18} & \colorbox{firstgrey}{7.37} & \colorbox{firstgrey}{4.56} & \colorbox{firstgrey}{2.00} \\
& InternVL3.5-14B & 46.33 & 14.81 & 3.48 & 1.05 & 0.00 & 0.00 & 0.00 & 0.00 & 66.43 & 45.12 & 23.65 & 11.85 & 5.93 & 1.93 & 1.28 & 0.40 \\
& InternVL3.5-8B & 54.72 & 25.09 & 5.39 & 1.05 & 1.22 & 0.18 & 0.00 & 0.00 & 63.99 & 40.24 & 20.00 & 6.79 & 3.49 & 0.53 & 0.36 & 0.20 \\
& InternVL3.5-4B & 54.55 & 22.13 & 6.43 & 2.09 & 0.52 & 0.00 & 0.00 & 0.00 & 63.64 & 32.93 & 16.00 & 5.75 & 2.27 & 0.53 & 0.18 & 0.00 \\
& Gemma-3-27b-it & 53.15 & 22.82 & 5.57 & 0.87 & 0.17 & 0.18 & 0.00 & 0.00 & 63.46 & 31.88 & 14.61 & 5.05 & 1.57 & 0.35 & 0.00 & \colorbox{secondgrey}{0.60} \\
& Gemma-3-12b-it & 51.22 & 21.78 & 6.09 & 1.05 & 0.17 & 0.00 & 0.00 & 0.00 & 52.80 & 27.53 & 9.74 & 2.79 & 1.75 & 0.35 & 0.00 & 0.00 \\
& Gemma-3-4b-it & 52.80 & 20.56 & 1.57 & 0.17 & 0.70 & 0.00 & 0.00 & 0.00 & 52.45 & 18.12 & 3.83 & 1.92 & 0.17 & 0.00 & 0.00 & 0.00 \\
& QVQ-72B-Preview & 60.84 & 29.79 & 8.17 & 2.09 & 0.70 & 0.00 & 0.00 & 0.00 & 66.96 & 40.24 & 16.87 & 6.97 & 3.84 & 1.23 & 0.55 & 0.00 \\
& Qwen2.5-VL-72B-Ins & \colorbox{firstgrey}{71.68} & 40.42 & \colorbox{secondgrey}{18.96} & 7.84 & \colorbox{secondgrey}{3.32} & \colorbox{firstgrey}{1.23} & 0.00 & 0.00 & 75.87 & 53.48 & 29.74 & 17.77 & 11.52 & 4.74 & 1.46 & 0.40 \\
& Qwen2.5-VL-32B-Ins & 51.40 & 32.75 & 10.09 & 3.48 & 0.52 & 0.00 & 0.00 & 0.00 & 39.34 & 33.45 & 19.13 & 8.89 & 6.11 & 2.11 & 0.91 & 0.00 \\
& Qwen2.5-VL-7B-Ins & 22.73 & 23.17 & 5.39 & 0.52 & 0.17 & 0.00 & 0.00 & 0.00 & 70.10 & 41.11 & 16.52 & 5.23 & 1.05 & 0.00 & 0.00 & 0.00 \\
& Qwen2.5-VL-3B-Ins & 45.98 & 13.76 & 5.91 & 0.70 & 0.17 & 0.00 & 0.00 & 0.00 & 56.64 & 32.75 & 13.39 & 5.75 & 1.05 & 0.18 & 0.00 & 0.00 \\
& Ovis2.5-9B & 47.55 & 23.00 & 10.61 & 2.96 & 1.05 & 0.18 & 0.00 & 0.00 & 62.76 & 35.54 & 16.00 & 6.27 & 1.75 & 0.35 & 0.00 & 0.00 \\
& Ovis2.5-2B & 39.69 & 17.77 & 5.91 & 0.87 & 0.52 & 0.00 & 0.00 & 0.00 & 48.43 & 23.87 & 8.52 & 1.57 & 0.00 & 0.00 & 0.00 & 0.00 \\
& MiniCPM-V-4.5 & 48.43 & 19.16 & 8.35 & 1.92 & 0.52 & 0.18 & 0.00 & 0.00 & 68.01 & 37.98 & 22.09 & 9.41 & 3.66 & 1.75 & 0.18 & 0.20 \\
& MiniCPM-o-2.6 & 26.05 & 17.07 & 5.22 & 1.57 & 0.00 & 0.00 & 0.00 & 0.00 & 38.64 & 27.35 & 11.30 & 2.44 & 0.52 & 0.18 & 0.00 & 0.00 \\
& Idefics3-8B-Llama3 & 48.08 & 16.20 & 2.26 & 0.52 & 0.17 & 0.00 & 0.00 & 0.00 & 46.33 & 16.72 & 2.96 & 1.57 & 0.00 & 0.00 & 0.00 & 0.00 \\
& Cosmos-Reason1 & 45.45 & 21.43 & 5.04 & 0.52 & 0.17 & 0.00 & 0.00 & 0.00 & 51.92 & 29.09 & 12.02 & 3.31 & 0.52 & 0.18 & 0.00 & 0.00 \\
& BAGEL & 25.87 & 17.77 & 3.83 & 2.09 & 0.17 & 0.00 & 0.00 & 0.00 & 56.29 & 35.89 & 14.61 & 7.14 & 2.97 & 0.18 & 0.18 & 0.00 \\

\midrule
& \textbf{Human Performance} & \textbf{90.38} & \textbf{92.16} & \textbf{89.74} & \textbf{85.71} & \textbf{88.31} & \textbf{87.02} & \textbf{85.58} & \textbf{84.00} & \textbf{91.78} & \textbf{90.24} & \textbf{88.70} & \textbf{88.15} & \textbf{89.53} & \textbf{92.28} & \textbf{87.73} & \textbf{85.00}\\
\bottomrule
\end{tabular}
}
\caption{\textbf{Evaluation on \name (Task Accuracy).}
{\setlength{\fboxsep}{2pt}\protect\colorbox{gray!60}{\rule{0pt}{1.1ex}Dark gray}}
indicates the best result within each category (Proprietary or Open-Weight Models), and
{\setlength{\fboxsep}{2pt}\protect\colorbox{gray!20}{\rule{0pt}{1.1ex}Light gray}}
denotes the second-best result within the category.}
\label{tab:app_task_acc}
\end{table*}

\begin{table*}[!htbp]
\centering
\setlength{\tabcolsep}{2pt}
\small
\resizebox{\textwidth}{!}{
\begin{tabular}{@{}p{3mm} l cccccccc cccccccc@{}}
\toprule
& \multirow{3}{*}{\textbf{Model}} & \multicolumn{8}{>{\columncolor{forwardbg}}c}{\textbf{Forward World Modeling}} & \multicolumn{8}{>{\columncolor{inversebg}}c}{\textbf{Inverse World Modeling}} \\
\addlinespace[0.5ex]
\cmidrule(l){3-10}\cmidrule(r){11-18}
\addlinespace[0.5ex]
& & \bf 3 & \bf 4 & \bf 5 & \bf 6 & \bf 7 & \bf 8 & \bf 9 & \bf 10 & \bf 3 & \bf 4 & \bf 5 & \bf 6 & \bf 7 & \bf 8 & \bf 9 & \bf 10 \\
\midrule
\multicolumn{18}{>{\columncolor{proprietary}}l}{\textit{\footnotesize Proprietary Models}}\\
& GPT-5 & 84.62 & 75.26 & \colorbox{secondgrey}{69.96} & \colorbox{firstgrey}{64.18} & \colorbox{secondgrey}{57.48} & \colorbox{secondgrey}{52.16} & \colorbox{firstgrey}{49.45} & \colorbox{firstgrey}{46.93} & \colorbox{secondgrey}{86.28} & \colorbox{secondgrey}{80.37} & \colorbox{firstgrey}{76.09} & \colorbox{secondgrey}{68.78} & \colorbox{secondgrey}{65.71} & \colorbox{secondgrey}{62.13} & \colorbox{secondgrey}{57.12} & \colorbox{secondgrey}{55.33} \\
& GPT-5 mini & \colorbox{firstgrey}{87.50} & \colorbox{secondgrey}{76.25} & \colorbox{firstgrey}{70.65} & \colorbox{secondgrey}{63.41} & \colorbox{firstgrey}{58.14} & \colorbox{firstgrey}{52.38} & \colorbox{secondgrey}{46.65} & \colorbox{secondgrey}{44.11} & 85.05 & 76.77 & \colorbox{secondgrey}{75.43} & 67.67 & 63.79 & 57.04 & 55.04 & 50.02 \\
& GPT-5 nano & 67.83 & 50.29 & 38.61 & 30.35 & 25.97 & 21.90 & 17.59 & 16.84 & 72.81 & 53.95 & 42.48 & 36.45 & 31.68 & 28.20 & 24.11 & 20.33 \\
& Gemini 2.5 Pro & \colorbox{secondgrey}{86.10} & \colorbox{firstgrey}{76.42} & 69.83 & 60.80 & 53.26 & 48.12 & 40.12 & 36.98 & \colorbox{firstgrey}{87.94} & \colorbox{firstgrey}{81.18} & 75.39 & \colorbox{firstgrey}{70.03} & \colorbox{firstgrey}{66.03} & \colorbox{firstgrey}{62.91} & \colorbox{firstgrey}{57.78} & \colorbox{firstgrey}{56.62} \\
& Gemini 2.5 Flash & 81.64 & 67.94 & 54.17 & 43.38 & 37.43 & 32.73 & 29.88 & 28.07 & 82.78 & 72.18 & 60.83 & 58.19 & 53.14 & 51.78 & 47.99 & 44.98 \\
& Gemini 2.5 Flash-Lite & 64.34 & 49.07 & 38.70 & 33.87 & 27.81 & 25.44 & 23.31 & 20.31 & 69.58 & 57.55 & 46.04 & 39.09 & 34.06 & 30.18 & 27.51 & 23.16 \\
& Claude Sonnet 4 & 65.65 & 45.82 & 36.65 & 30.52 & 26.61 & 22.78 & 21.49 & 20.16 & 73.25 & 56.85 & 48.87 & 43.07 & 37.00 & 32.71 & 30.50 & 28.49 \\
\midrule
\multicolumn{18}{>{\columncolor{openweight}}l}{\textit{\footnotesize Open-Weight Models}}\\
& GLM-4.5V & 74.30 & 59.99 & 47.65 & 38.78 & 30.83 & 25.69 & 21.60 & 19.67 & \colorbox{secondgrey}{80.59} & \colorbox{secondgrey}{69.28} & \colorbox{secondgrey}{57.04} & \colorbox{secondgrey}{51.53} & \colorbox{secondgrey}{46.95} & \colorbox{secondgrey}{41.68} & \colorbox{secondgrey}{37.36} & \colorbox{secondgrey}{37.93} \\
& GLM-4.1V-Thinking & 67.31 & 49.48 & 38.43 & 31.29 & 25.80 & 21.50 & 20.14 & 18.73 & 75.35 & 56.27 & 46.57 & 36.79 & 29.61 & 24.56 & 23.91 & 25.80 \\
& Llama-4-Scout-17B-16E-Ins & 68.18 & 42.62 & 34.30 & 30.52 & 28.50 & 26.57 & 25.94 & \colorbox{firstgrey}{31.20} & 66.00 & 50.00 & 41.30 & 37.04 & 29.73 & 25.61 & 22.45 & 26.54 \\
& Llama-4-Mav-17B-128E-Ins & 72.47 & 52.09 & 43.87 & 35.30 & 29.90 & 25.89 & 22.79 & 20.49 & 72.55 & 62.60 & 50.52 & 43.10 & 35.17 & 31.68 & 28.10 & 25.80 \\
& InternVL3.5-241B-A28B & \colorbox{secondgrey}{75.79} & \colorbox{firstgrey}{62.25} & \colorbox{firstgrey}{50.83} & \colorbox{firstgrey}{45.85} & \colorbox{firstgrey}{37.84} & \colorbox{firstgrey}{32.88} & \colorbox{secondgrey}{27.85} & 25.24 & \colorbox{firstgrey}{82.26} & \colorbox{firstgrey}{70.09} & \colorbox{firstgrey}{60.61} & \colorbox{firstgrey}{53.38} & 45.90 & 39.35 & 34.12 & 30.56 \\
& InternVL3.5-14B & 54.90 & 36.53 & 27.87 & 25.47 & 22.02 & 18.73 & 18.29 & 20.60 & 69.06 & 59.52 & 49.00 & 43.45 & 37.61 & 32.28 & 29.31 & 28.58 \\
& InternVL3.5-8B & 64.42 & 44.83 & 31.48 & 24.32 & 23.62 & 21.50 & 19.30 & 15.47 & 65.03 & 56.10 & 45.35 & 37.67 & 35.02 & 29.62 & 26.41 & 23.60 \\
& InternVL3.5-4B & 63.11 & 42.04 & 30.26 & 26.13 & 21.73 & 20.28 & 19.64 & 21.98 & 64.95 & 50.12 & 41.61 & 35.78 & 29.00 & 26.57 & 27.55 & 24.04 \\
& Gemma-3-27b-it & 63.29 & 44.66 & 32.04 & 25.82 & 22.11 & 19.50 & 16.74 & 16.29 & 64.95 & 48.37 & 40.04 & 33.87 & 28.53 & 23.63 & 21.74 & 19.36 \\
& Gemma-3-12b-it & 62.33 & 43.55 & 32.78 & 25.68 & 22.45 & 20.40 & 17.70 & 16.71 & 53.23 & 43.79 & 34.43 & 29.90 & 25.57 & 22.31 & 21.60 & 18.16 \\
& Gemma-3-4b-it & 61.98 & 41.17 & 35.70 & 35.16 & 30.51 & 26.17 & 26.73 & 25.80 & 53.06 & 36.41 & 29.52 & 26.38 & 22.66 & 24.44 & 33.71 & 33.62 \\
& QVQ-72B-Preview & 69.14 & 52.96 & 40.83 & 36.27 & 33.16 & 30.63 & 26.30 & 24.76 & 71.33 & 58.77 & 48.43 & 44.36 & 40.26 & 39.30 & 36.66 & 36.58 \\
& Qwen2.5-VL-72B-Ins & \colorbox{firstgrey}{78.15} & \colorbox{secondgrey}{60.05} & \colorbox{secondgrey}{49.87} & \colorbox{secondgrey}{41.92} & \colorbox{secondgrey}{36.77} & \colorbox{secondgrey}{31.73} & \colorbox{firstgrey}{28.03} & 25.07 & 77.80 & 65.85 & 53.30 & 48.19 & 44.07 & 37.57 & 33.76 & 36.27 \\
& Qwen2.5-VL-32B-Ins & 67.83 & 55.46 & 44.35 & 35.75 & 27.52 & 26.42 & 22.01 & 18.07 & 63.55 & 59.70 & 54.57 & 51.01 & \colorbox{firstgrey}{49.36} & \colorbox{firstgrey}{47.17} & \colorbox{firstgrey}{41.47} & \colorbox{firstgrey}{40.16} \\
& Qwen2.5-VL-7B-Ins & 26.84 & 43.90 & 32.00 & 23.07 & 19.66 & 16.69 & 11.82 & 11.31 & 70.54 & 56.45 & 42.43 & 32.89 & 25.07 & 19.52 & 16.72 & 17.42 \\
& Qwen2.5-VL-3B-Ins & 58.22 & 35.31 & 30.57 & 24.08 & 20.36 & 17.44 & 14.87 & 15.07 & 57.43 & 49.13 & 40.48 & 34.88 & 28.33 & 26.14 & 22.97 & 20.51 \\
& Ovis2.5-9B & 58.39 & 42.51 & 34.96 & 31.08 & 24.61 & 20.78 & 18.11 & 16.96 & 64.86 & 51.74 & 41.65 & 35.47 & 30.95 & 26.64 & 23.70 & 23.25 \\
& Ovis2.5-2B & 46.94 & 38.85 & 32.65 & 26.86 & 25.63 & 22.21 & 22.49 & 24.87 & 54.28 & 44.08 & 35.43 & 29.06 & 27.84 & 25.56 & 27.62 & 29.29 \\
& MiniCPM-V-4.5 & 60.75 & 38.73 & 33.65 & 25.47 & 24.81 & 21.40 & 21.56 & 18.33 & 69.23 & 53.08 & 47.35 & 39.55 & 34.87 & 30.63 & 27.05 & 25.71 \\
& MiniCPM-o-2.6 & 35.31 & 39.37 & 29.48 & 31.78 & 27.66 & 26.39 & 24.59 & \colorbox{secondgrey}{27.42} & 54.11 & 48.26 & 44.70 & 40.00 & 38.28 & 36.12 & 33.23 & 31.71 \\
& Idefics3-8B-Llama3 & 60.23 & 36.99 & 31.83 & 24.25 & 21.29 & 20.80 & 20.46 & 17.71 & 47.38 & 33.86 & 27.26 & 23.48 & 19.87 & 18.50 & 17.04 & 15.16 \\
& Cosmos-Reason1 & 56.28 & 41.86 & 34.75 & 28.40 & 26.46 & 26.49 & 25.41 & 24.88 & 58.30 & 45.93 & 44.25 & 38.50 & 35.72 & 34.56 & 31.50 & 28.64 \\
& BAGEL & 30.24 & 40.19 & 29.65 & 25.37 & 22.75 & 19.45 & 17.84 & 15.87 & 56.73 & 52.85 & 40.09 & 35.44 & 29.67 & 24.39 & 28.70 & 18.91 \\
\midrule
& \textbf{Human Performance} & \textbf{93.62} & \textbf{95.30} & \textbf{95.04} & \textbf{93.87} & \textbf{95.43} & \textbf{95.41} & \textbf{94.75} & \textbf{95.13} & \textbf{92.05} & \textbf{93.56} & \textbf{94.35} & \textbf{94.25} & \textbf{95.96} & \textbf{97.74} & \textbf{96.30} & \textbf{96.29} \\
\bottomrule
\end{tabular}
}
\caption{\textbf{Evaluation on \name (Pairwise Accuracy).}
{\setlength{\fboxsep}{2pt}\protect\colorbox{gray!60}{\rule{0pt}{1.1ex}Dark gray}}
indicates the best result within each category (Proprietary or Open-Weight Models), and
{\setlength{\fboxsep}{2pt}\protect\colorbox{gray!20}{\rule{0pt}{1.1ex}Light gray}}
denotes the second-best result within the category.}
\label{tab:app_pair_acc}
\end{table*}

\subsubsection{Contact Experiment}
\begin{table*}[!htbp]
\centering
\setlength{\tabcolsep}{2pt}
\small
\resizebox{\textwidth}{!}{
\begin{tabular}{@{}p{3mm} l cccccccc cccccccc@{}}
\toprule
& \multirow{3}{*}{\textbf{Metric}} &
\multicolumn{8}{>{\columncolor{forwardbg}}c}{\textbf{Forward (with contact changes)}} &
\multicolumn{8}{>{\columncolor{inversebg}}c}{\textbf{Inverse (with contact changes)}} \\
\addlinespace[0.5ex]
\cmidrule(l){3-10}\cmidrule(r){11-18}
\addlinespace[0.5ex]
& & \bf 3 & \bf 4 & \bf 5 & \bf 6 & \bf 7 & \bf 8 & \bf 9 & \bf 10
  & \bf 3 & \bf 4 & \bf 5 & \bf 6 & \bf 7 & \bf 8 & \bf 9 & \bf 10 \\
\midrule
& Task Accuracy
& 86.67 & 43.33 & 36.67 & 20.00 &  3.45 &  3.33 & 0.00 & 0.00
& 90.00 & 73.33 & 30.00 & 26.67 & 16.67 &  3.33 & 6.67 & 0.00 \\
& Pairwise Accuracy
& 90.00 & 72.22 & 60.00 & 53.33 & 48.28 & 42.38 & 31.67 & 34.44
& 90.00 & 82.22 & 55.83 & 57.33 & 46.11 & 32.86 & 38.75 & 27.78 \\
\bottomrule
\end{tabular}
}
\caption{\textbf{Effect of including contact changes in key-frame selection.}
Task and pairwise accuracies (\%) of InternVL3.5--241B when key frames are triggered by both state changes and contact changes.}
\label{tab:contact_ablation}
\end{table*}

To verify that our conclusions are not an artifact of using only ``semantic scene graph'' predicates in ENACT, we add an ablation in which key frames are also gated on changes in binary contact relations between objects (e.g., touch / no-touch). Concretely, we augment the symbolic predicate set so that both state changes and contact changes trigger key-frame sampling, while keeping the rest of the pipeline unchanged, and re-evaluate InternVL3.5--241B on the resulting trajectories.

As shown in Table~\ref{tab:contact_ablation}, the qualitative trends remain the same as in our main results. Inverse world modeling consistently outperforms forward modeling across all horizons (e.g., $86.67\%$ vs.\ $90.00\%$ task accuracy at 3 steps, and $3.45\%$ vs.\ $16.67\%$ at 7 steps), and both task and pairwise accuracies still drop substantially as the number of interaction steps increases for both directions. This suggests that our findings are robust to the choice of symbolic key-frame criteria and are not driven by sparsity introduced by state-change-only sampling.

\subsubsection{How Does Action Representation Affect VLMs' Performance?}
\definecolor{emojirow}{RGB}{245,245,255}
\begin{table*}[!htbp]
\centering
\setlength{\tabcolsep}{2pt}
\small
\resizebox{\textwidth}{!}{
\begin{tabular}{@{}p{3mm} l cccccccc cccccccc@{}}
\toprule
& \multirow{3}{*}{\textbf{Metric}} &
\multicolumn{8}{>{\columncolor{forwardbg}}c}{\textbf{Forward}} &
\multicolumn{8}{>{\columncolor{inversebg}}c}{\textbf{Inverse}} \\
\addlinespace[0.5ex]
\cmidrule(l){3-10}\cmidrule(r){11-18}
\addlinespace[0.5ex]
& & \bf 3 & \bf 4 & \bf 5 & \bf 6 & \bf 7 & \bf 8 & \bf 9 & \bf 10
  & \bf 3 & \bf 4 & \bf 5 & \bf 6 & \bf 7 & \bf 8 & \bf 9 & \bf 10 \\
\midrule
\multicolumn{18}{>{\columncolor{proprietary}}l}{\textit{\footnotesize Vanilla (Natural Language)}}\\
& Task Accuracy
& 68.97 & 35.17 & 27.59 &  8.97 &  6.90 &  2.07 & 0.69 & 0.00
& 83.45 & 60.69 & 44.14 & 24.83 & 13.79 &  7.59 & 4.14 & 0.00 \\
& Pairwise Accuracy
& 76.21 & 57.70 & 57.41 & 41.24 & 38.85 & 30.34 & 29.05 & 26.52
& 83.45 & 69.66 & 61.72 & 53.10 & 47.01 & 40.39 & 34.40 & 27.25 \\
\midrule
\multicolumn{18}{>{\columncolor{openweight}}l}{\textit{\footnotesize Symbolic Predicates}}\\
& Task Accuracy
& 67.59 & 40.00 & 20.00 &  9.66 &  4.14 & 0.69 & 1.38 & 0.00
& 79.86 & 51.03 & 40.00 & 23.45 &  8.97 & 5.52 & 2.76 & 0.73 \\
& Pairwise Accuracy
& 74.48 & 61.61 & 48.97 & 41.52 & 35.75 & 32.12 & 30.34 & 25.68
& 79.86 & 62.99 & 58.62 & 51.59 & 39.43 & 36.85 & 32.07 & 24.33 \\
\midrule
\multicolumn{18}{>{\columncolor{emojirow}}l}{\textit{\footnotesize Emoji-Style Encodings}}\\
& Task Accuracy
& 65.52 & 44.14 & 18.62 & 11.03 &  8.28 & 0.69 & 0.69 & 0.00
& 77.24 & 48.28 & 35.17 & 23.45 & 12.41 & 7.59 & 2.76 & 0.00 \\
& Pairwise Accuracy
& 73.10 & 64.60 & 48.62 & 39.45 & 38.05 & 27.49 & 26.72 & 23.93
& 77.93 & 61.38 & 59.14 & 45.38 & 40.80 & 36.06 & 27.16 & 25.30 \\
\bottomrule
\end{tabular}
}
\caption{\textbf{Effect of action representation on InternVL3.5--241B.}
Task and pairwise accuracies (\%) on a 2{,}304-QA subset of \name under three action encodings.}
\label{tab:app_action_repr}
\end{table*}

Our primary goal is to evaluate VLMs under the standard interface of natural-language actions, but this leaves open whether the inverse advantage is merely a consequence of language priors, i.e., models being better at mapping visuals to familiar verbs than to unfamiliar symbolic actions.
To test this, we construct a subset of ENACT trajectories with 2{,}304 QAs and compare three action--predicate encodings while keeping the underlying videos and questions fixed:
(1) the original natural-language descriptions (``vanilla''), 
(2) structured symbolic predicates, and 
(3) emoji-style encodings.
Table~\ref{tab:app_action_repr} reports task and pairwise accuracies of InternVL3.5--241B across horizons for all three settings.

Across all representations, we observe the same qualitative pattern as in our main results:
Inverse world modeling consistently outperforms forward modeling at comparable horizons, and performance for both directions degrades sharply as the number of interaction steps increases.
While absolute accuracies vary slightly across encodings, the inverse~$>$~forward gap is preserved even with purely symbolic or emoji-style actions, suggesting that our conclusions are not driven solely by natural-language priors.

\subsubsection{How Often Do Accepted Predictions Omit Parts of Transitions?}
\begin{table}[t]
\centering
\setlength{\tabcolsep}{8pt}
\small
\begin{tabular}{lcc}
\toprule
\textbf{Model} & \textbf{Data-Level Mismatch (\%)} & \textbf{Pair-Level Mismatch (\%)} \\
\midrule
GPT-5            & 1.65 & 15.31 \\
Human            & 2.26 &  3.62 \\
InternVL3.5-241B & 1.49 & 13.96 \\
\bottomrule
\end{tabular}
\caption{\textbf{Mismatch rates among semantically accepted predictions.}
Data-level and pair-level mismatch ratios (\%) for cases where the semantic verifier accepts a prediction but its predicate set is a strict subset of the ground-truth transition.}
\label{tab:omit_transitions}
\end{table}

Our semantic verifier operates with subset inclusion: a predicted transition is accepted as semantically correct if its predicate set is a subset of the ground-truth transition and does not contain any predicates that contradict the ground truth. In other words, we allow partial correctness (omitting some true predicates), but never accept hallucinated predicates that conflict with the annotated transition.

To quantify how often such omissions occur among \emph{accepted} predictions, we measure mismatch ratios at two granularities:
(i) a \textbf{data-level} mismatch ratio, computed per QA as the fraction of semantically accepted predictions whose predicate set is a strict subset of the ground truth; and
(ii) a \textbf{pair-level} mismatch ratio, computed per ordered pair in the reordering task as the fraction of accepted pairs where at least one element is a strict subset of the corresponding ground truth transition.
Results for GPT-5, InternVL3.5--241B, and human annotators are shown in Table~\ref{tab:omit_transitions}.

At the data level, all models (and humans) exhibit very low mismatch rates ($\approx$1--2\%), indicating that most semantically accepted predictions recover the full transition.
At the pair level, models show higher mismatch ratios than humans (around 14--15\% vs.\ 3.6\%), reflecting that they occasionally capture only a subset of the true transition when comparing two candidate steps.

Overall, this suggests that while our verifier does grant some partial credit, such cases are relatively rare at the QA level and do not dominate the evaluation.

\subsection{Additional Ablation Experiments Common Setup}
\label{app_3_4:common_setup}

To gain deeper insights into model sensitivities, we conducted a series of controlled ablation experiments. This section outlines the common experimental framework that applies to our analyses of Image Realism (Section~\ref{app_3_3:image_realism}), Camera Configurations (Section~\ref{app_3_4:camera_config}), and Robot Appearance (Section~\ref{app_3_5_1:robot_appearance}).

For these experiments, we selected two representative models. Given its strong balance of performance and computational cost in our main results, we chose \textbf{GPT-5 mini} as our primary model to represent state-of-the-art proprietary VLMs. To include a strong open-weight counterpart, we also selected \textbf{InternVL3.5-241B-A28B}, which demonstrated robust performance among open models.

\begin{figure}[!htbp]
    \centering
    \includegraphics[width=\linewidth]{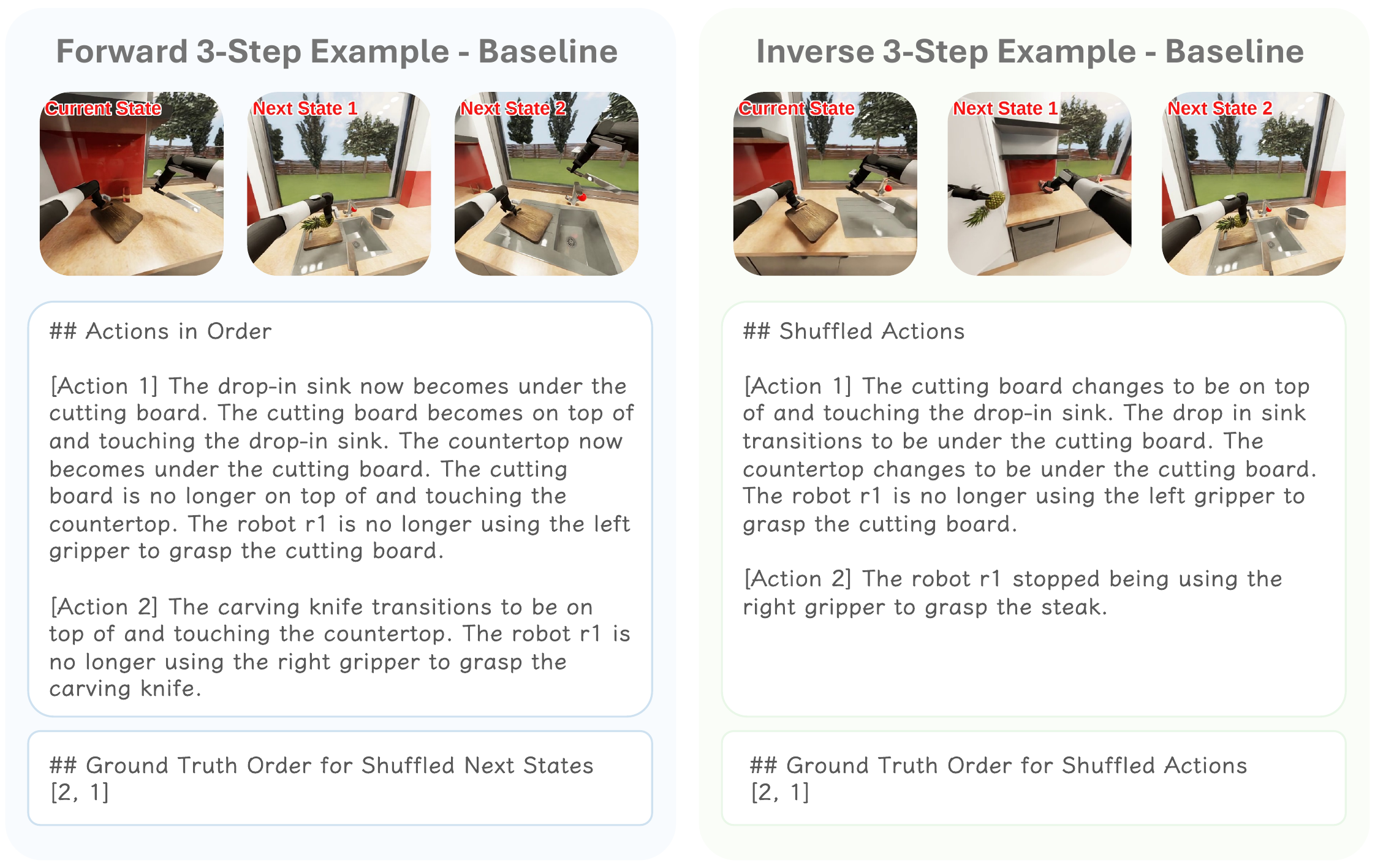}
    \caption{Illustrative trajectories of \textbf{Forward World Modeling} and \textbf{Inverse World Modeling} for a representative baseline question.}
    \label{fig:b3_baseline}
\end{figure}

\begin{figure}[htbp]
    \centering
    \includegraphics[width=\linewidth]{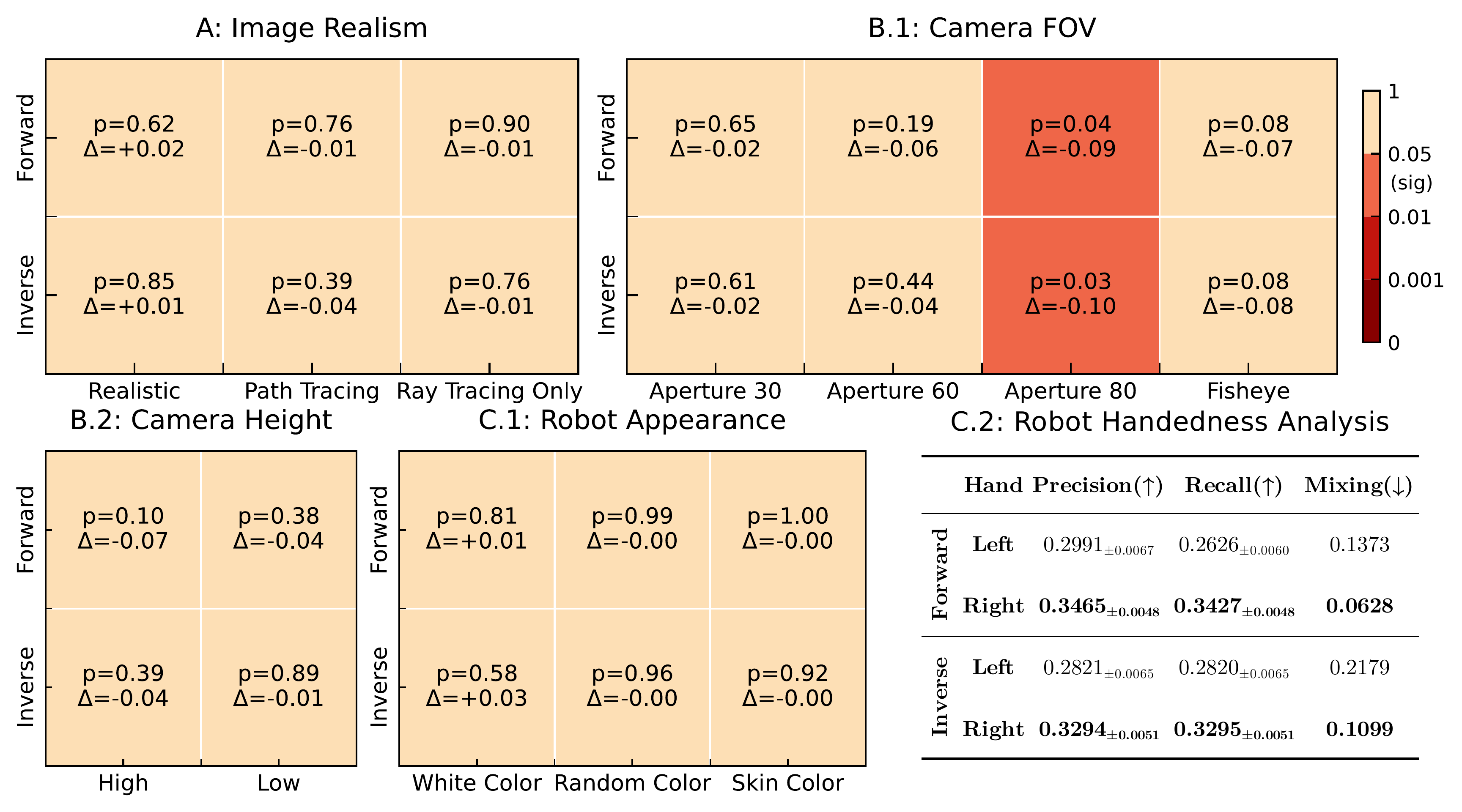}
    \caption{\textbf{Ablation experiment results with InternVL3.5-241B-A28B on \name.} Heatmaps show two-tailed unpaired p-values against the baseline, using \textit{Pairwise Accuracy}. $p<0.05$ is considered \textit{significant}. Darker red means more significant. $\Delta$ is the performance change from the baseline. If \textit{significant} and $\Delta<0$, the setting is worse than the baseline. C.2 reports the robot's performance on the left- and right-hand predicates, where \textit{Mixing} is the proportion of ground truth left or right cases that are predicted as the other hand (i.e., mixing one hand into the other hand). Note that, although InternVL3.5-241B-A28B performance is less significant than GPT-5 mini, the $|\Delta|$ across unnatural camera configurations still remains high ($>0.05$) when the same settings are significant for GPT-5 mini. }
    \label{fig_app:intern_behavior_analysis}
\end{figure}

 In our summary heatmaps (Figure~\ref{fig:behavior_analysis} for GPT-5 mini and Figure~\ref{fig_app:intern_behavior_analysis} for InternVL3.5-241B-A28B), we use $\Delta$ to visualize the performance difference between a variant and the baseline. To assess the statistical significance of these differences, we perform a two-tailed unpaired Welch's t-test. An unpaired test is appropriate as each question is evaluated in an independent session. We specifically use Welch's t-test as it does not assume equal variance between the two groups being compared (baseline vs. variant). We report the p-value for each comparison and consider a result to be statistically significant if $p < 0.05$. We qualitatively classify any performance change where $|\Delta| < 0.05$ as a small change.

We show one baseline question and its images for both forward and inverse settings in Figure~\ref{fig:b3_baseline}, and for other settings, we  \textbf{\textit{only show their images, as they all share the same question text and answers.}}

\subsection{Sensitivity to Image Realism}~\label{app_3_3:image_realism}
Although the BEHAVIOR simulator is designed to be photo-realistic, we were curious whether a ``sim-to-real'' gap might still exist due to subtle differences in rendering quality. Specifically, we sought to investigate if such a gap affects performance on our world modeling tasks and to quantify the impact of rendering fidelity on the reasoning capabilities of state-of-the-art Vision-Language Models, such as GPT-5 mini. In the following sections, we detail the experimental setup for evaluating model performance across various levels of image realism.

\begin{figure}[htbp]
    \centering
    \includegraphics[width=\linewidth]{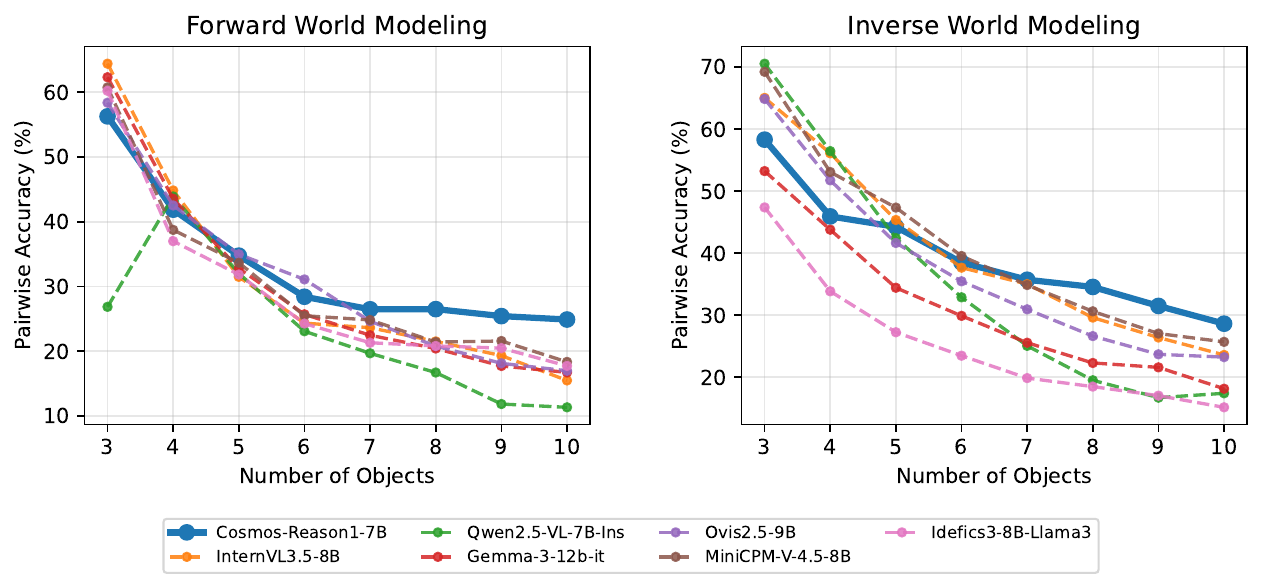}
    \caption{Comparison between Cosmos-Reason1 and other similar-sized models.}
    \label{fig_app:cosmos}
\end{figure}

\subsubsection{Realistic: Generated Images as Real World Proxy}
Since our activities are diverse and complex, reproducing simulator outputs in the real world on a one-to-one basis would incur prohibitively high costs. However, with the advent of powerful image generation models with the ability of image-scale reproduction  (e.g., GPT-image-1), it is feasible to use them as a real-world proxy to convert frames rendered by simulator into realistic styles, which provides a cost-effective and well-aligned alternative.

Constructing prompts for high-accuracy style transfer poses several challenges. First, since our segmented frames are extracted from a replayed robot trajectory, the generated realistic frames corresponding to the trajectory must preserve consistent content and style, including object shapes and appearances, lighting conditions, material properties, and camera parameters. Second, image generation models often demonstrate instability and errors in understanding fine-grained structures of robotic arms (particularly the gripper) and in interpreting robotic actions. To mitigate these issues, we establish a detailed set of rules and incorporate them into the prompt design (Figure~\ref{app_fig:realistic_img_prompt}), which improves both stability and fidelity in the generated outputs.

\begin{figure}[htbp]
  \centering
  
  \begin{subfigure}[t]{0.5\linewidth}
    \centering
    \includegraphics[width=\linewidth,page=1]{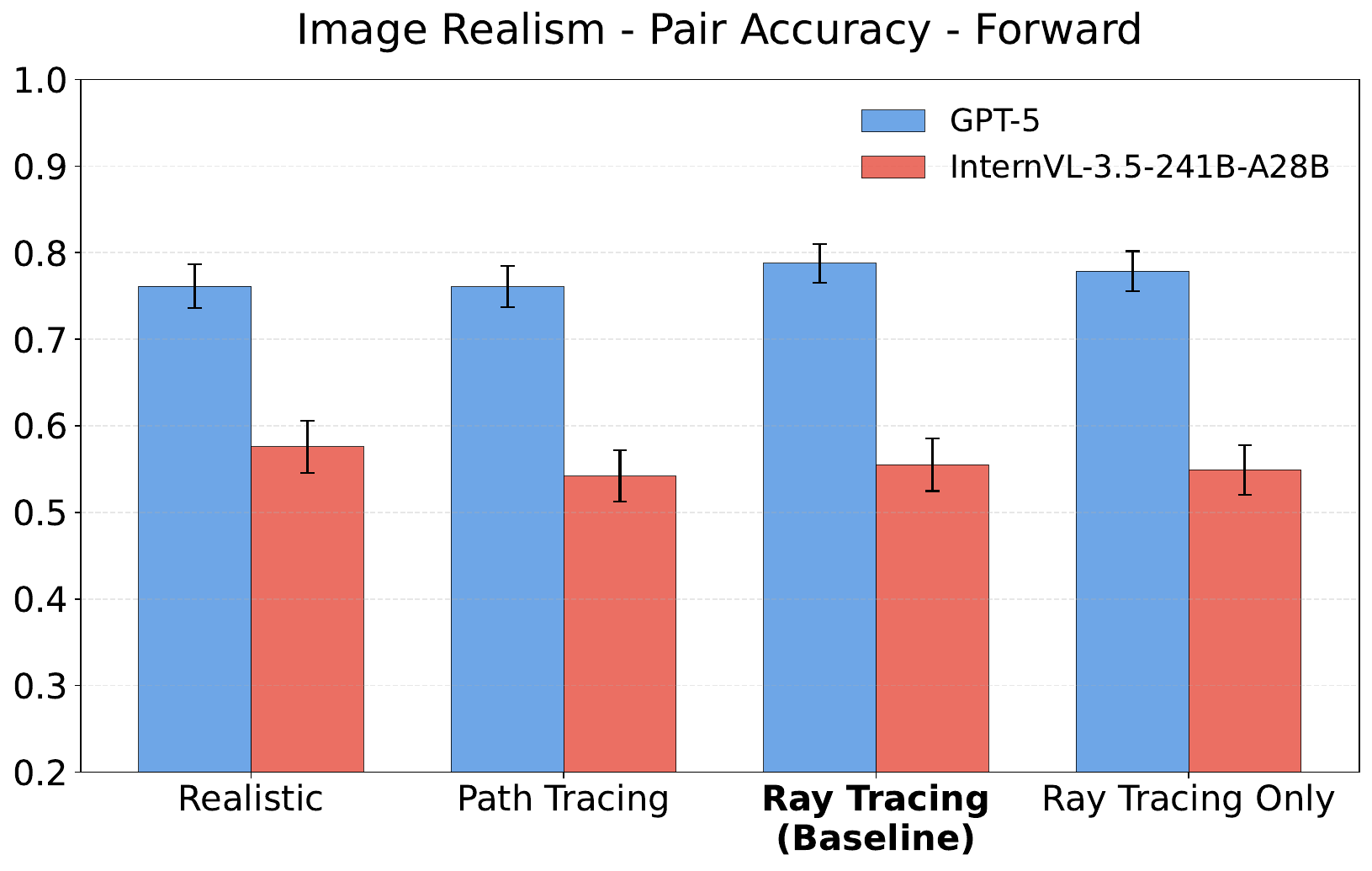}
    \caption{}
    \label{fig:f_probe_image_realism}
  \end{subfigure}\hfill
  \begin{subfigure}[t]{0.5\linewidth}
    \centering
    \includegraphics[width=\linewidth]{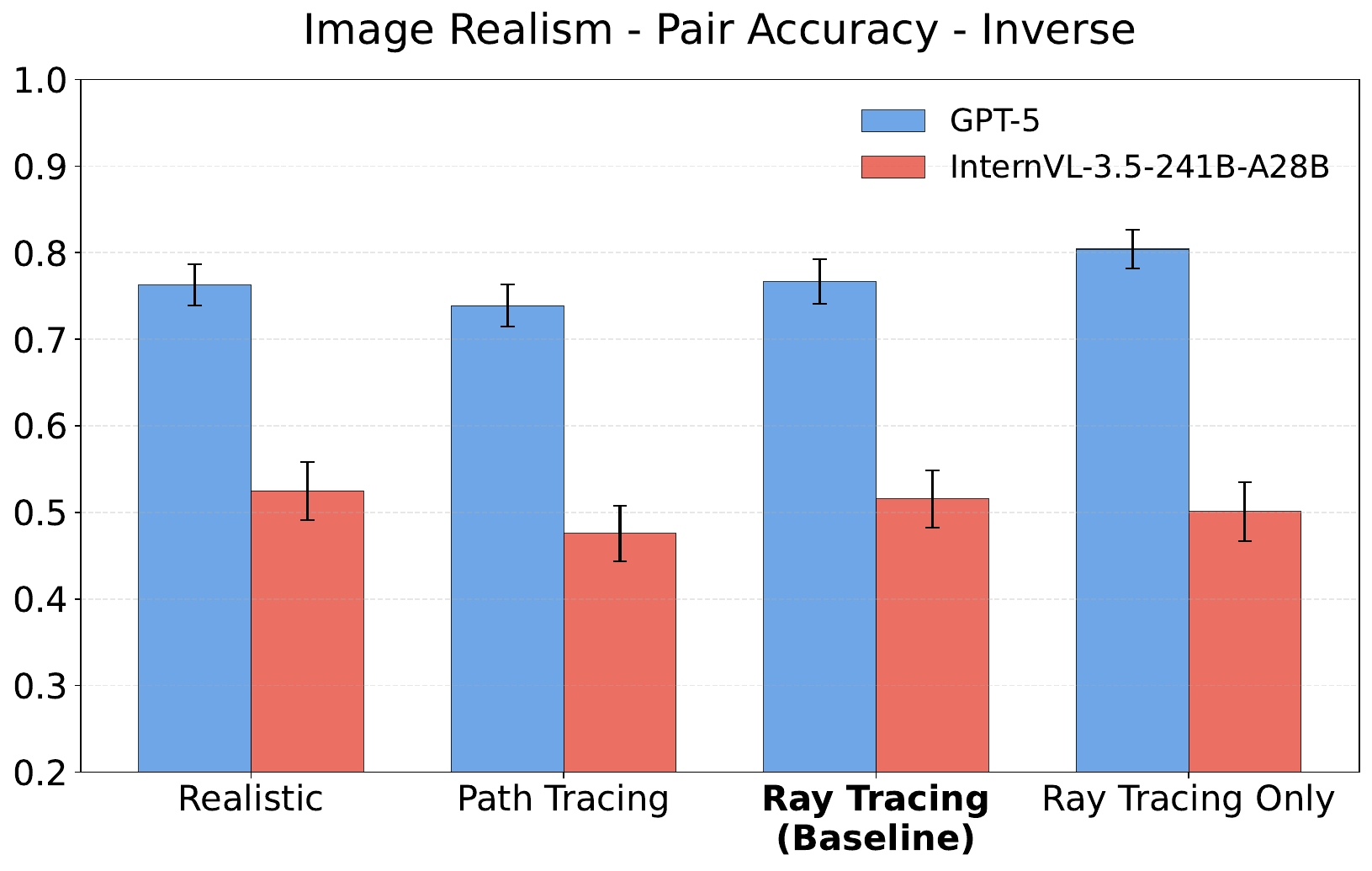}
    \caption{}
    \label{fig:i_probe_image_realism}
  \end{subfigure}

    \caption{\textbf{Ablating image realism with GPT-5 and InternVL3.5-241B-A28B.} (a) Forward dynamics; (b) Inverse dynamics. Bar plots report Pairwise Accuracy across four rendering settings—Realistic, Path Tracing, Ray Tracing (Baseline), and Ray Tracing Only. Error bars denote ±SEM. The baseline x-tick is bolded.}
  \label{fig:fi_probe_image_realism}
\end{figure}

\begin{figure}[htbp]
    \centering
    \includegraphics[width=\linewidth]{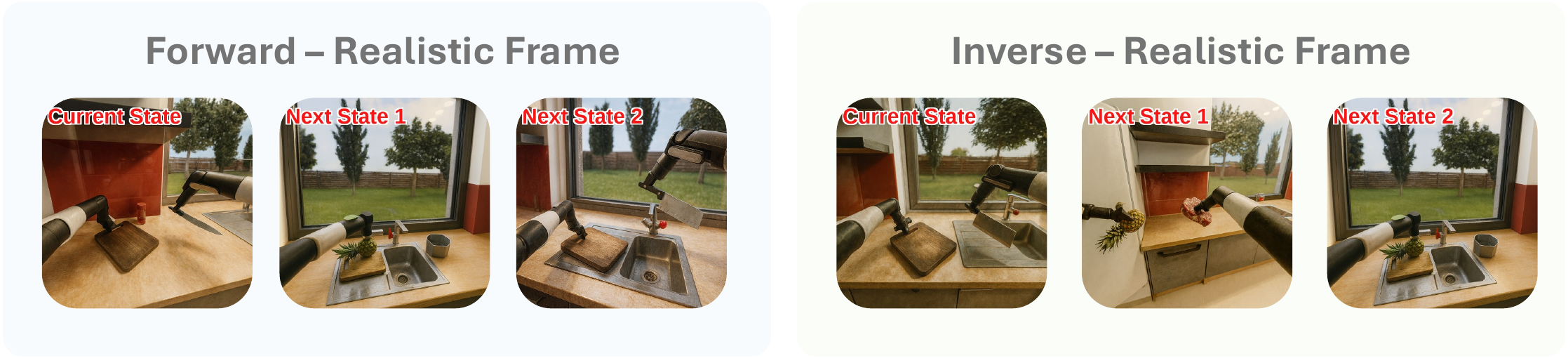}
    \caption{Examples of simulator frames converted into realistic styles for both \textbf{Forward World Modeling} (left) and \textbf{Inverse World Modeling} (right) trajectories.}
    \label{fig:b4_img_realism}
\end{figure}

\begin{tcolorbox}[colback=black!5!white, colframe=black!75!white, 
title=Project Instruction and Prompt for Image Generation, boxrule=0.5mm, width=\textwidth, arc=2mm, auto outer arc=true, breakable]
\scriptsize
\begin{verbatim}
## Below are the instructions and regulations, treat them as the sole, global reference 
for all image generations you are going to perform.

## Core Objective
Convert simulator screenshots into photorealistic PBR images. Change style only; 
do not change content.

## Content Lock (Content-Locked)
Preserve the count, position, size, geometry, and pose of all objects. 
The robot hand and knife angles, shapes, and actions must match exactly.

## Camera crop and viewpoint must remain unchanged.
The outdoor scene must remain daytime; tree and fence silhouettes must not change. 
If realism conflicts with content, content fidelity takes precedence.

## Style Requirements
* Lighting: Warm under-cabinet tungsten (3200-3600 K) + soft window daylight fill.
* Tone: Filmic contrast, smooth highlight roll-off, no crushed blacks or blown 
highlights.
* Camera: approximately 35 mm, f/2.8-4, shallow DOF; subject sharp with gently blurred 
background.
* Shadows: Realistic soft shadows, contact shadows, and ambient occlusion.

## Materials:
* Metal knife and trims: Brushed, anisotropic metal.
* Robot: Matte polymer.
* Cutting board and countertop: Sealed/oiled wood grain.
* Glass/walls: Glossy glass with realistic reflections and refractions.
* Post-processing: Subtle camera grain; light vignette.
* Prohibited: Cartoonish look, plastic sheen, bloom, oversaturation, hard outline 
sharpening, fake lighting effects.

## Acceptance Criteria
* Edge alignment: SSIM >= 0.95 (along object boundaries).
* Segmentation: IoU >= 0.98 for robot, knife, cutting board, outdoors.
* Color difference: delta Hue <= 3°, delta L <= 6.
* Knife shape error: <= 1 px.
* Outdoor tree/fence silhouette error: <= 1-2 px.

## Implementation Suggestions
* Use low denoise strength 0.20-0.35, CFG 4-6.
* Negative prompt: forbid new objects, geometry changes, 
cartoonish/oversaturated/plastic textures.
* Detail pass: add micro-surface material detail + light film grain.

Now, review and summarize what you have learned from these instructions.

Following the instructions you have learned, transform the given image into realistic 
photograph style.
\end{verbatim}
\end{tcolorbox}
\captionof{figure}{The prompt used to generate realistic photographic style images from segmented frames (of a replayed robot trajectory).}
\label{app_fig:realistic_img_prompt}

\subsubsection{Path Tracing Setup}
\begin{figure}[htbp]
    \centering
    \includegraphics[width=\linewidth]{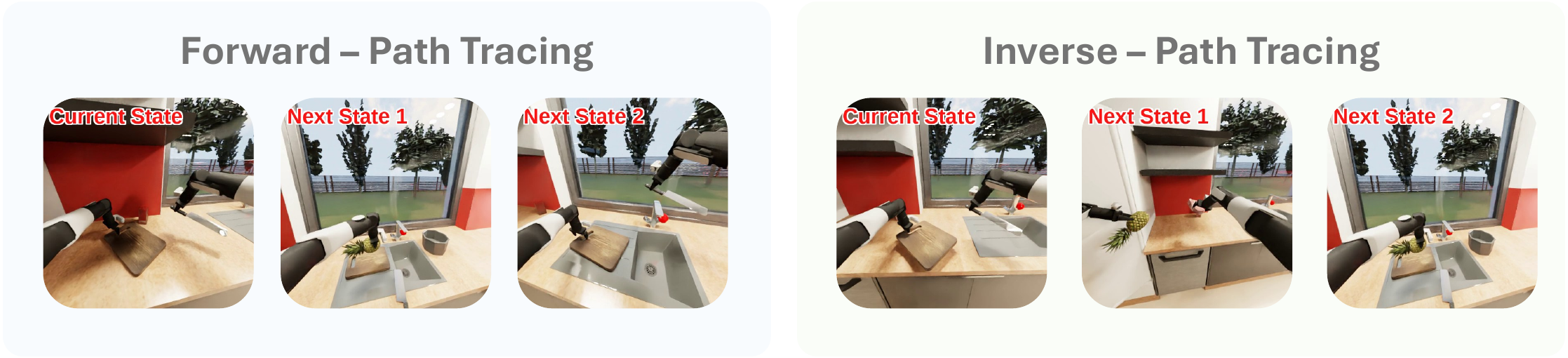}
    \caption{The figure illustrates \textbf{Forward World Modeling} (left) and \textbf{Inverse World Modeling} (right) trajectories rendered using the path tracing engine in NVIDIA Isaac Sim.}
    \label{fig:b4_path_tracing}
\end{figure}

To generate the highest-fidelity images for our analysis of image realism, we utilized path tracing. This was achieved directly through the built-in, real-time path tracing engine provided by the NVIDIA Isaac Sim simulator. An example can be seen in Figure~\ref{fig:b4_path_tracing}.

\subsubsection{Ray Tracing Only Setup}
\begin{figure}[htbp]
    \centering
    \includegraphics[width=\linewidth]{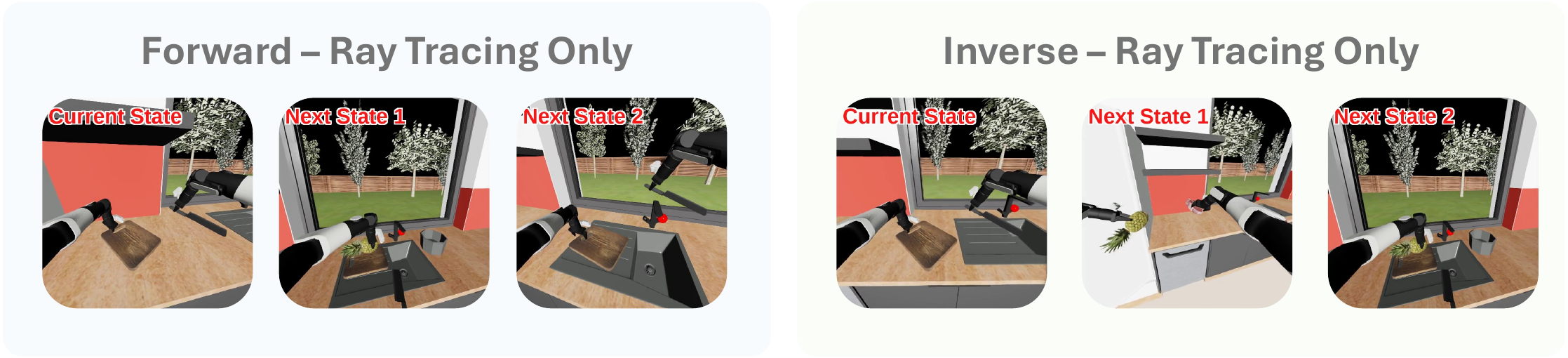}
    \caption{Examples of an intermediate rendering style created with a simplified ray tracing pipeline for \textbf{Forward World Modeling} (left) and \textbf{Inverse World Modeling} (right) trajectories.}
    \label{fig:b4_ray_tracing_only}
\end{figure}

This setup was designed to represent an intermediate rendering quality (representing `unrealistic'). While it still utilizes the ray tracing pipeline as its foundation, we manually disabled several advanced lighting and post-processing effects to reduce visual fidelity. Specifically, we turned off the following features: reflections, DLSS, ambient occlusion, sampled lighting, ambient light, and flow. The resulting visual style, which lacks these richer effects, can be seen in Figure~\ref{fig:b4_ray_tracing_only}.

\begin{table*}[!htbp]
\centering
\setlength{\tabcolsep}{2pt}
\small
\resizebox{\textwidth}{!}{
\begin{tabular}{@{}p{3mm} l cccccccc@{}}
\toprule
& \textbf{Metric} & \bf 3 & \bf 4 & \bf 5 & \bf 6 & \bf 7 & \bf 8 & \bf 9 & \bf 10 \\
\midrule
\multicolumn{10}{>{\columncolor{forwardbg}}l}{\textit{\footnotesize Forward (with random perturbations)}}\\
& Task Accuracy
& $68.51 \pm 1.44$ & $37.47 \pm 2.11$ & $25.75 \pm 1.74$ & $10.80 \pm 1.59$ & $4.60 \pm 2.11$ & $1.61 \pm 0.80$ & $0.92 \pm 0.40$ & $0.00 \pm 0.00$ \\
& Pairwise Accuracy
& $75.52 \pm 1.19$ & $59.77 \pm 1.82$ & $54.48 \pm 2.54$ & $42.25 \pm 1.30$ & $37.36 \pm 1.30$ & $30.90 \pm 0.59$ & $28.76 \pm 0.43$ & $25.98 \pm 1.01$ \\
\midrule
\multicolumn{10}{>{\columncolor{inversebg}}l}{\textit{\footnotesize Inverse (with random perturbations)}}\\
& Task Accuracy
& $83.45 \pm 0.69$ & $58.39 \pm 3.40$ & $43.91 \pm 1.05$ & $22.76 \pm 2.07$ & $13.79 \pm 0.00$ & $6.67 \pm 2.22$ & $4.14 \pm 2.07$ & $0.73 \pm 1.26$ \\
& Pairwise Accuracy
& $83.45 \pm 0.69$ & $68.74 \pm 1.40$ & $61.49 \pm 0.72$ & $52.46 \pm 0.69$ & $45.59 \pm 1.23$ & $39.70 \pm 1.64$ & $35.09 \pm 2.57$ & $29.01 \pm 1.64$ \\
\bottomrule
\end{tabular}
}
\caption{\textbf{Robustness to random perturbations in predicate deltas.}
Mean $\pm$ standard deviation (\%) over three random seeds on a 2{,}304-QA subset of \name for InternVL3.5--241B.}
\label{tab:predicate_noise}
\end{table*}

\subsubsection{Robustness to Noise in Predicate Deltas}
\label{sec:app:predicate-noise}

To assess whether our conclusions are sensitive to noise in the abstract transitions, we run a robustness study on a subset of 2{,}304 QAs sampled from \name.
For each trajectory, we randomly perturb the symbolic predicates in the abstract deltas and re-run both forward and inverse evaluations for InternVL3.5--241B over three random seeds.
Table~\ref{tab:predicate_noise} reports the mean $\pm$ standard deviation across seeds.

Across all horizons, we observe only very small standard deviations, and the qualitative trends remain unchanged:
Inverse world modeling consistently outperforms forward world modeling at comparable horizons, and performance for both directions degrades sharply as the number of steps increases.
These results indicate that our findings are stable across activities and robust to random perturbations in the predicate deltas, rather than being driven by a few particularly clean or favorable trajectories.

\subsection{Sensitivity to Camera Configurations}~\label{app_3_4:camera_config}

\subsubsection{Camera Aperture Setup}

\begin{figure}[htbp]
    \centering
    \includegraphics[width=\linewidth]{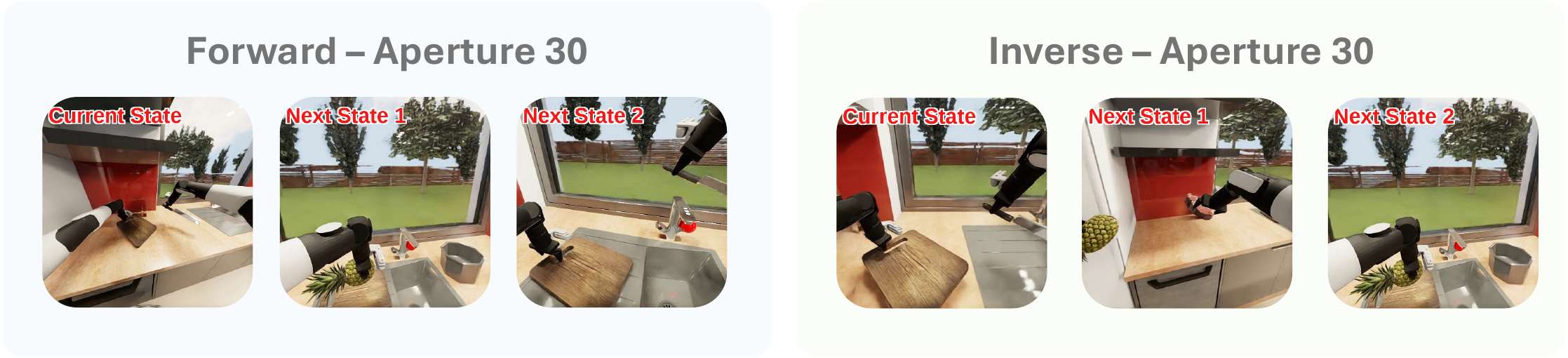}
    \caption{Example trajectories of \textbf{Forward World Modeling} (left) and \textbf{Inverse World Modeling} (right), captured with a camera aperture of 30.}
    \label{fig:b4_aperture30}
\end{figure}

\begin{figure}[htbp]
    \centering
    \includegraphics[width=\linewidth]{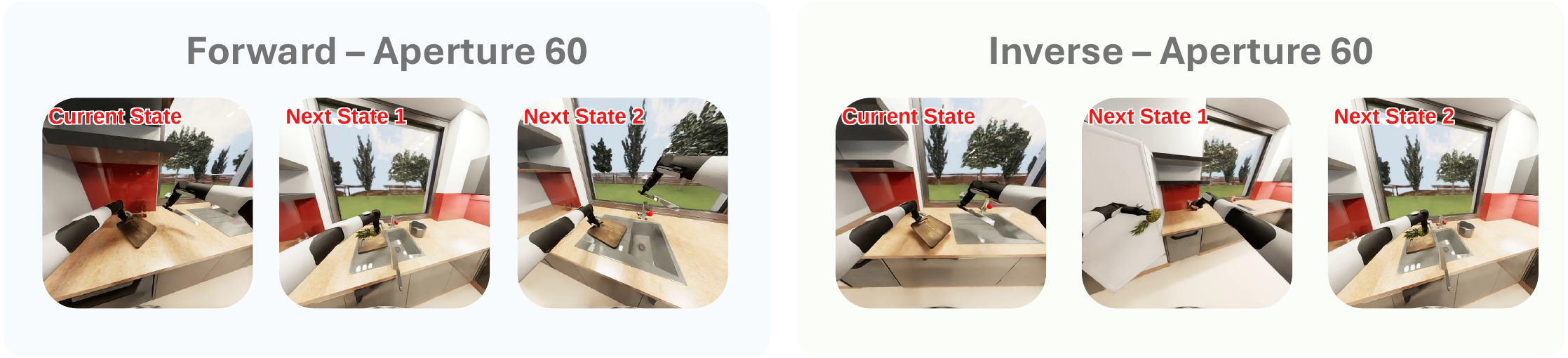}
    \caption{Example trajectories of \textbf{Forward World Modeling} (left) and \textbf{Inverse World Modeling} (right), captured with a camera aperture of 60.}
    \label{fig:b4_aperture60}
\end{figure}

\begin{figure}[htbp]
    \centering
    \includegraphics[width=\linewidth]{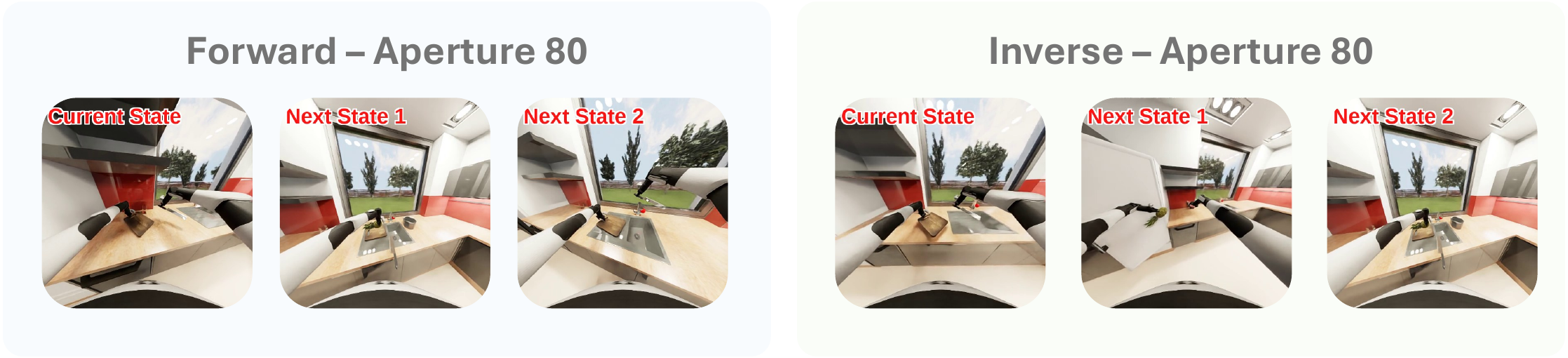}
    \caption{Example trajectories of \textbf{Forward World Modeling} (left) and \textbf{Inverse World Modeling} (right), captured with a camera aperture of 80.}
    \label{fig:b4_aperture80}
\end{figure}

Our default baseline is aperture 40. We also investigate apertures 30, 60, and 80. Examples can refer to Figure~\ref{fig:b4_aperture30},~\ref{fig:b4_aperture60}, and~\ref{fig:b4_aperture80}.

\begin{figure}[t]
  \centering
  
  \begin{subfigure}[t]{0.5\linewidth}
    \centering
    \includegraphics[width=\linewidth,page=1]{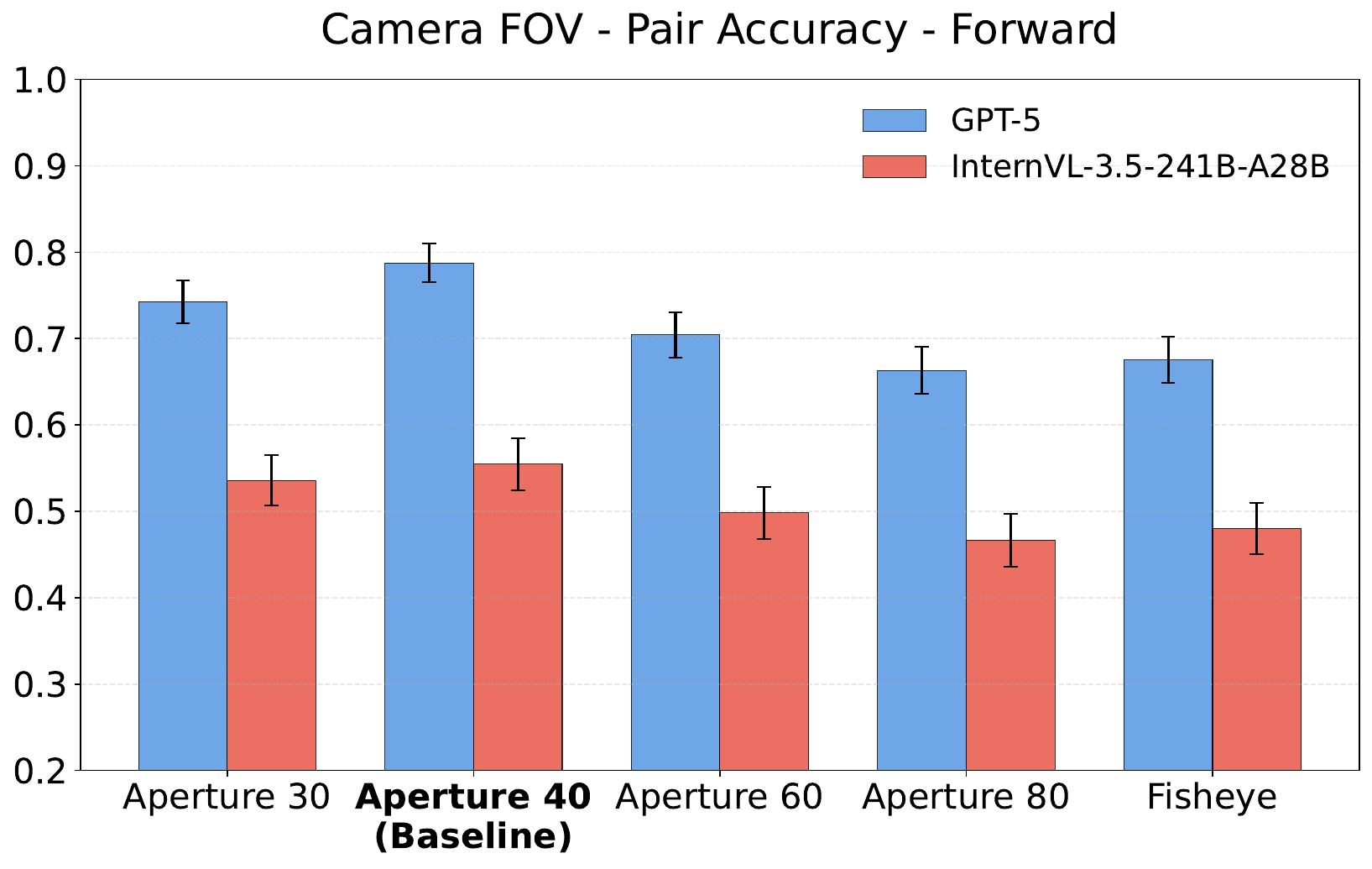}
    \caption{}
    \label{fig:f_probe_camera_fov}
  \end{subfigure}\hfill
  \begin{subfigure}[t]{0.5\linewidth}
    \centering
    \includegraphics[width=\linewidth]{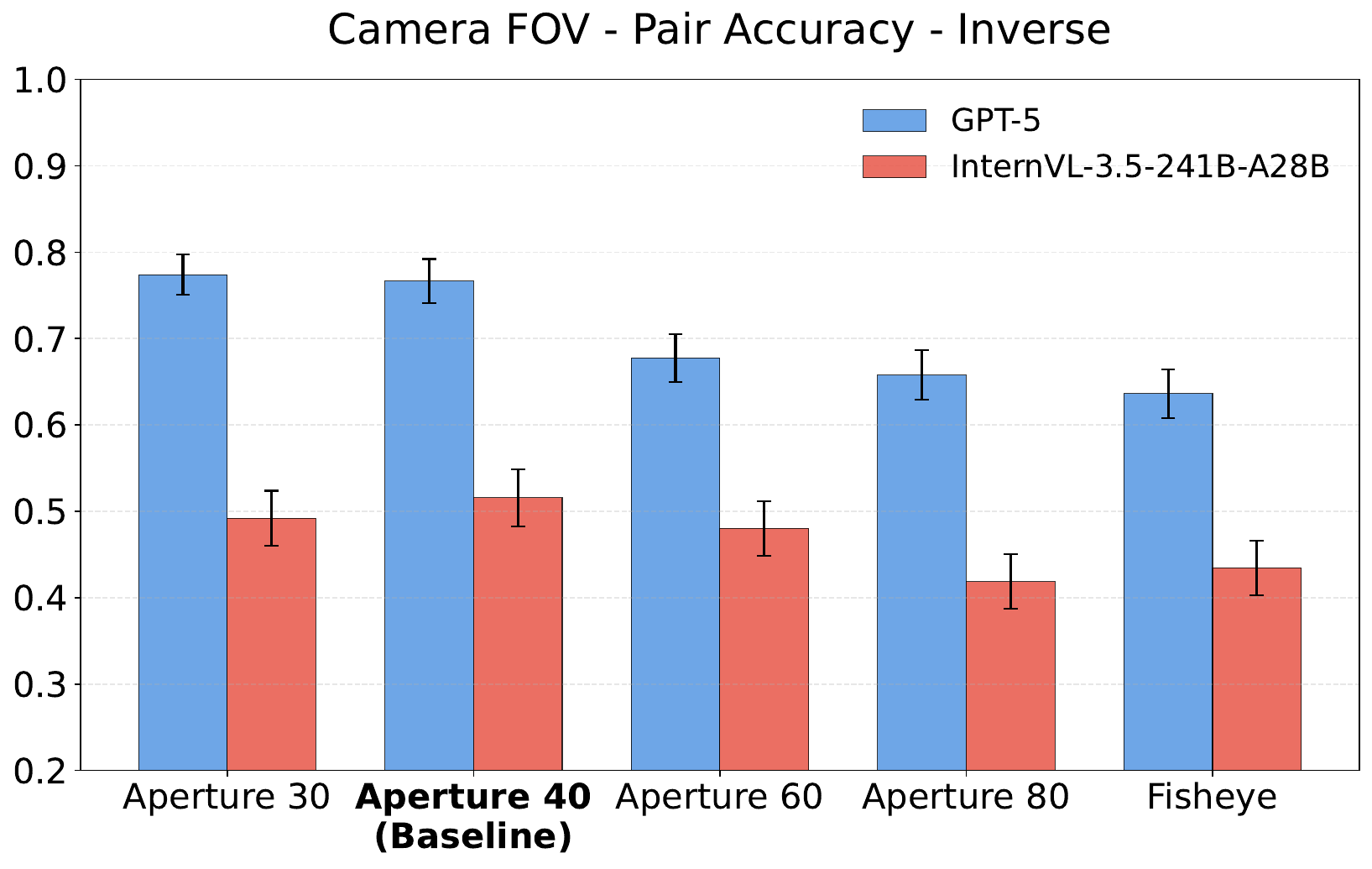}
    \caption{}
    \label{fig:i_probe_camera_fov}
  \end{subfigure}
    \caption{\textbf{Ablating camera field-of-view (FOV) with GPT-5 and InternVL-3.5-241B-A28B.} (a) Forward dynamics; (b) Inverse dynamics. Bar plots report Pairwise Accuracy across five lens settings—Aperture 30, Aperture 40 (Baseline), Aperture 60, Aperture 80, and Fisheye. Error bars denote ±SEM; the baseline tick is bolded.}
  \label{fig:fi_probe_camera_fov}
\end{figure}

\subsubsection{Fisheye Lens Setup}
\begin{figure}[!htbp]
    \centering
    \includegraphics[width=\linewidth]{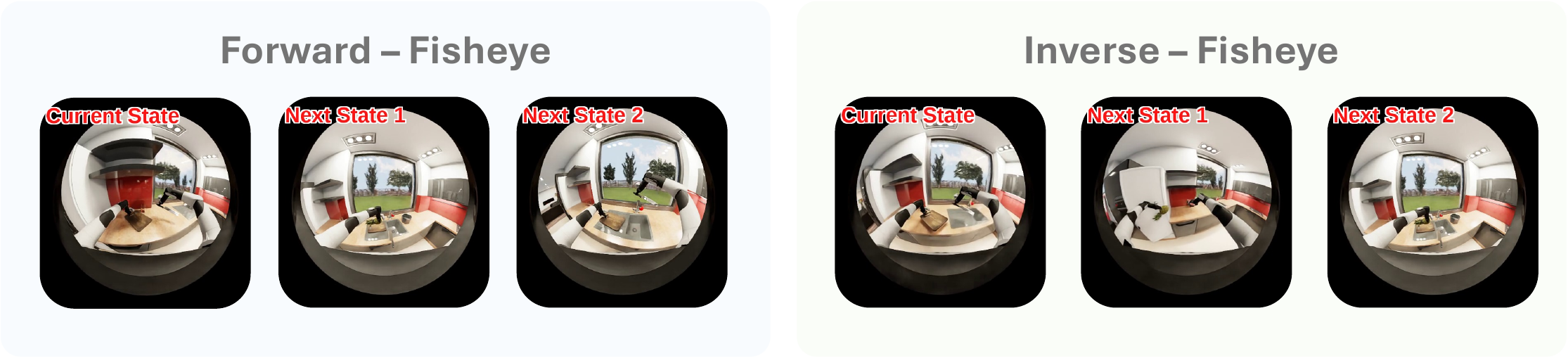}
    \caption{Example trajectories of \textbf{Forward World Modeling} (left) and \textbf{Inverse World Modeling} (right), captured with a fisheye-style camera.}
    \label{fig:b4_fisheye}
\end{figure}

Isaac Sim provides the fisheye lens settings. We choose \texttt{fisheyePolynomial}, which is the most similar to a daily fisheye lens, such as GoPro, as our evaluated target. The effect can be seen in the example Figure~\ref{fig:b4_fisheye}.

\subsubsection{Camera Height Setup}
\begin{figure}[!htbp]
    \centering
    \includegraphics[width=\linewidth]{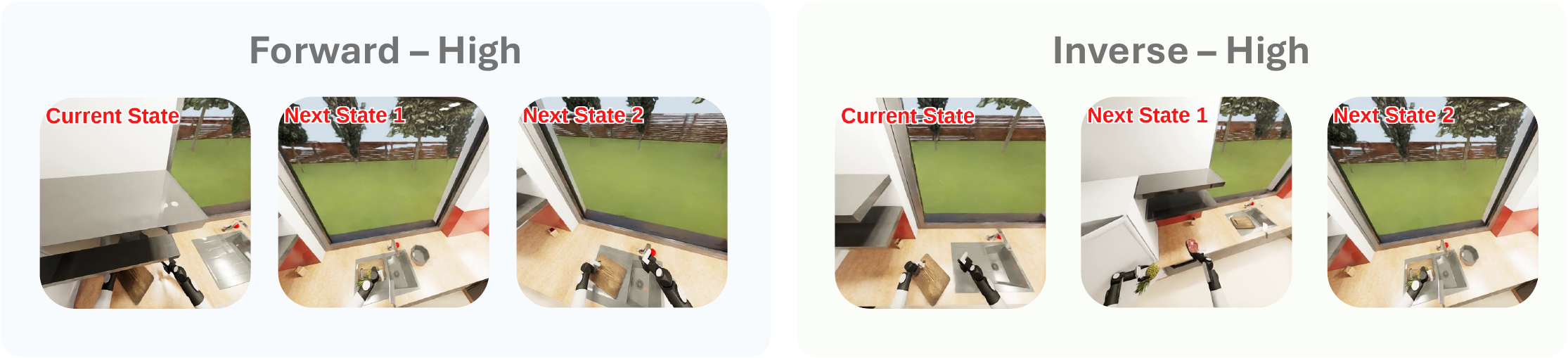}
    \caption{Example trajectories of \textbf{Forward World Modeling} (left) and \textbf{Inverse World Modeling} (right), captured from a camera height of 2.25 m.}
    \label{fig:b4_high}
\end{figure}

\begin{figure}[!htbp]
    \centering
    \includegraphics[width=\linewidth]{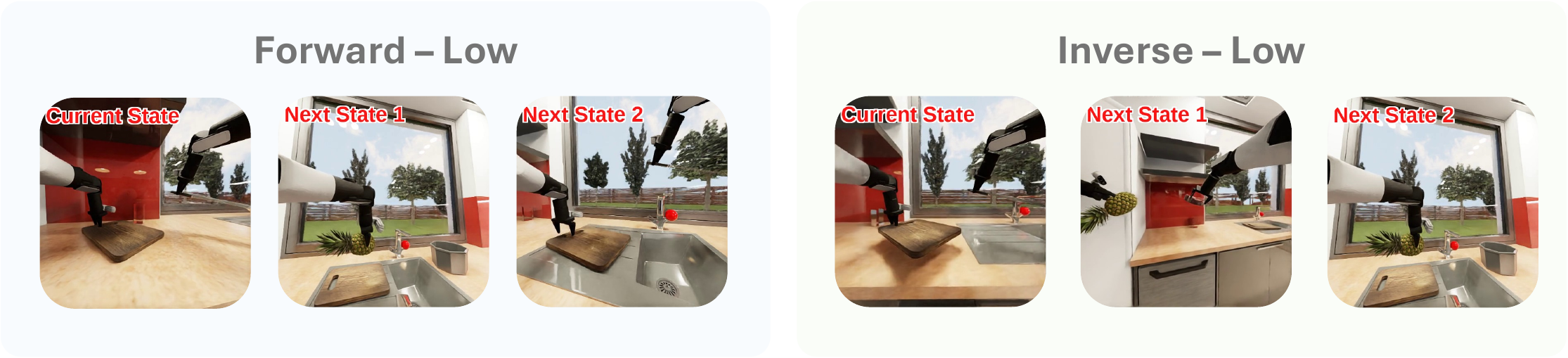}
    \caption{Example trajectories of \textbf{Forward World Modeling} (left) and \textbf{Inverse World Modeling} (right), captured from a camera height of 1.5 m.}
    \label{fig:b4_low}
\end{figure}

The default setting height is $1.75\ \mathrm{m}$, we also investigate the high ($+0.5\mathrm{m}$) setting and low ($-0.25\mathrm{m}$)setting, and the examples are shown in Figure~\ref{fig:b4_high} and~\ref{fig:b4_low}.

\begin{figure}[!htbp]
  \centering
  \begin{subfigure}[t]{0.5\linewidth}
    \centering
    \includegraphics[width=\linewidth,page=1]{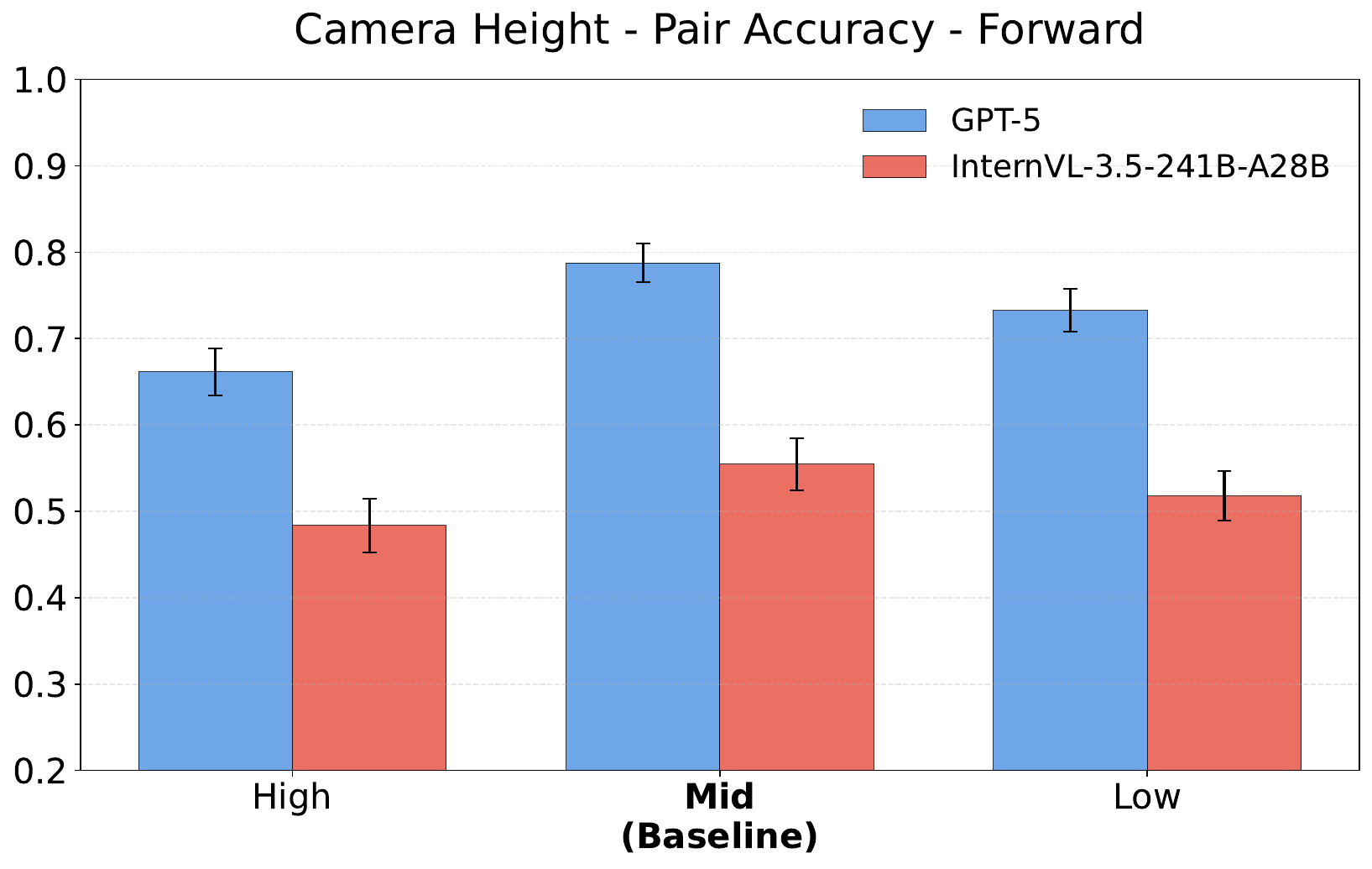}
    \caption{}
    \label{fig:f_probe_camera_height}
  \end{subfigure}\hfill
  \begin{subfigure}[t]{0.5\linewidth}
    \centering
    \includegraphics[width=\linewidth]{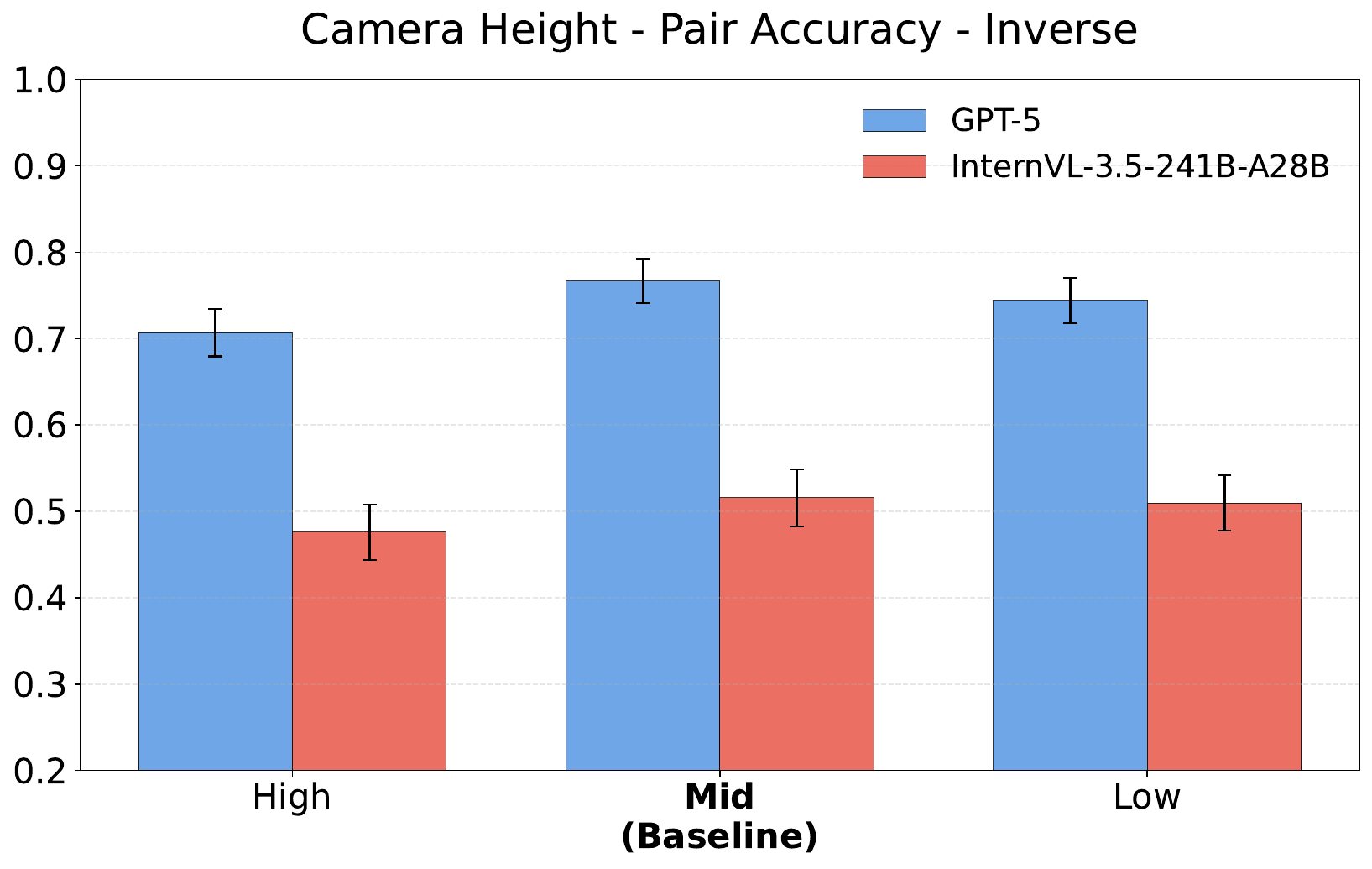}
    \caption{}
    \label{fig:i_probe_camera_height}
  \end{subfigure}

    \caption{\textbf{Ablating camera height with GPT-5 and InternVL-3.5-241B-A28B.} (a) Forward dynamics; (b) Inverse dynamics. Bar plots report Pairwise Accuracy across three viewpoints (High, Mid baseline, and Low). Error bars denote $\pm$SEM; the baseline tick is bolded.}
  \label{fig:fi_probe_camera_height}
\end{figure}

\subsection{Do VLMs Have Embodied Biases?}

\subsubsection{Robot Appearance}~\label{app_3_5_1:robot_appearance}
\begin{figure}[!htbp]
    \centering
    \includegraphics[width=\linewidth]{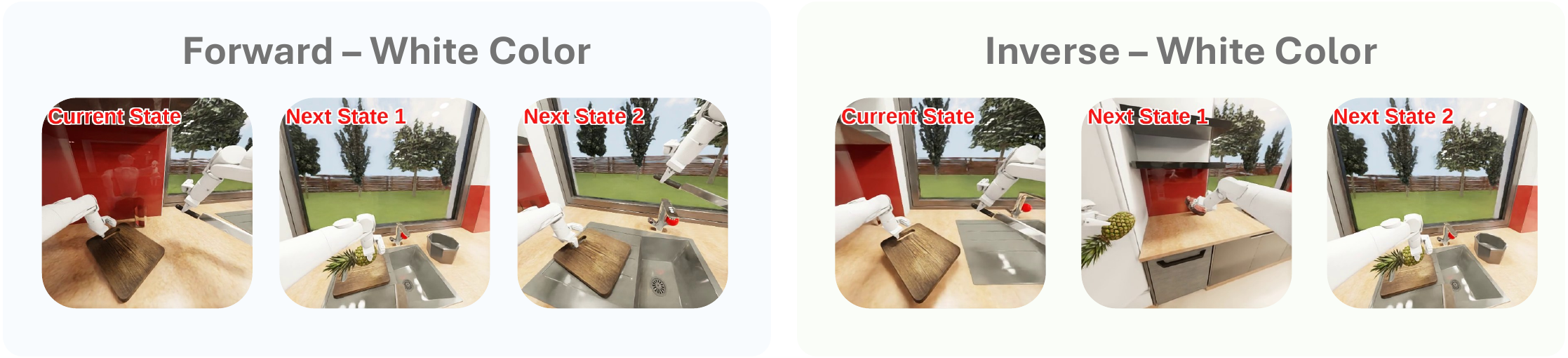}
    \caption{Example trajectories of \textbf{Forward World Modeling} (left) and \textbf{Inverse World Modeling} (right), with the robot gripper rendered in white.}
    \label{fig:b4_white_color}
\end{figure}

\begin{figure}[!htbp]
    \centering
    \includegraphics[width=\linewidth]{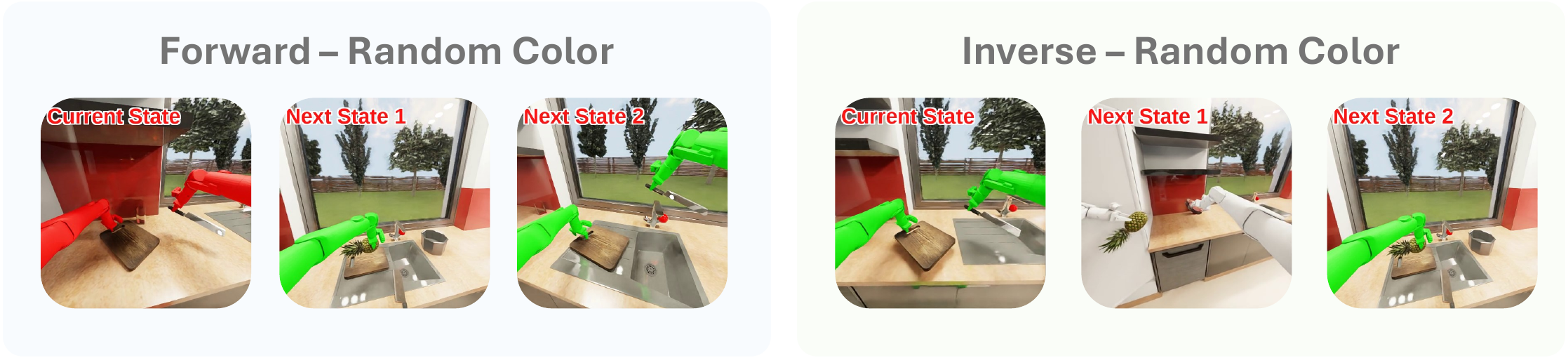}
    \caption{Example trajectories of \textbf{Forward World Modeling} (left) and \textbf{Inverse World Modeling} (right), with the robot gripper rendered in a random color at each frame.}
    \label{fig:b4_random_color}
\end{figure}

\begin{figure}[!htbp]
    \centering
    \includegraphics[width=\linewidth]{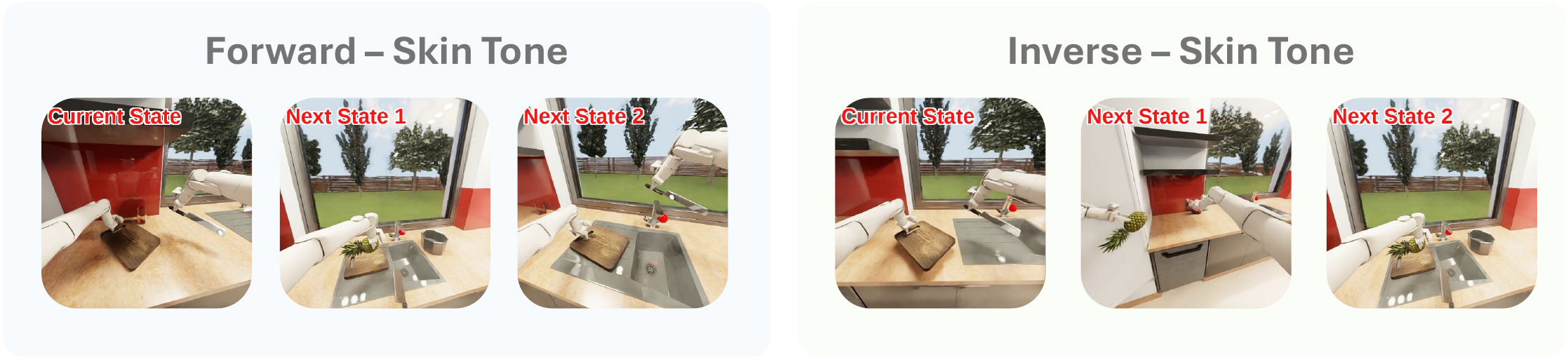}
    \caption{Example trajectories of \textbf{Forward World Modeling} (left) and \textbf{Inverse World Modeling} (right), with the robot gripper rendered in a human skin–like color.}
    \label{fig:b4_skin_tone}
\end{figure}

We test three variants: White Color, Random Color (robot color is randomized at each frame), and Skin Color (robot is rendered with a human-like skin tone). Examples can be referred to Figure~\ref{fig:b4_white_color},~\ref{fig:b4_random_color}, and~\ref{fig:b4_skin_tone}.

\begin{figure}[t]
  \centering
  
  \begin{subfigure}[t]{0.5\linewidth}
    \centering
    \includegraphics[width=\linewidth,page=1]{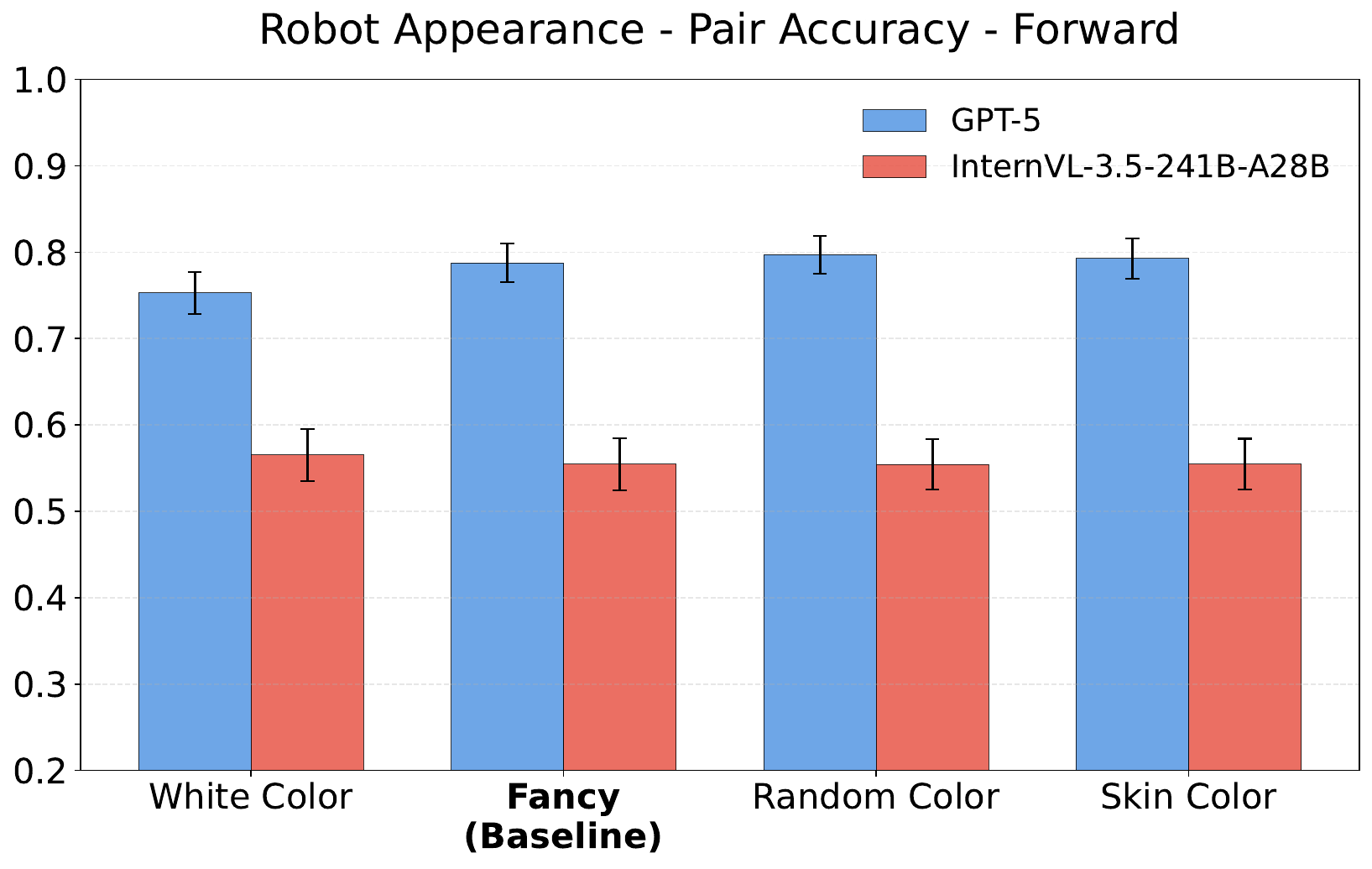}
    \caption{}
    \label{fig:f_probe_robot_appearance}
  \end{subfigure}\hfill
  \begin{subfigure}[t]{0.5\linewidth}
    \centering
    \includegraphics[width=\linewidth]{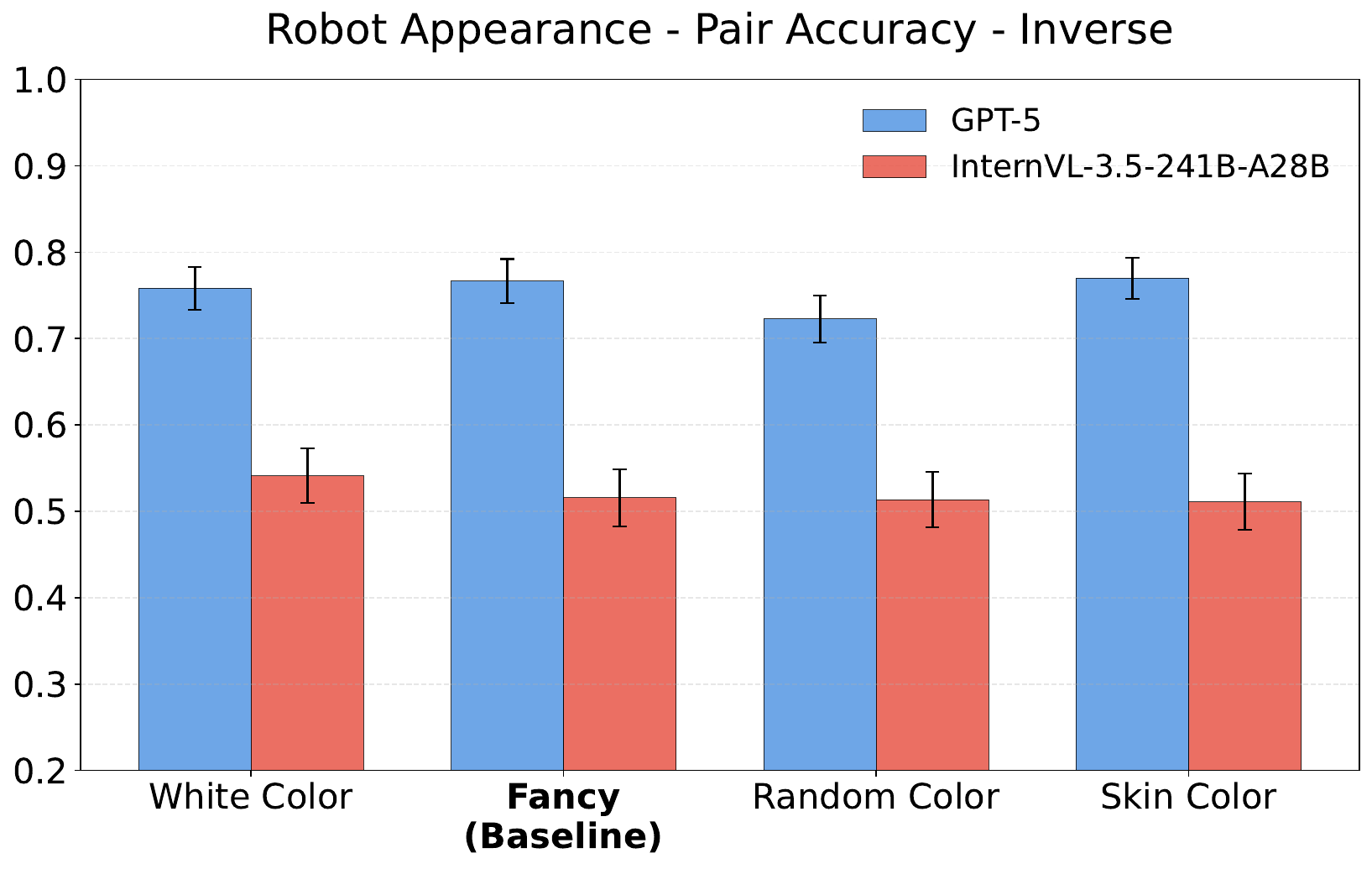}
    \caption{}
    \label{fig:i_probe_robot_appearance}
  \end{subfigure}

    \caption{\textbf{Ablating robot appearance with GPT-5 and InternVL-3.5-241B-A28B.}(a) Forward dynamics; (b) Inverse dynamics. Bar plots report Pairwise Accuracy across four styles—White Color, Fancy (Baseline), Random Color, and Skin Color. Error bars denote ±SEM; the baseline tick is bolded.}
  \label{fig:fi_probe_robot_appearance}
\end{figure}

\subsubsection{Handedness} \label{app_3_6_2:hand}

\begin{figure}[t]
  \centering
  
  \begin{subfigure}[t]{0.5\linewidth}
    \centering
    \includegraphics[width=\linewidth,page=1]{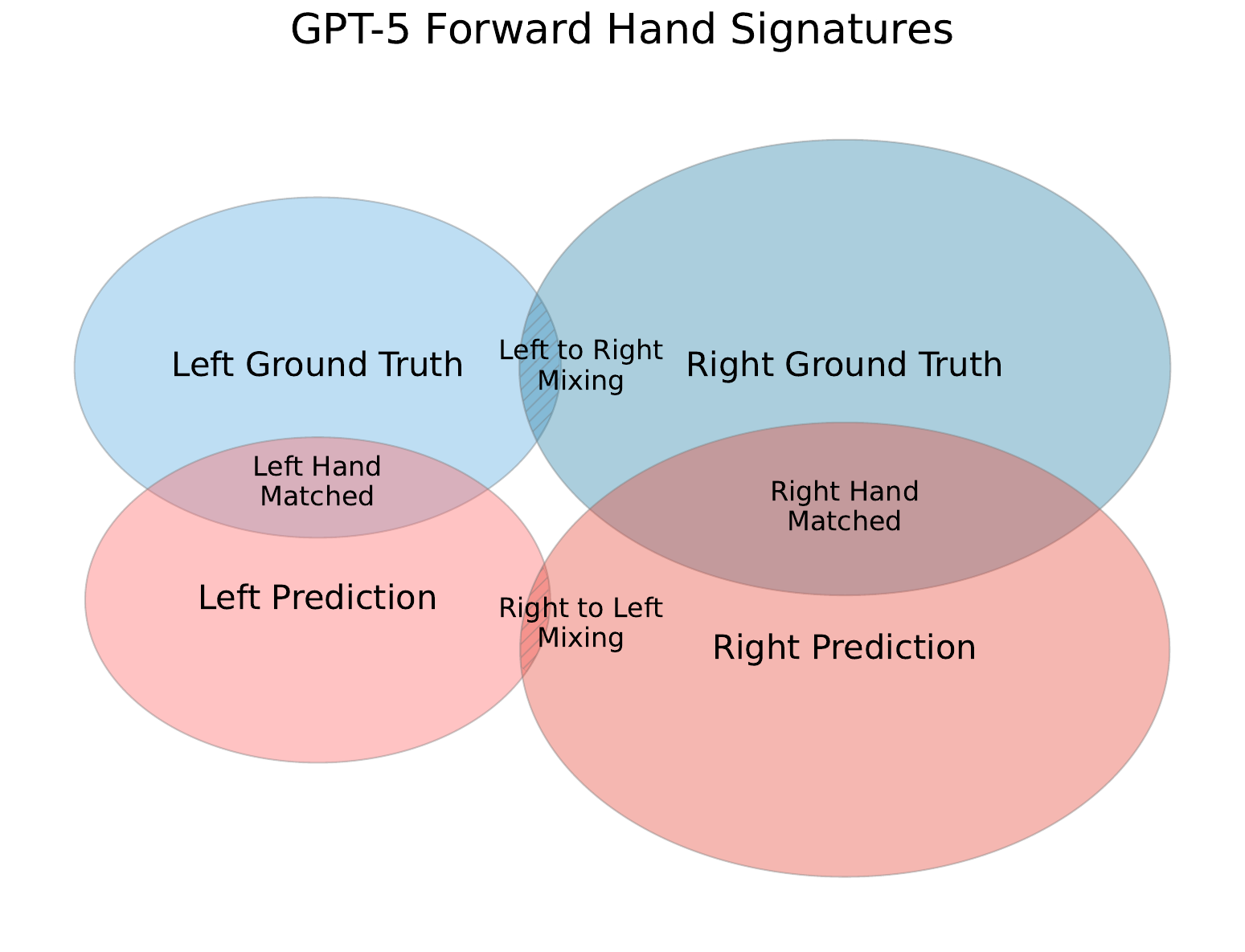}
    \caption{}
    \label{fig:f_ellipse}
  \end{subfigure}\hfill
  \begin{subfigure}[t]{0.5\linewidth}
    \centering
    \includegraphics[width=\linewidth]{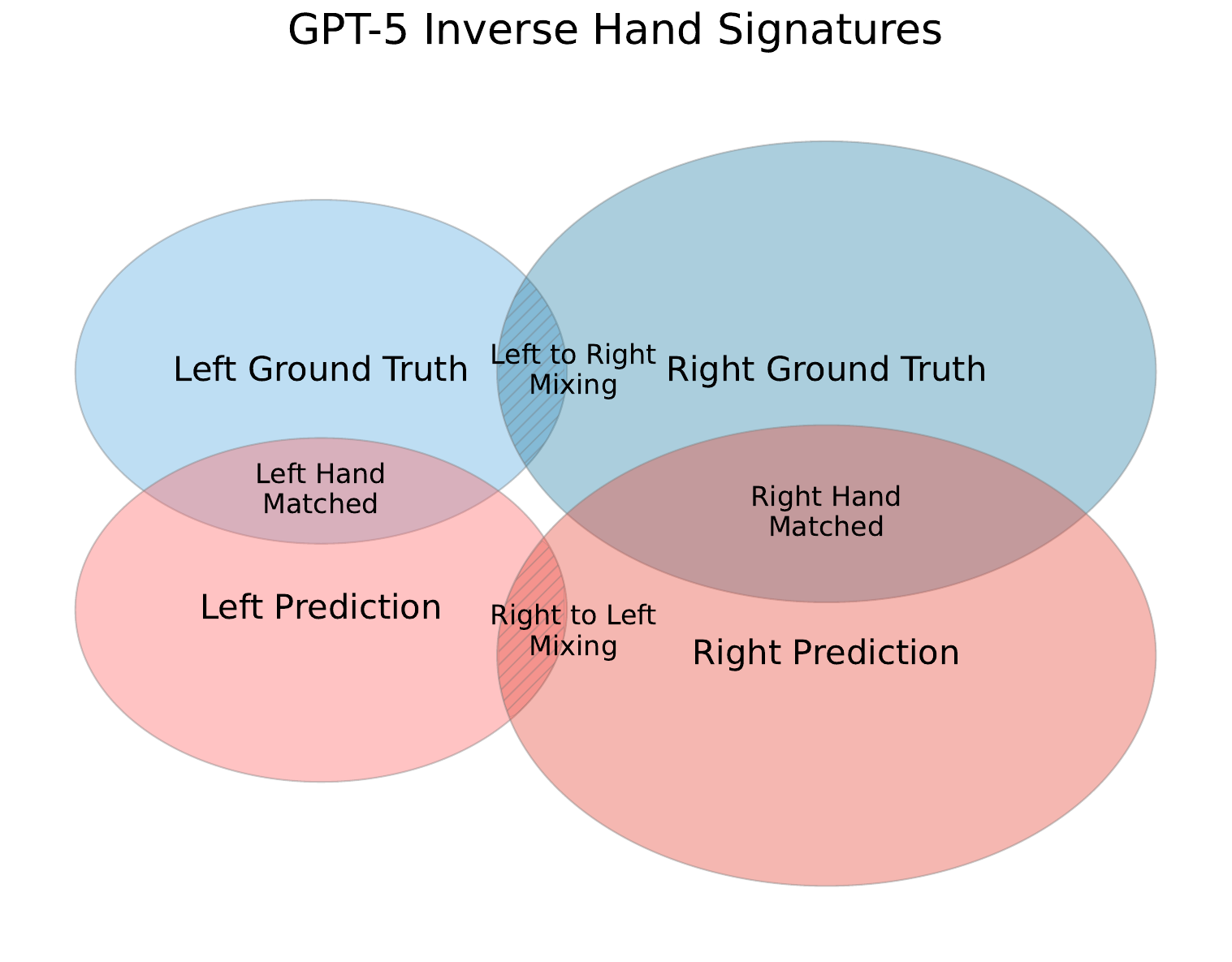}
    \caption{}
    \label{fig:i_ellipse}
  \end{subfigure}

    \caption{Illustration of overlaps between ground truth and GPT-5 predictions sets for left and right hands related signatures in forward and inverse tasks. The size of ellipses project the total counts of signatures, and overlaps denote matched signatures (center regions) or mixing errors (cross-hand overlaps). }
  \label{fig:ellipse_fi}
\end{figure}

Based on our experimental setup (\ref{app_4_1_3:handedness}). We further examine whether predictions involving agent interactions reflect real-world handedness asymmetry (typically favoring the right hand). 

\begin{figure}[htbp]
  \centering
  
  \begin{subfigure}[t]{0.5\linewidth}
    \centering
    \includegraphics[width=\linewidth,page=1]{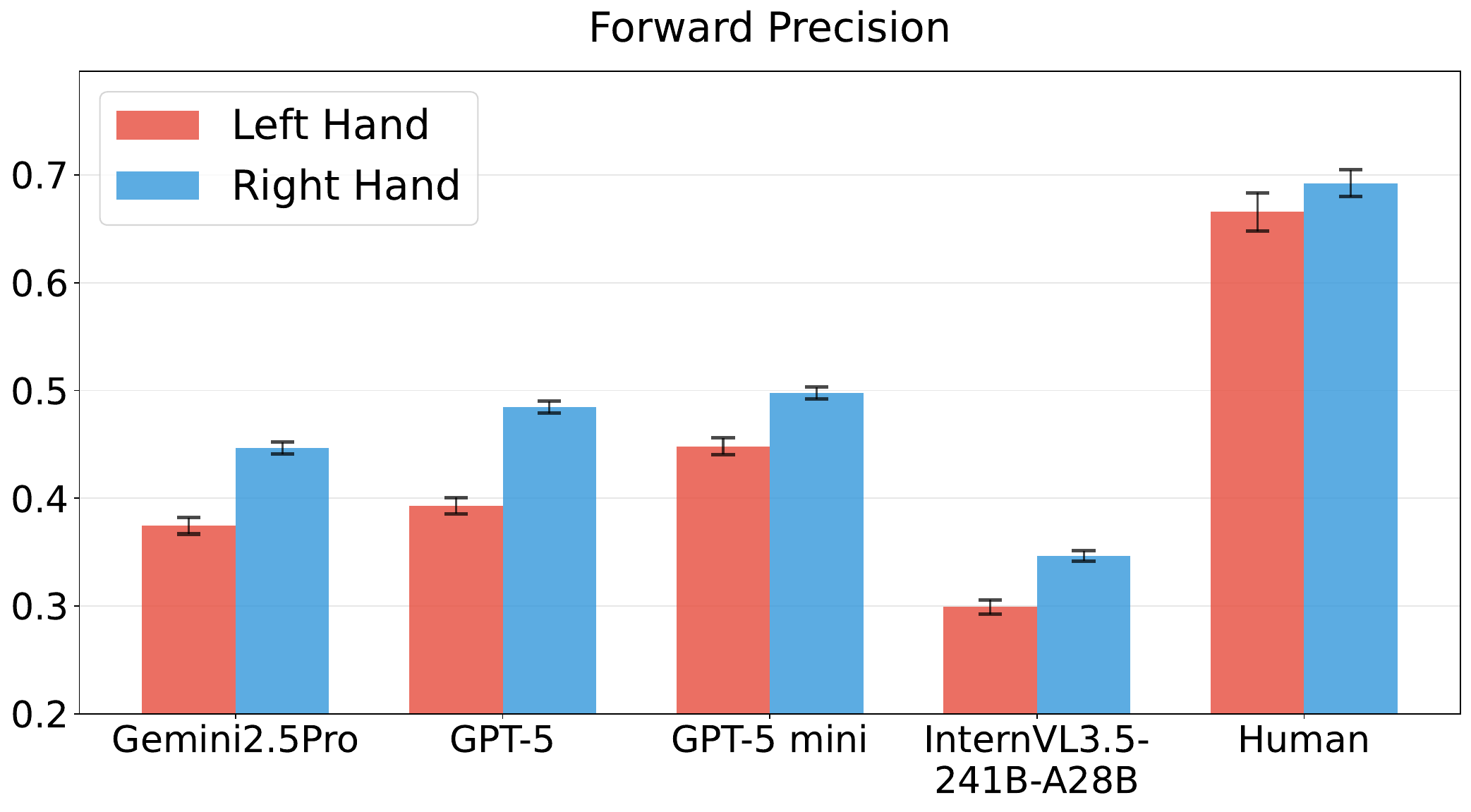}
    \caption{}
    \label{fig:precision_forward}
  \end{subfigure}\hfill
  \begin{subfigure}[t]{0.5\linewidth}
    \centering
    \includegraphics[width=\linewidth]{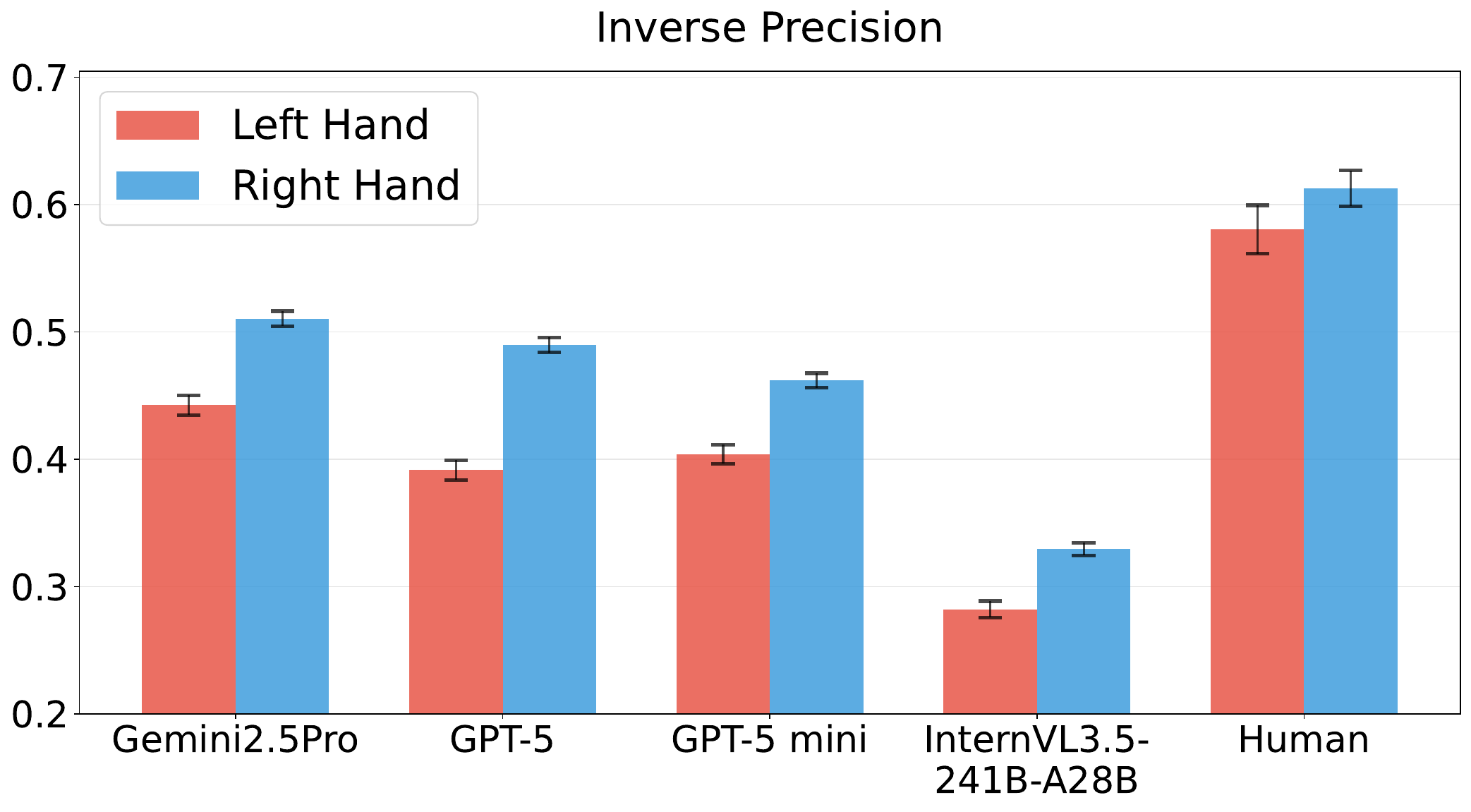}
    \caption{}
    \label{fig:precision_inverse}
  \end{subfigure}

    \caption{Precision of left/right hand related components prediction in (a) forward and (b) inverse tasks, with models Gemini2.5Pro, GPT-5, GPT-5 mini, InternVL3.5-241B-A28B, and Human. Error bars indicate the standard error (SE).}
  \label{fig:precision_fi}
\end{figure}

\begin{figure}[htbp]
  \centering
  
  \begin{subfigure}[t]{0.5\linewidth}
    \centering
    \includegraphics[width=\linewidth,page=1]{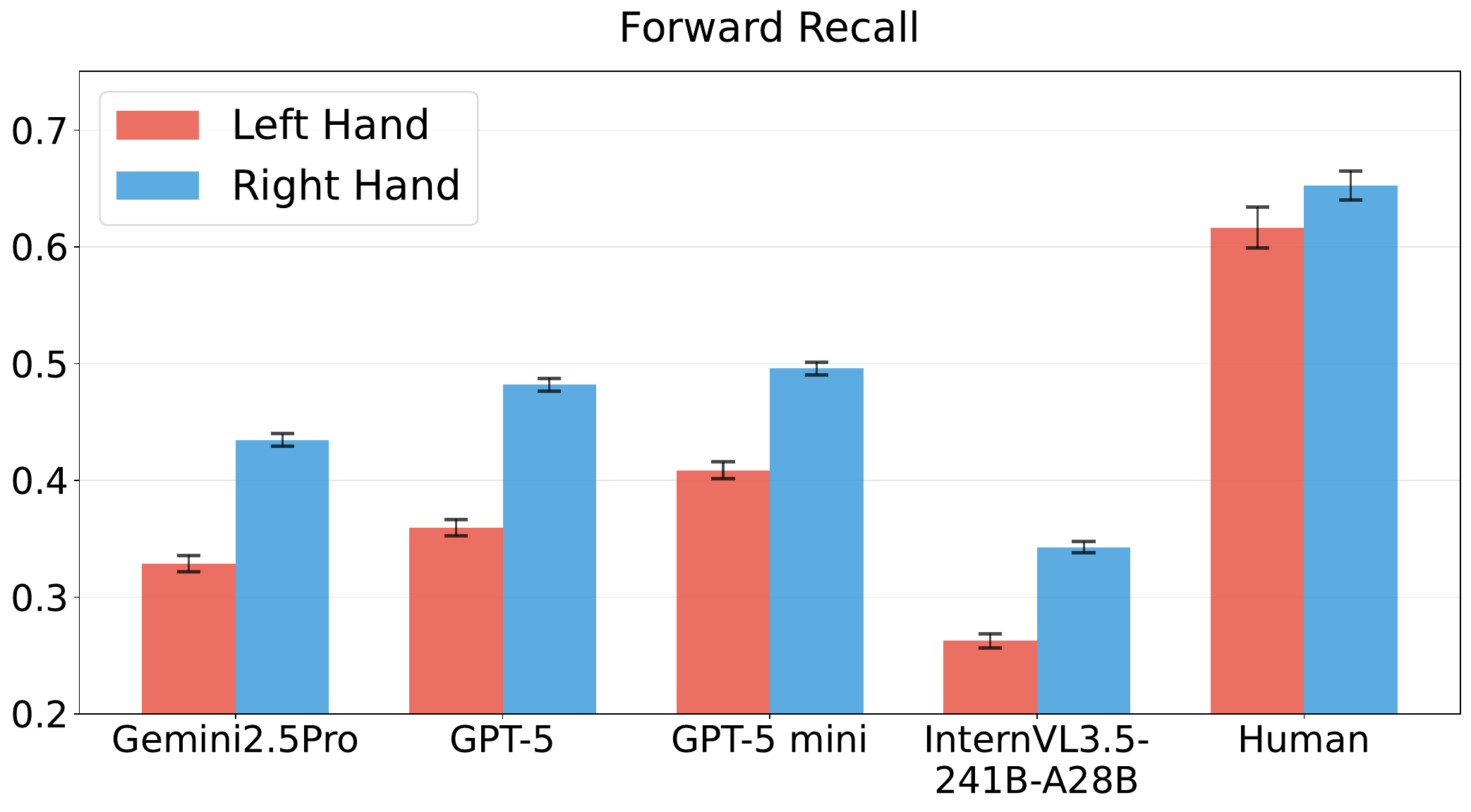}
    \caption{}
    \label{fig:recall_forward}
  \end{subfigure}\hfill
  \begin{subfigure}[t]{0.5\linewidth}
    \centering
    \includegraphics[width=\linewidth]{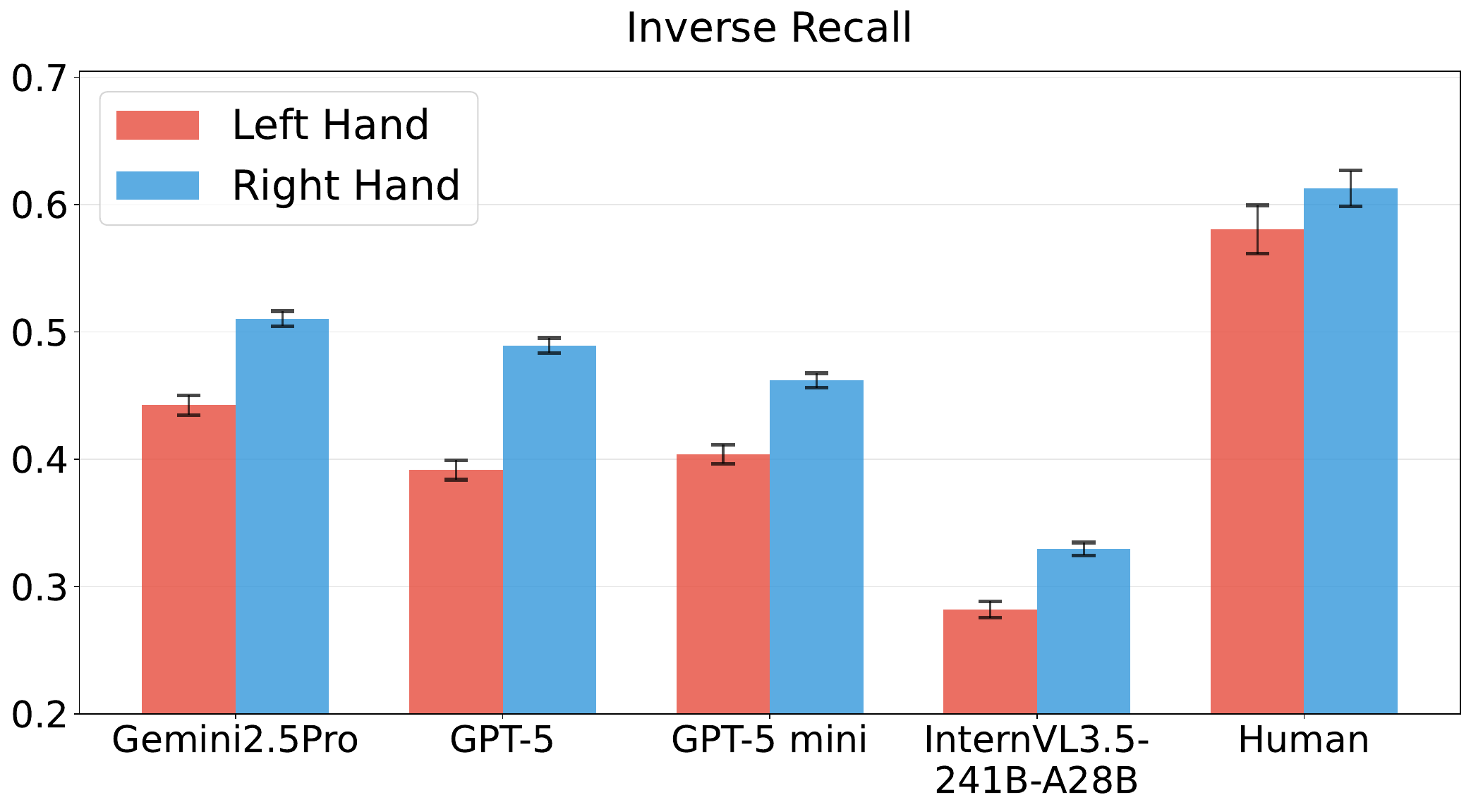}
    \caption{}
    \label{fig:recall_inverse}
  \end{subfigure}

    \caption{Recall of left/right hand related components prediction in forward task, with models Gemini2.5Pro, GPT-5, GPT-5 mini, InternVL3.5-241B-A28B and Human. Error bars indicate the standard error (SE).}
  \label{fig:recall_fi}
\end{figure}

\begin{figure}[htbp]
  \centering
  
  \begin{subfigure}[t]{0.5\linewidth}
    \centering
    \includegraphics[width=\linewidth,page=1]{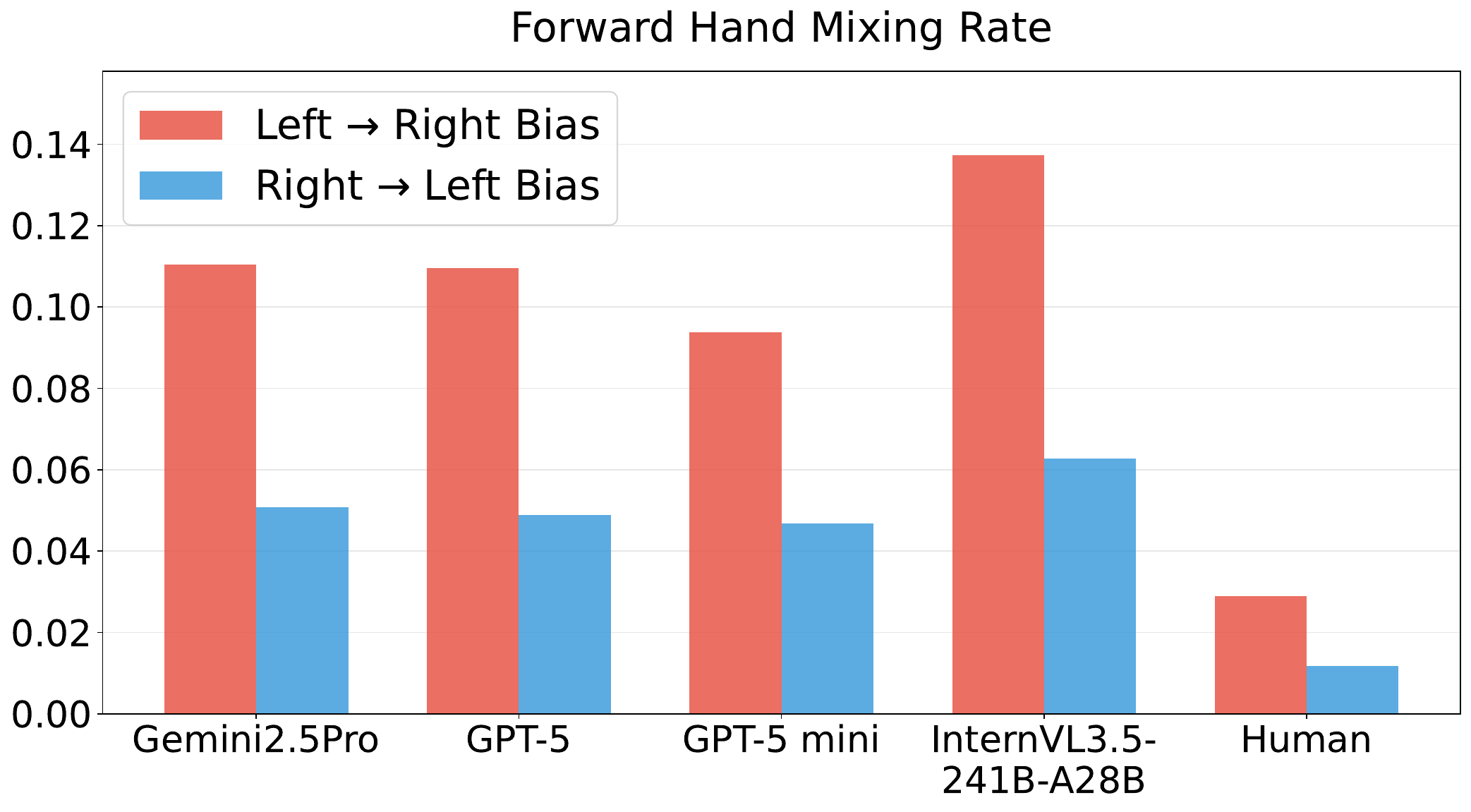}
    \caption{}
    \label{fig:full_mixing_forward}
  \end{subfigure}\hfill
  \begin{subfigure}[t]{0.5\linewidth}
    \centering
    \includegraphics[width=\linewidth]{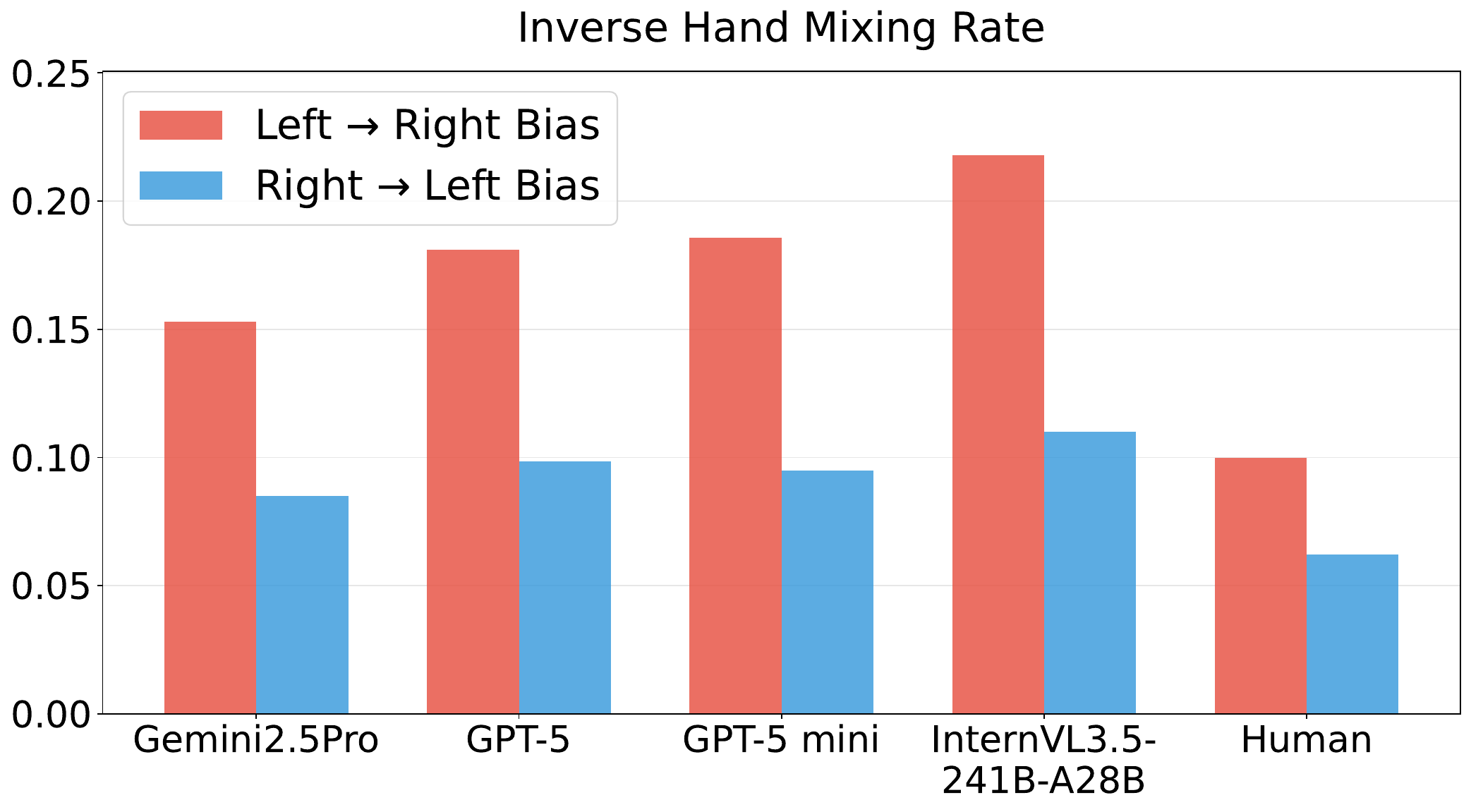}
    \caption{}
    \label{fig:full_mixing_inv}
  \end{subfigure}

    \caption{Hand-mixing rate, i.e.the ratio of left/right hand-mixing to all ground truth left/right and components in (a) forward and (b) inverse task, with models Gemini2.5Pro, GPT-5, GPT-5 mini, InternVL3.5-241B-A28B and Human. Error bars indicate the standard error (SE).}
  \label{fig:full_mixing_fi}
\end{figure}

In both humans and models, and across both task types, right-hand precision and recall systematically exceed those of the left (Figures~\ref{fig:precision_fi}, \ref{fig:recall_fi}). Furthermore, left-to-right mixing rate (ground-truth left-hand components wrongly predicted as right-hand ones) substantially exceeds the reverse (Figure~\ref{fig:full_mixing_fi}).

\section{Error Analysis}
\label{app:03_error_analysis}
\subsection{Methodology for Error Calculations}

\paragraph{Signature Modeling from Scene Graph-level Differences}

For error analysis, it is found hard to recognize predicate-level or semantic-level errors through natural language-based actions, i.e. visible differences between consecutive states. Hence, we parse the raw (natural language) action predicates as a signature into a sequence of unique state-change signature $(a_0^{\mathrm{sig}}, a_{1}^{sig}, \dots)$.

\[
c_i ::= (\gamma, e_1, \rho, e_2) 
   \;\mid\; (\gamma, e, \rho) 
   \;\mid\; (\mathrm{transition}, e, \rho_{\mathrm{from}} \rightarrow \rho_{\mathrm{to}}), 
\quad \gamma \in \{\mathrm{add}, \mathrm{remove}\}
\]

To further structure these signatures, each signature \(a_i^{\mathrm{sig}}\) is then modeled as a finite set of components $\{c_1, c_2, \dots\}$. Each component $c_i$ represents an atomic unit of state change. We distinguish three categories of components: edge components (addition or removal \(\gamma\) of predicates \(\rho\) between two entities \(e_1\) and \(e_2\)), node components (addition or removal \(\gamma\) of the predicate \(\rho\) of entity \(e\)), and node transition components (transition from the previous predicate \(\rho_{from}\) to new predicate \(\rho_{to}\) of an entity \(e\)).  

\paragraph{Error Modeling from Signatures}

We categorize errors from two perspectives: structural and semantic. Structural errors concern the form of actions and include entity substitution (object replacement), predicate substitution (relation/attribute replacement), polarity inversion (add, remove or transition), omission, and hallucination. Semantic errors concern interpretation and are grouped into spatial relations (misplaced object positions), functional states (incorrect functionalities or status), material states (wrong physical properties), and agent interaction (misattributed agent actions).

Both perspectives are based on comparing component sets of paired ground-truth and predicted signatures. For each pair, we compute set-level differences and classify components into missing (in ground truth only), matched (in both), and hallucinated (in prediction only). To support this categorization, we preprocess the signature dataset into structured data with these three groups of components, as outlined in Algorithm~\ref{alg:error-calc-parsing}.

To categorize structural errors, we define criteria for each component type (edge, node, transition node). Entity Substitution occurs when entities differ while other fields match; Predicate Substitution when the predicate differs; and Polarity Inversion when only the operation (add/remove) differs. After pairwise classification, remaining unmatched ground-truth components are categorized to Omission, and unmatched predicted components as Hallucination.

After structural error categorization, each component is further labeled by semantic error type: Spatial Relations, Functional States, Material States, or Agent Interactions. Labeling uses a predefined mapping table that links all observed predicates to their semantic categories. When a component contains a listed predicate, the table is consulted to assign its semantic label. The overall workflow of error detection and categorization is illustrated in Algorithm~\ref{alg:error-calc-categorization}.

\paragraph{Handedness Asymmetry Error Modeling \label{app_4_1_3:handedness}}

To systematically capture handedness asymmetry, we compute for left- and right-hand components: precision (correct matches over predicted), recall (correct matches over ground truth), and the hand-mixing rate (the fraction of ground-truth left-hand components predicted as right, or vice versa).

For computing the hand-mixing rate, we define left–right mixing at the level of each signature-level difference (with missing, matched, and hallucinated components). If the missing set contains left- (or right-) hand usage, while the hallucinated set lacks the same hand but includes the opposite one, then all missing components involving that hand are counted as left-to-right (or right-to-left) mixing, as outlined in Algorithm~\ref{alg:hand-confusion-dataset}.

\subsection{Structural Error Analysis}
\begin{figure}[t]
  \centering
  
  \begin{subfigure}[t]{0.5\linewidth}
    \centering
    \includegraphics[width=\linewidth,page=1]{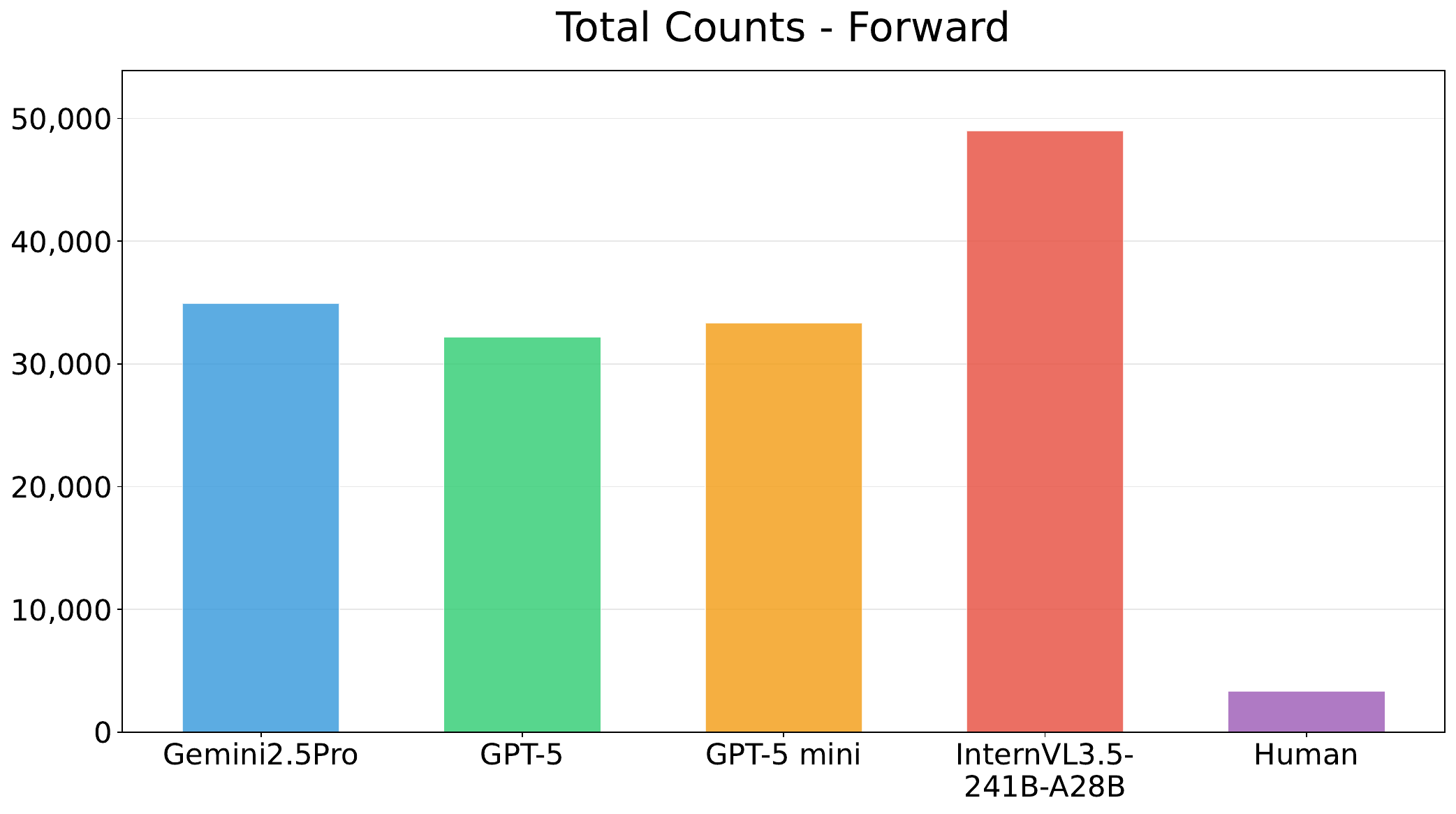}
    \caption{}
    \label{fig:total_f}
  \end{subfigure}\hfill
  \begin{subfigure}[t]{0.5\linewidth}
    \centering
    \includegraphics[width=\linewidth]{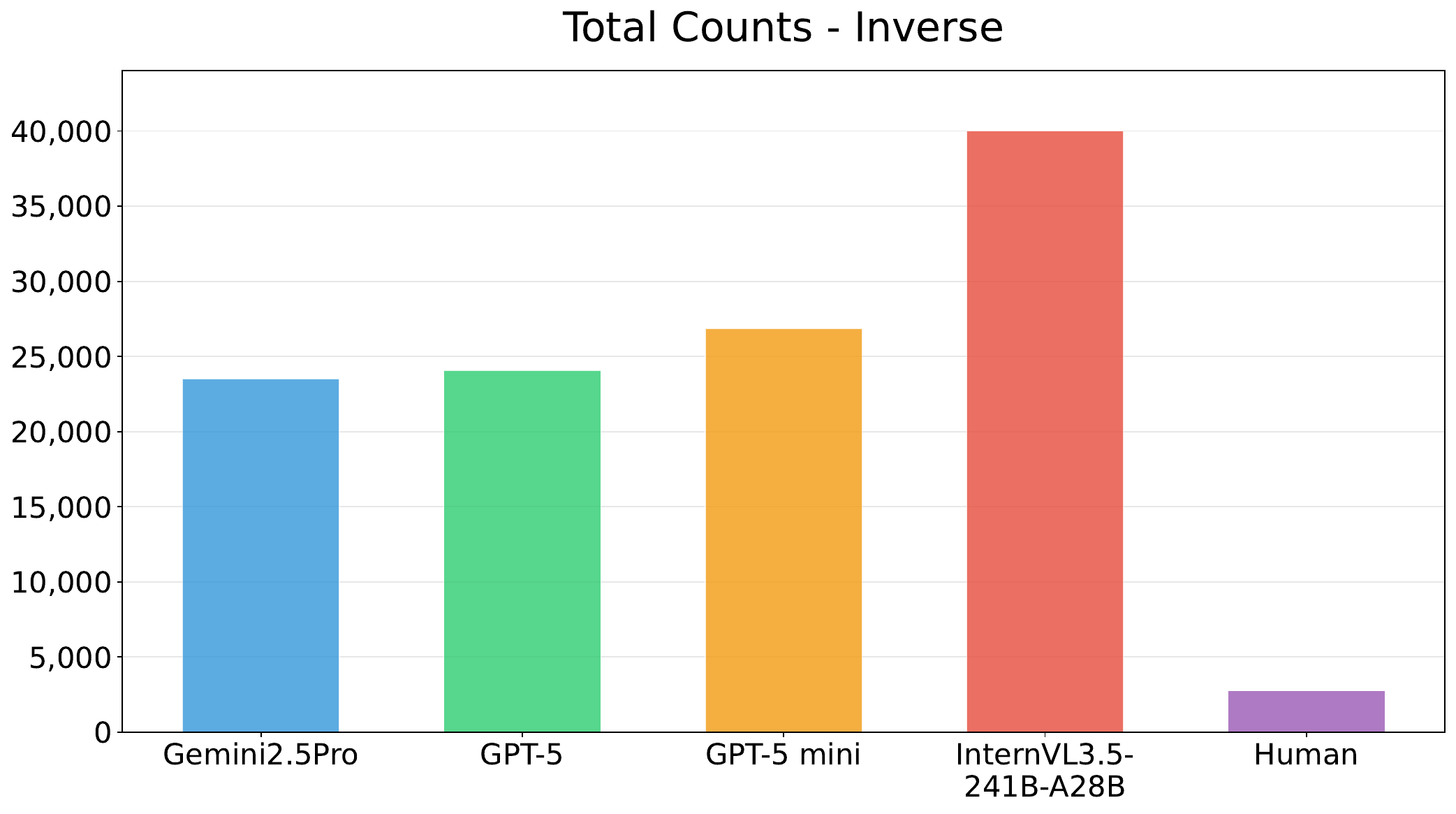}
    \caption{}
    \label{total_i}
  \end{subfigure}

    \caption{The amount of total errors made by Gemini2.5Pro, GPT-5, GPT-5 mini, InternVL3.5-241B-A28B, and Human, under (a) forward tasks and (b) inverse tasks.}
  \label{fig:total_fi}
\end{figure}

We compared error patterns in forward and inverse tasks across Gemini-2.5 Pro, GPT-5, GPT-5 mini, InternVL-3.5-241B-A28B, and human predictions (Figures~\ref{fig:total_fi}). 

\begin{figure}[htbp]
    \centering
    \includegraphics[width=\linewidth]{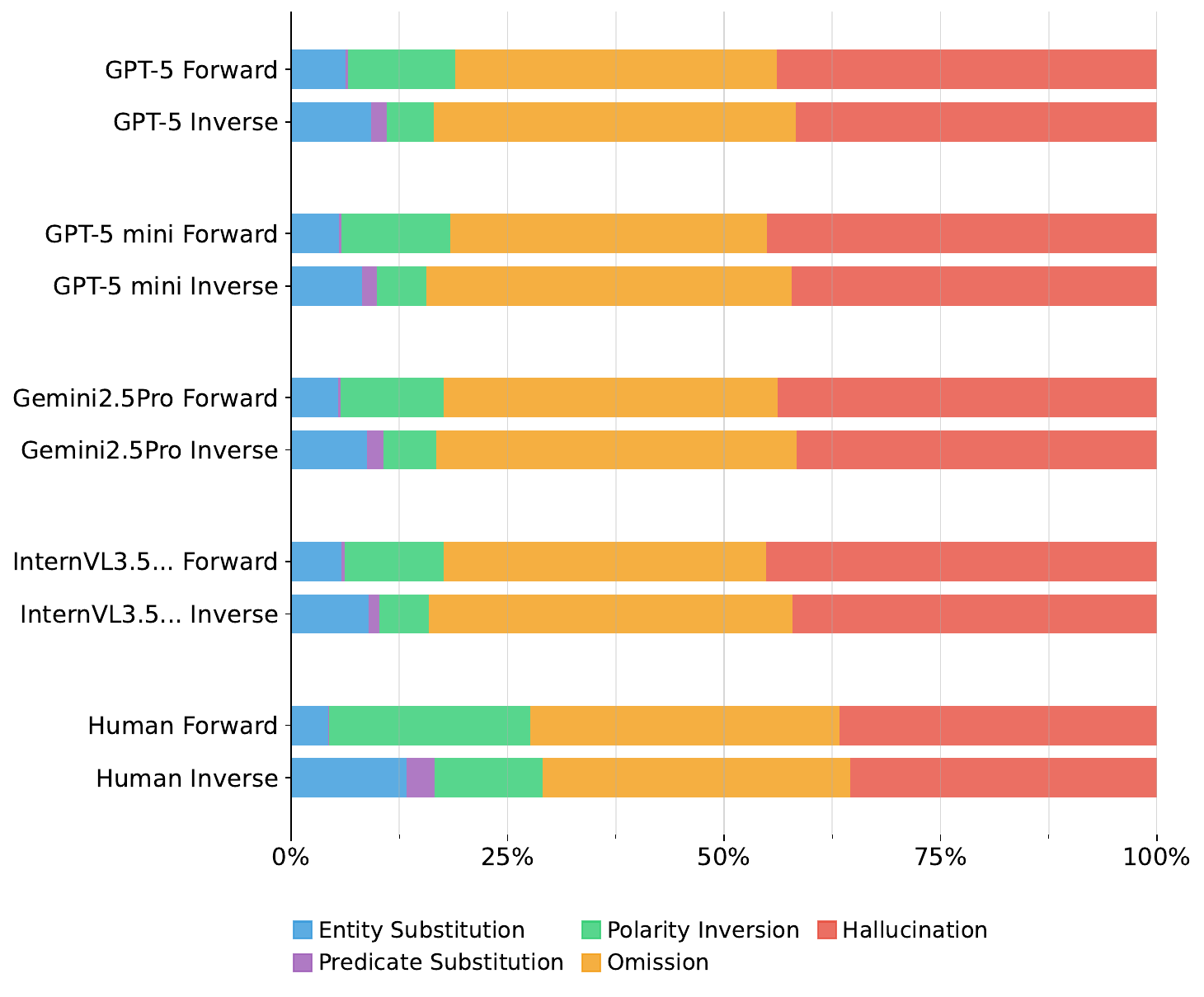}
    \caption{The structural error distributions of typical LLMs (GPT-5, GPT-5 mini, Gemini2.5Pro and InternVL3.5-241B-A28B (referred as InternVL3.5... in figure)) and Human-level prediction in both forward and inverse tasks.}
    \label{fig:structural_errors_statistics}
\end{figure}

\begin{figure}[t]
  \centering
  
  \begin{subfigure}[t]{0.5\linewidth}
    \centering
    \includegraphics[width=\linewidth,page=1]{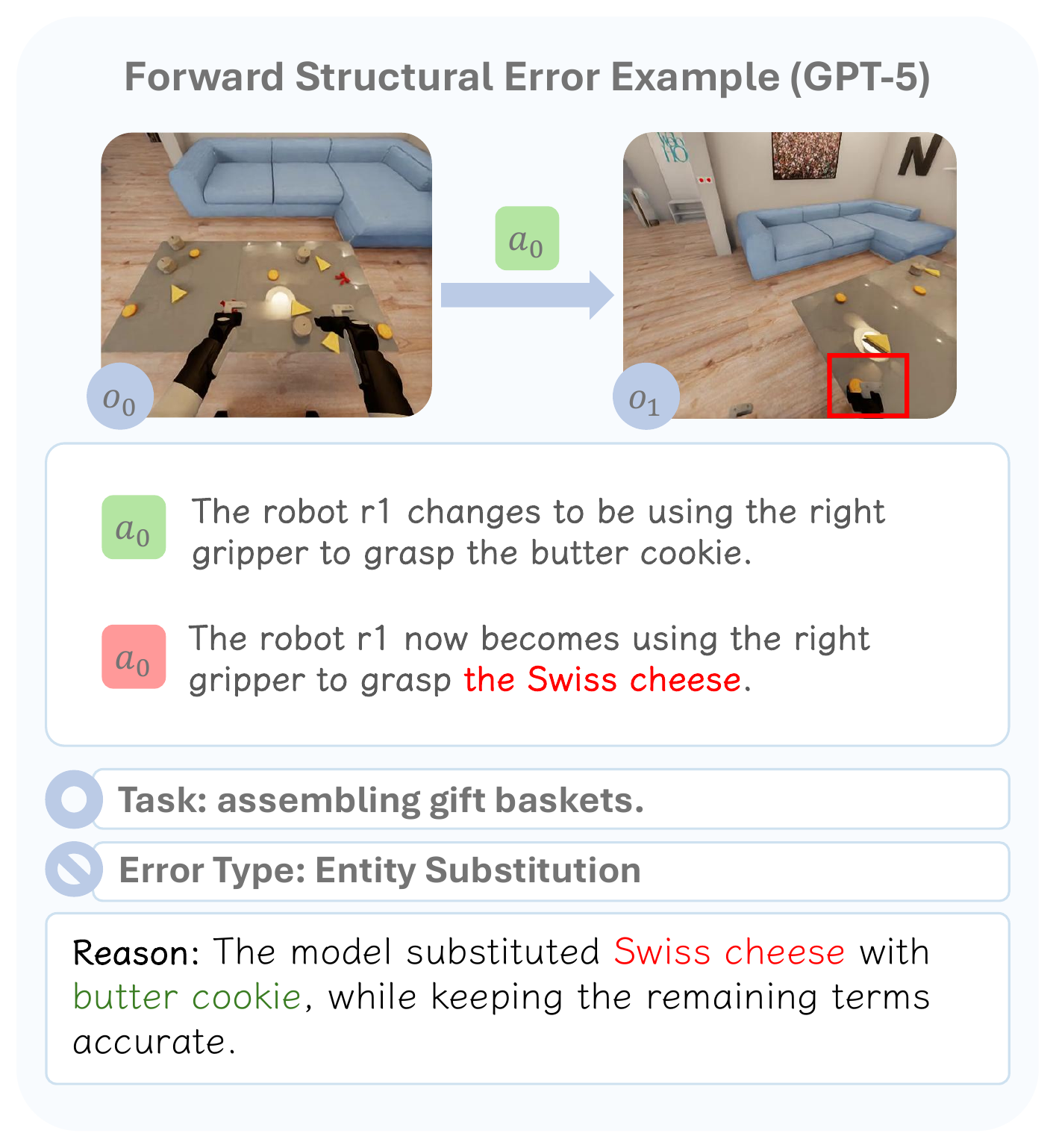}
    \label{fig:f_st_ES}
  \end{subfigure}\hfill
  \begin{subfigure}[t]{0.5\linewidth}
    \centering
    \includegraphics[width=\linewidth]{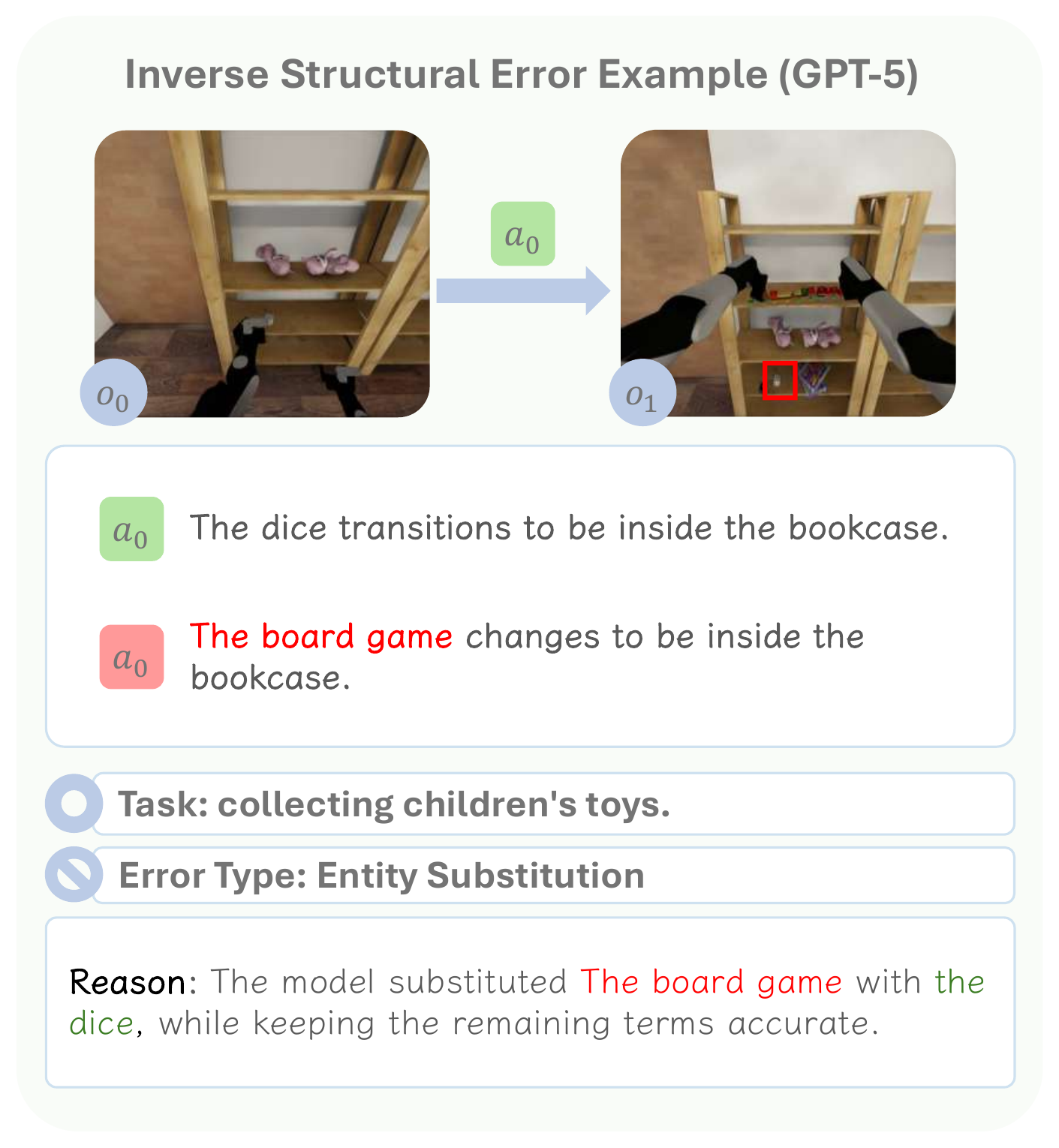}
    \label{i_st_ES}
  \end{subfigure}

    \caption{Example of structural error \textbf{Entity Substitution} by GPT-5 under forward and inverse tasks.}
  \label{fig:fi_st_ES}
\end{figure}

\begin{figure}[t]
  \centering
  \begin{subfigure}[t]{0.5\linewidth}
    \centering
    \includegraphics[width=\linewidth,page=1]{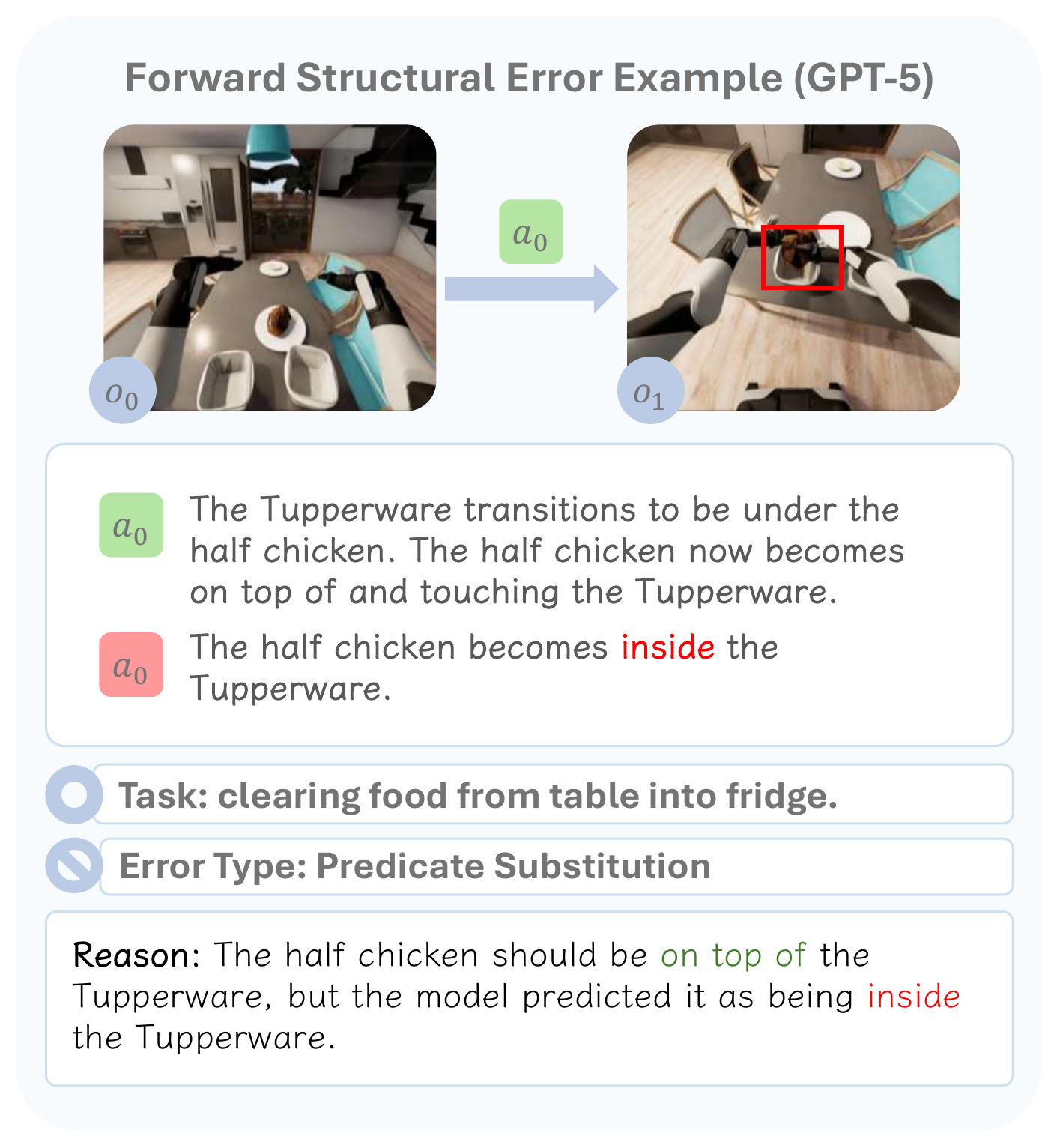}
    \label{f_st_PS}
  \end{subfigure}\hfill
  \begin{subfigure}[t]{0.5\linewidth}
    \centering
    \includegraphics[width=\linewidth]{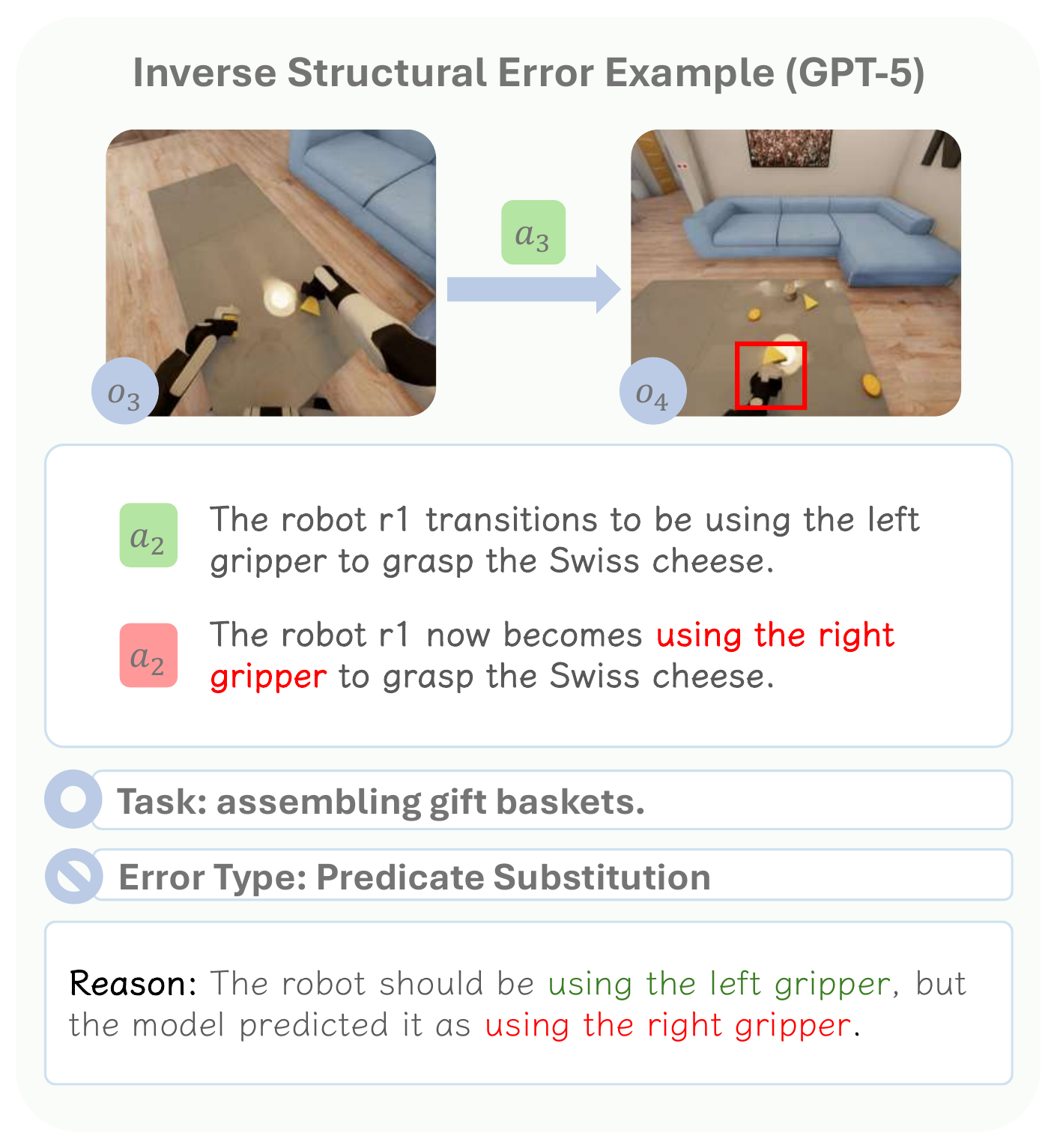}
    \label{i_st_PS}
  \end{subfigure}
  
  \caption{Example of structural error \textbf{Predicate Substitution} by GPT-5 under forward and inverse tasks.}
  \label{fig:fi_st_PS}
\end{figure}

\begin{figure}[t]
  \centering
  
  \begin{subfigure}[t]{0.5\linewidth}
    \centering
    \includegraphics[width=\linewidth,page=1]{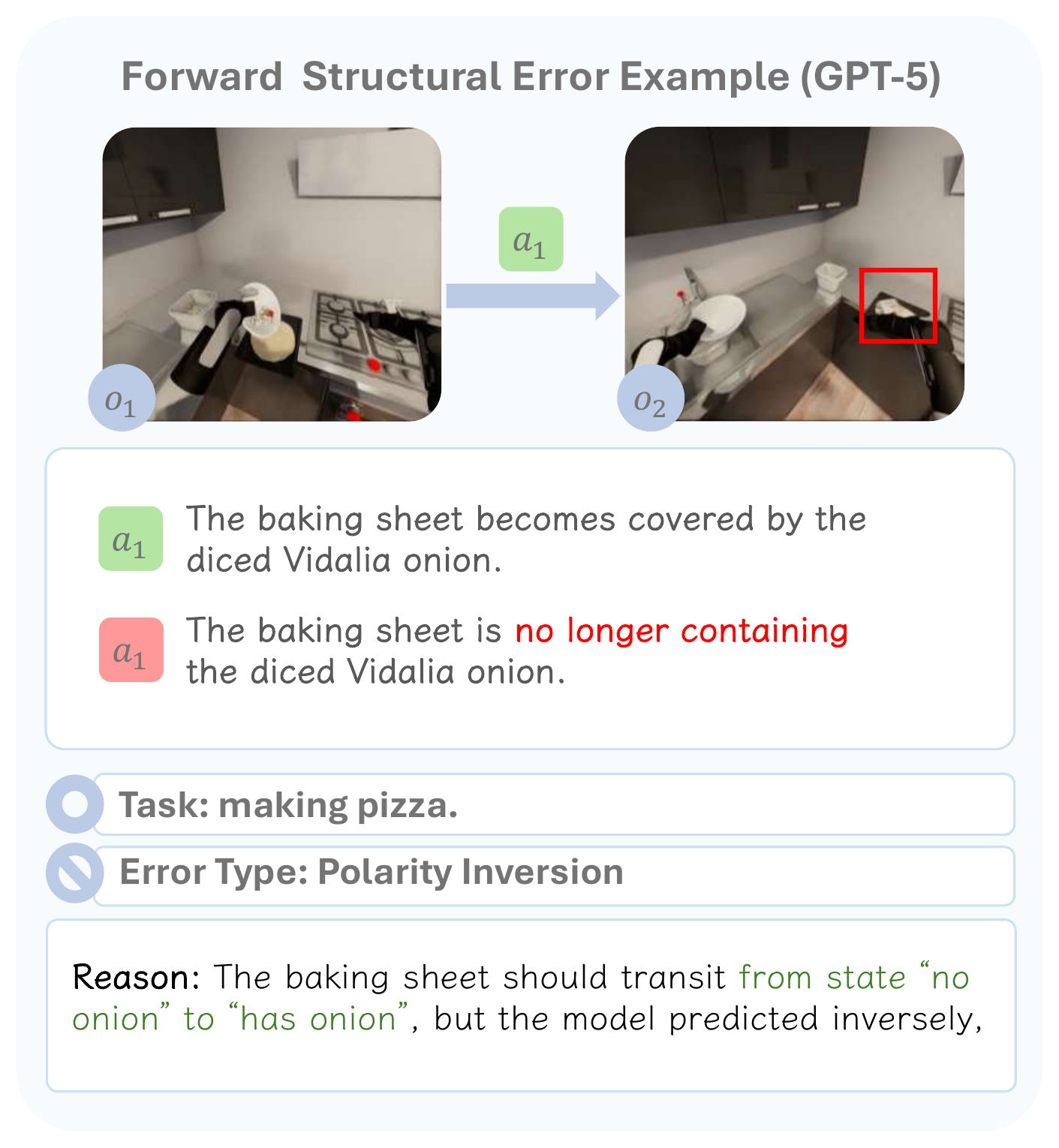}
    \label{fig:f_st_PI}
  \end{subfigure}\hfill
  \begin{subfigure}[t]{0.5\linewidth}
    \centering
    \includegraphics[width=\linewidth]{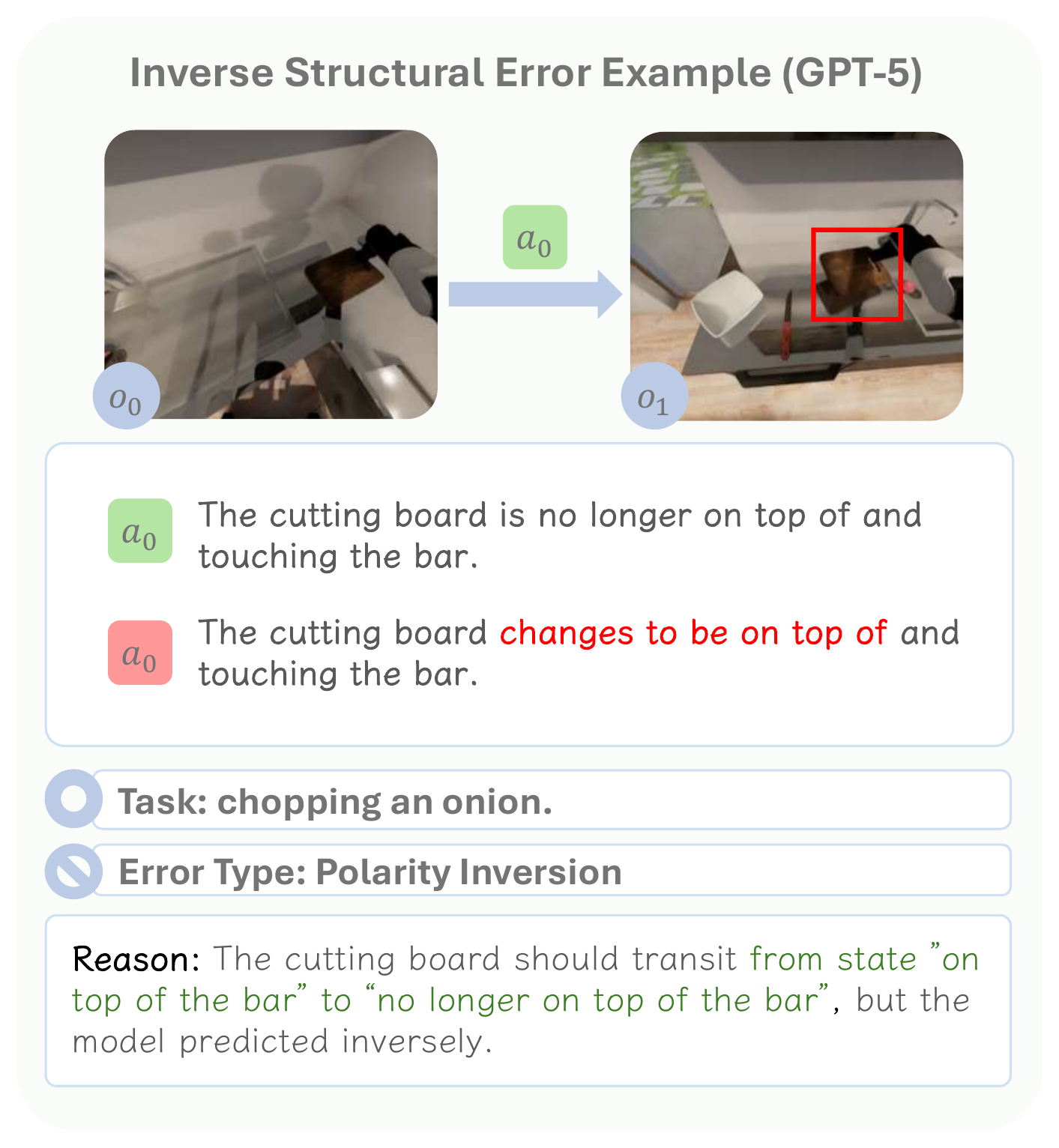}
    \label{i_st_PI}
  \end{subfigure}
  
  \caption{Example of structural error \textbf{Polarity Inversion} by GPT-5 under forward and inverse tasks.}
  \label{fig:fi_st_PI}
\end{figure}

\begin{figure}[t]
  \centering
  
  \begin{subfigure}[t]{0.5\linewidth}
    \centering
    \includegraphics[width=\linewidth,page=1]{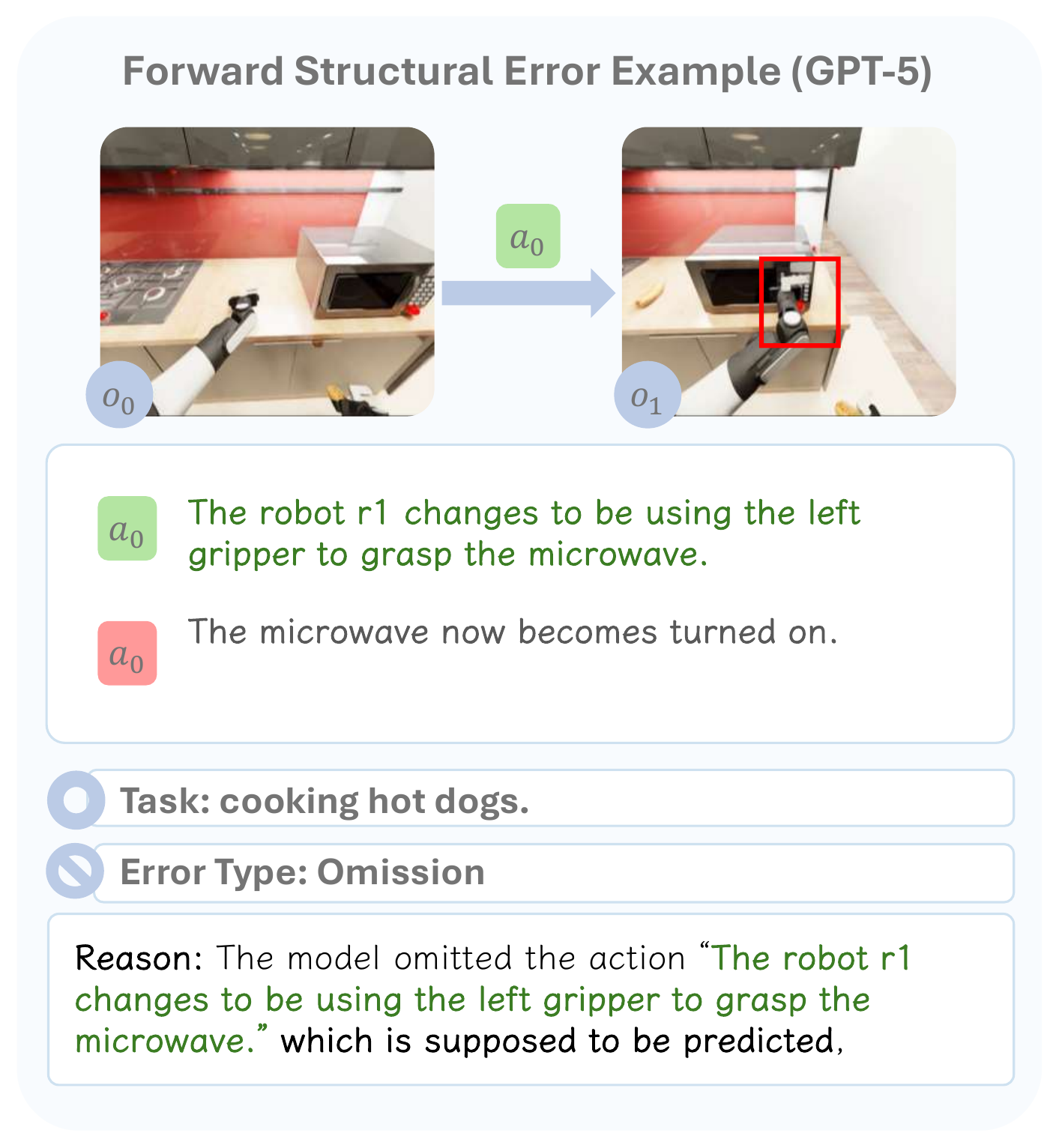}
    \label{f_st_OM}
  \end{subfigure}\hfill
  \begin{subfigure}[t]{0.5\linewidth}
    \centering
    \includegraphics[width=\linewidth]{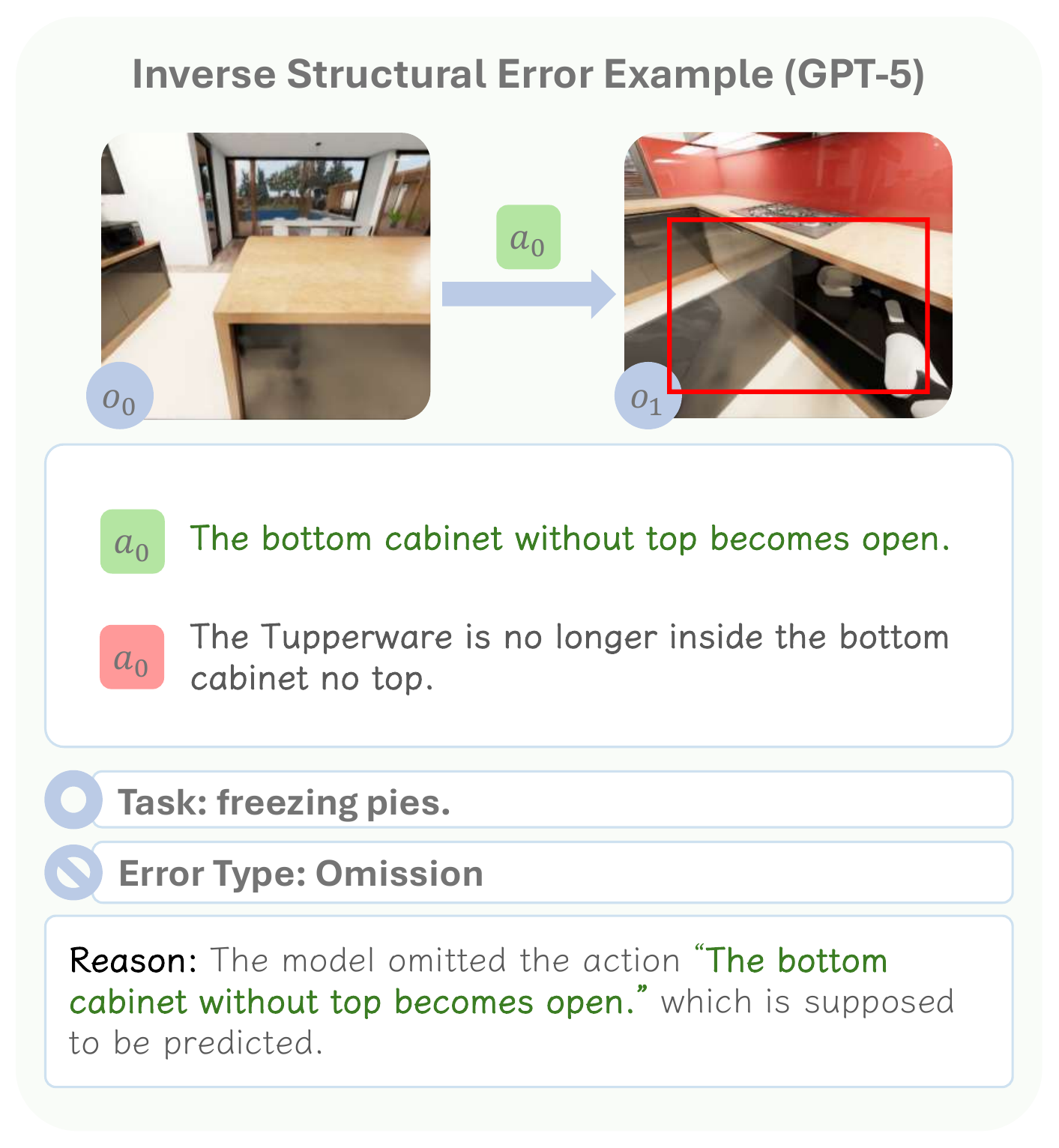}
    \label{i_st_OM}
  \end{subfigure}
  
  \caption{Example of structural error \textbf{Omission} by GPT-5 under forward and inverse tasks.}
  \label{fig:fi_st_OM}
\end{figure}

\begin{figure}[t]
  \centering

  \begin{subfigure}[t]{0.5\linewidth}
    \centering
    \includegraphics[width=\linewidth,page=1]{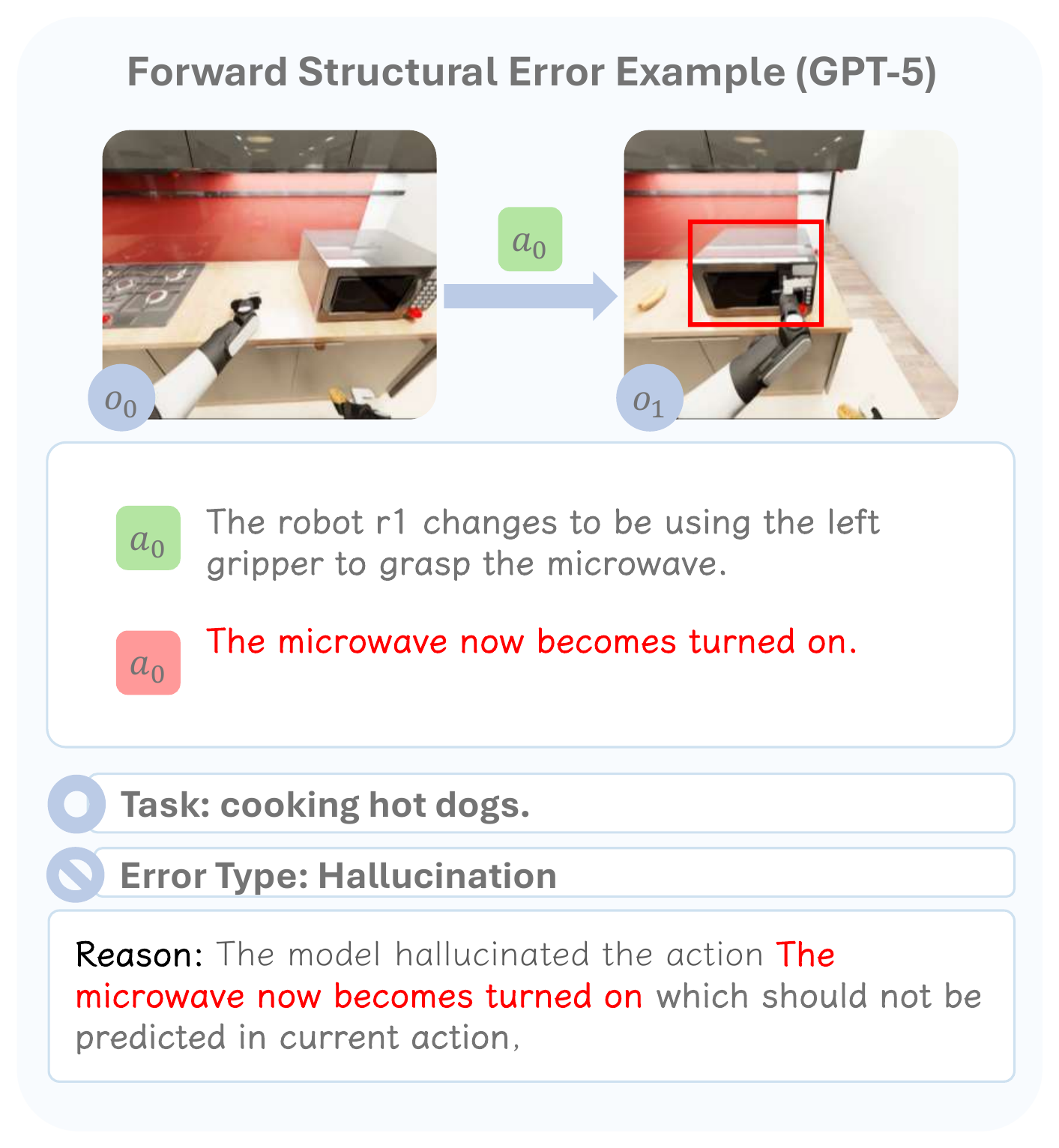}
    \label{f_st_HA}
  \end{subfigure}\hfill
  \begin{subfigure}[t]{0.5\linewidth}
    \centering
    \includegraphics[width=\linewidth]{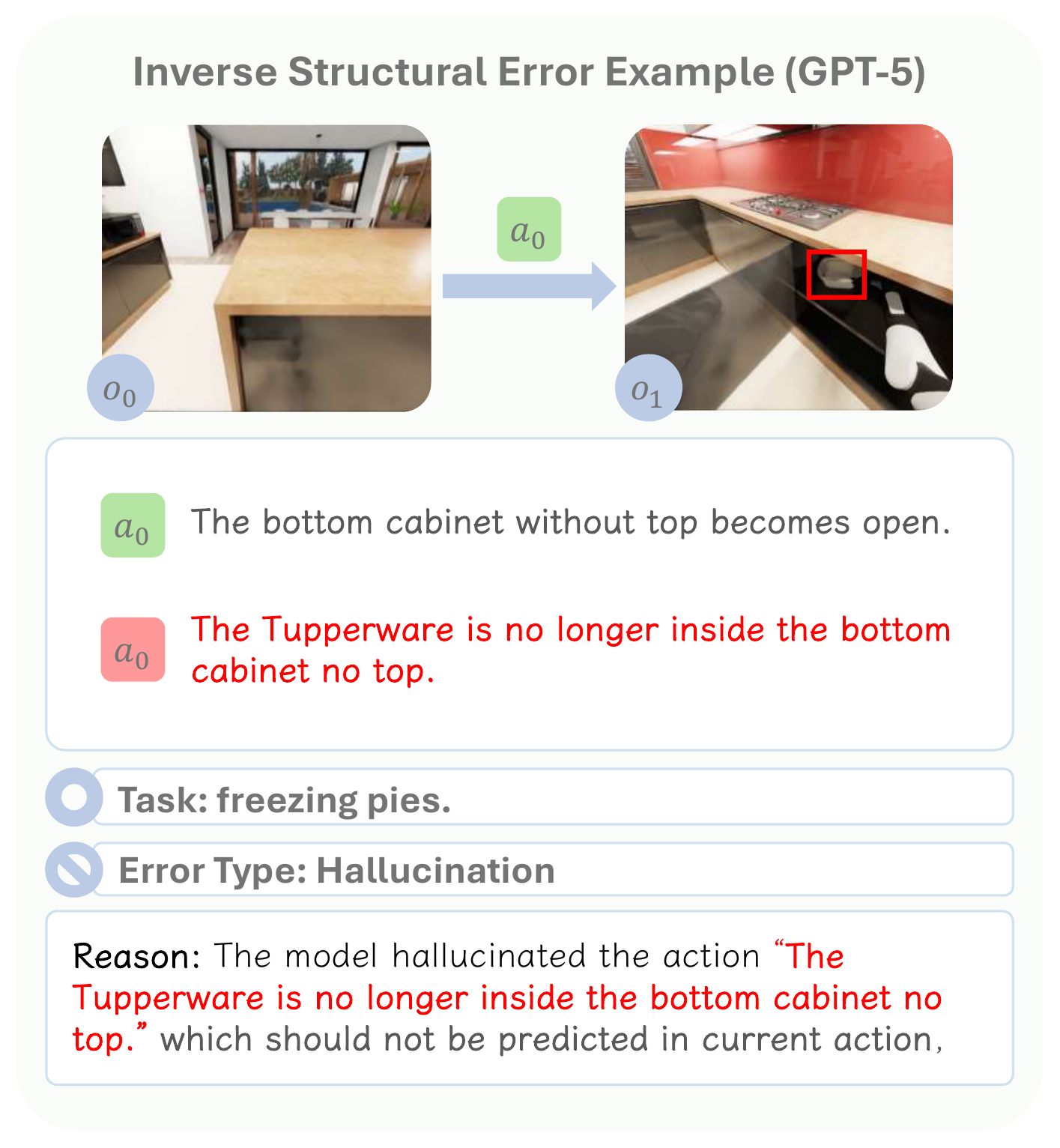}
    \label{i_st_HA}
  \end{subfigure}
  
  \caption{Example of structural error \textbf{Hallucination} by GPT-5 under forward and inverse tasks.}
  \label{fig:fi_st_HA}
\end{figure}

\subsection{Semantic Error Analysis}
\begin{figure}[htbp]
    \centering
    \includegraphics[width=\linewidth]{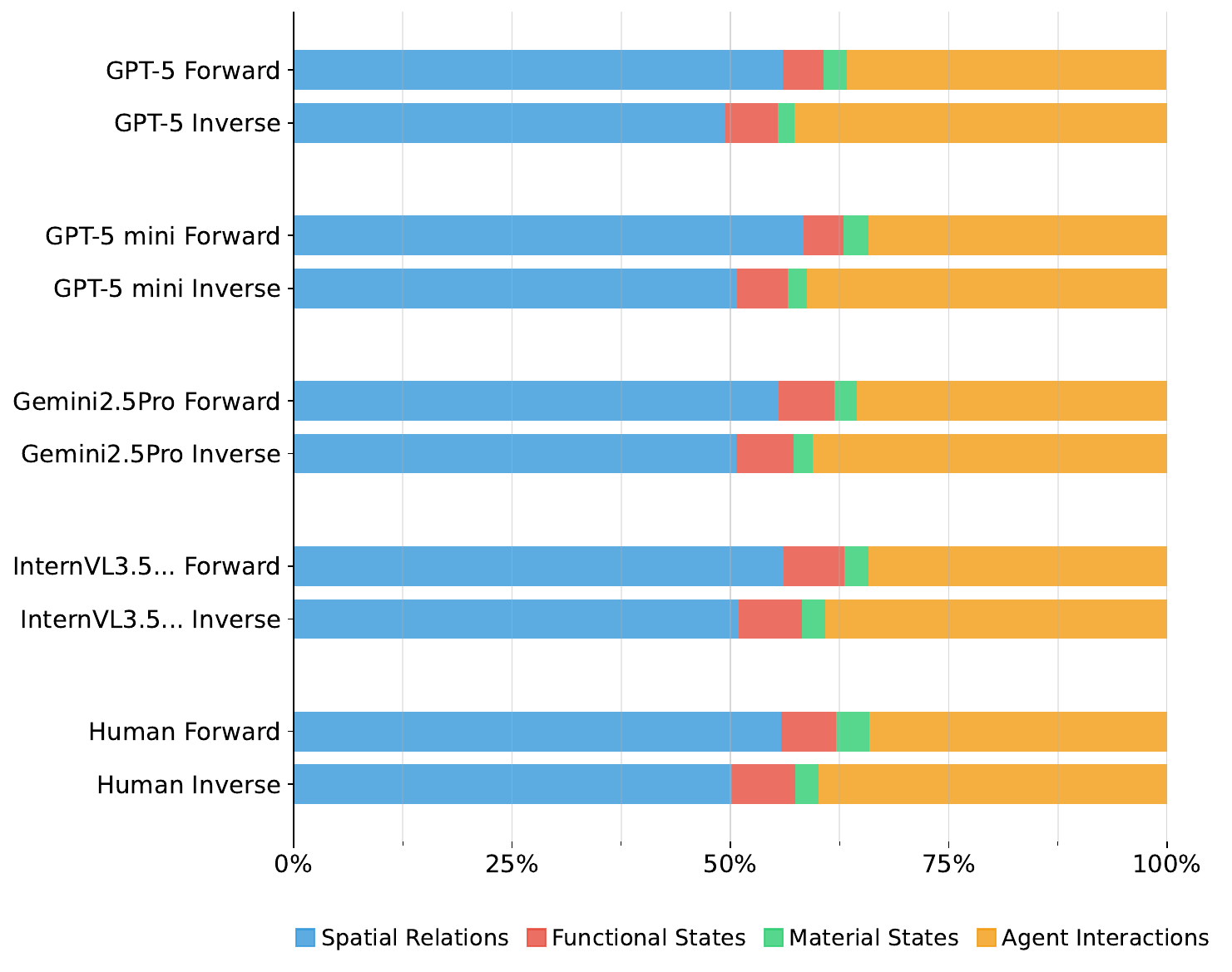}
    \caption{The semantic error distributions of typical LLMs (GPT-5, GPT-5 mini, Gemini2.5Pro and InternVL3.5-241B-A28B (referred as InternVL3.5... in figure)) and Human-level prediction in both forward and inverse tasks.}
    \label{fig:semantic_errors_statistics}
\end{figure}

\begin{figure}[t]
  \centering
  
  \begin{subfigure}[t]{0.5\linewidth}
    \centering
    \includegraphics[width=\linewidth,page=1]{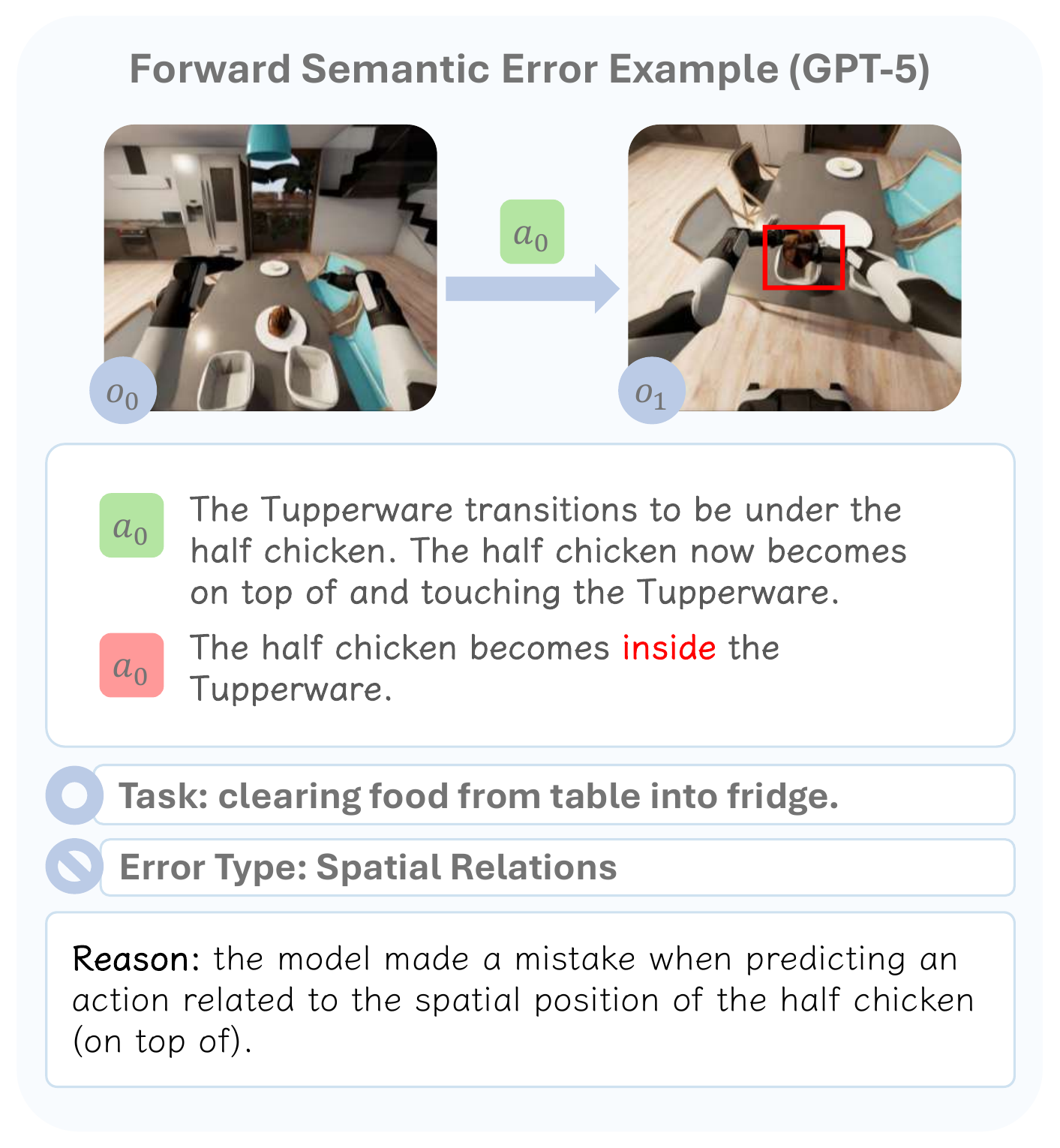}
    \label{fig:f_sem_SR}
  \end{subfigure}\hfill
  \begin{subfigure}[t]{0.5\linewidth}
    \centering
    \includegraphics[width=\linewidth]{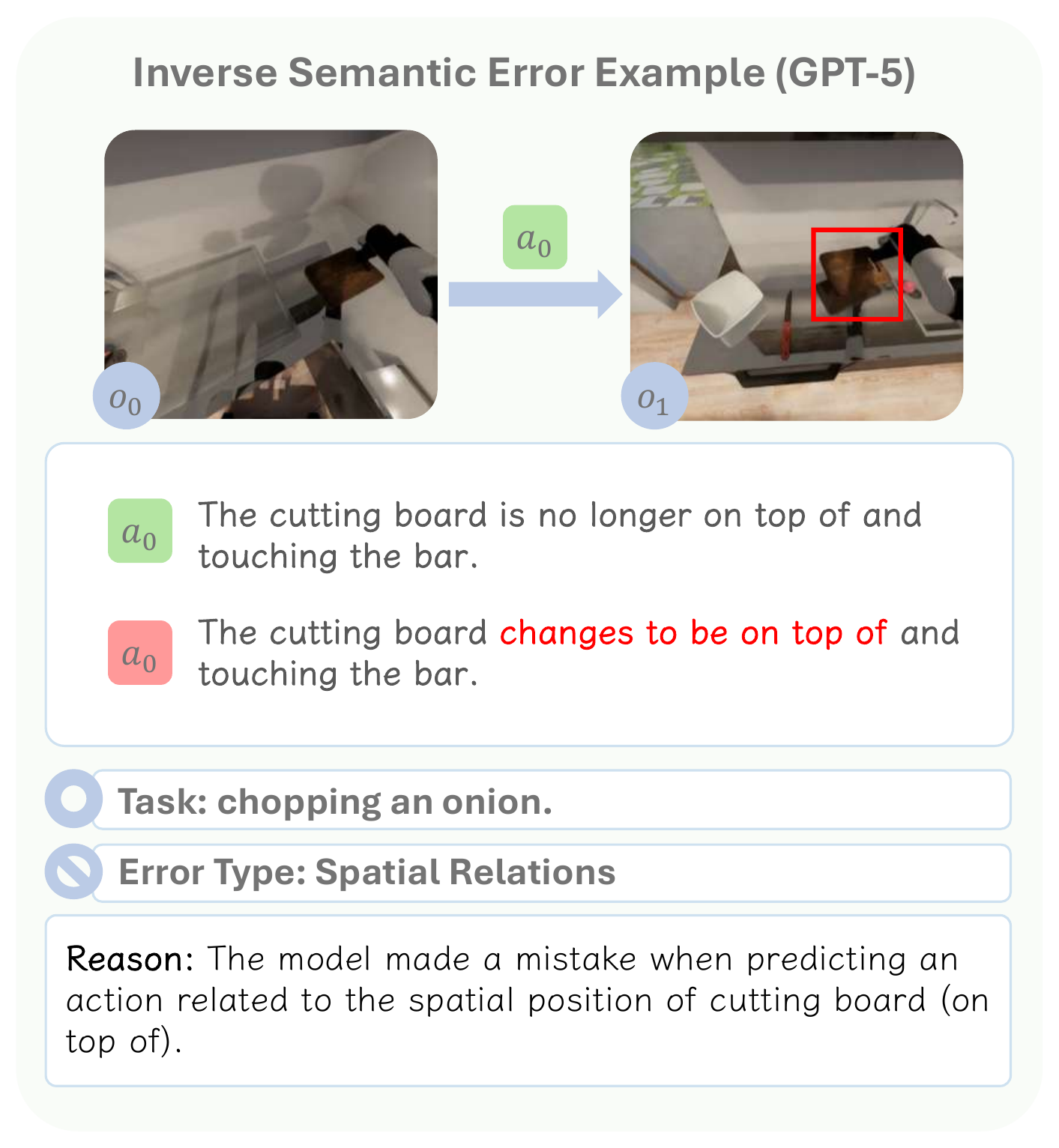}
    \label{i_sem_SR}
  \end{subfigure}

    \caption{Example of semantic error \textbf{Spatial Relations} by GPT-5 under forward and inverse tasks.}
  \label{fig:fi_sem_SR}
\end{figure}

\begin{figure}[t]
  \centering
  \begin{subfigure}[t]{0.5\linewidth}
    \centering
    \includegraphics[width=\linewidth,page=1]{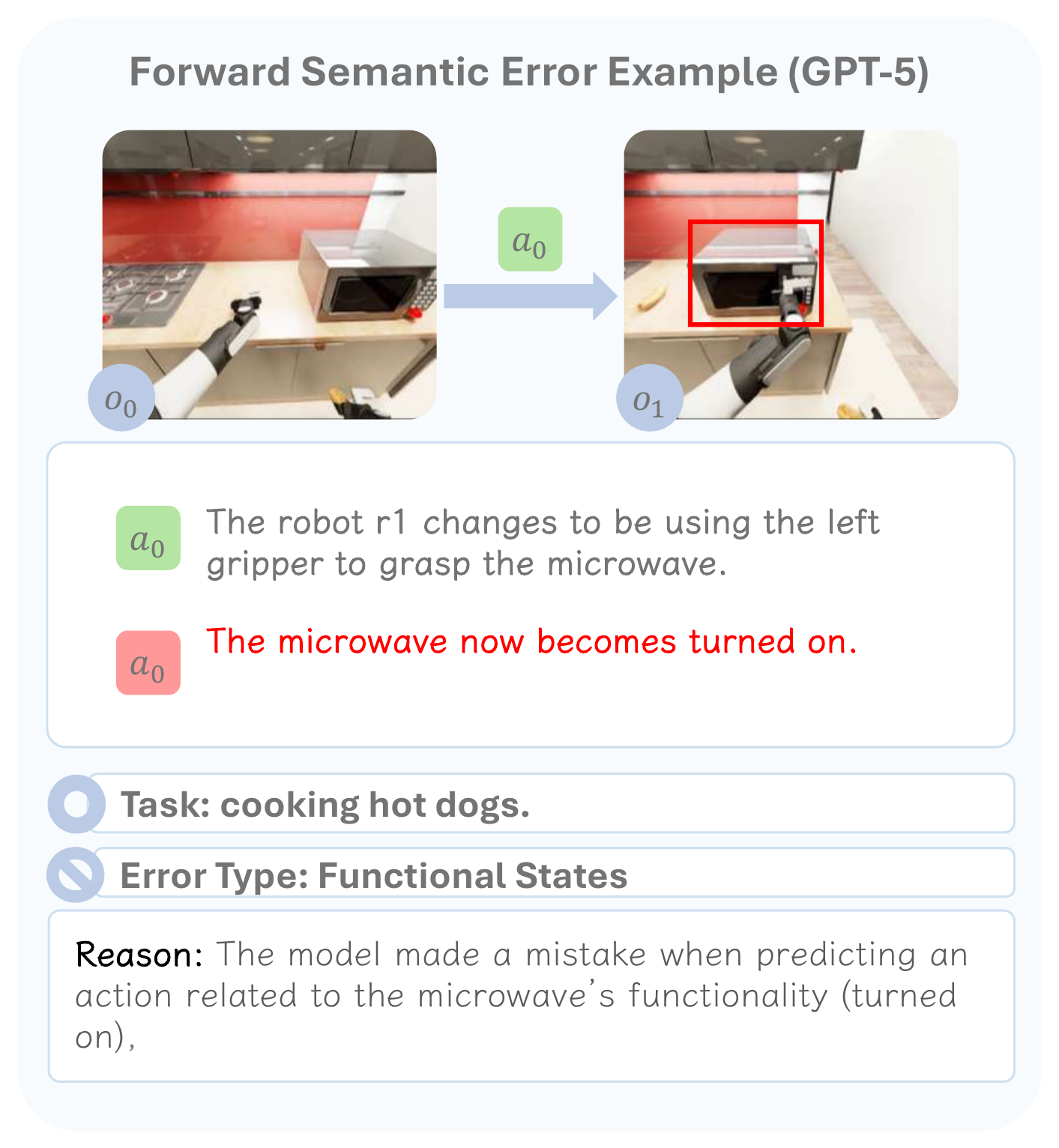}
    \label{f_sem_FS}
  \end{subfigure}\hfill
  \begin{subfigure}[t]{0.5\linewidth}
    \centering
    \includegraphics[width=\linewidth]{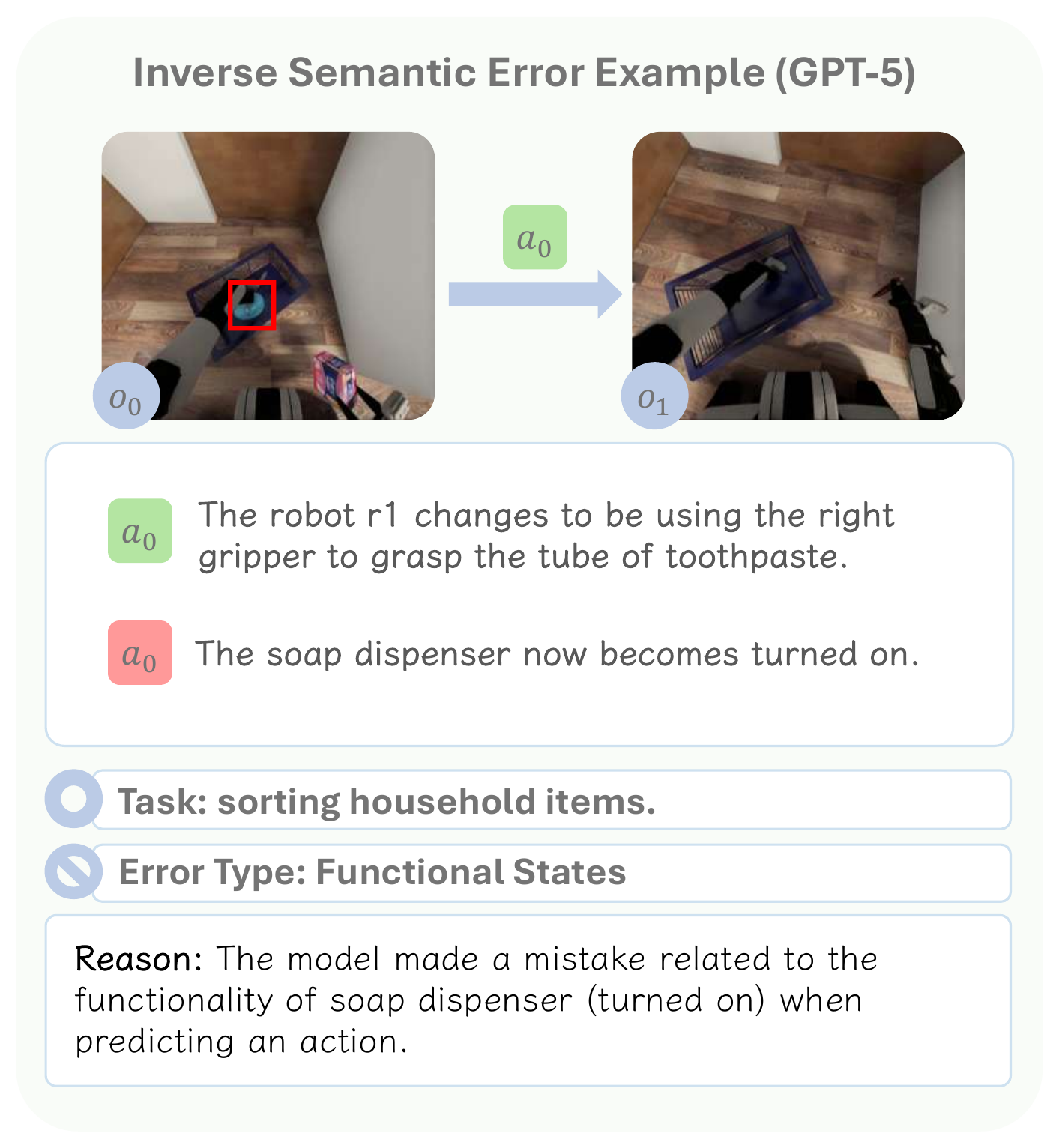}
    \label{i_sem_FS}
  \end{subfigure}
  
  \caption{Example of semantic error \textbf{Functional States} by GPT-5 under forward and inverse tasks.}
  \label{fig:fi_sem_FS}
\end{figure}

\begin{figure}[t]
  \centering
  
  \begin{subfigure}[t]{0.5\linewidth}
    \centering
    \includegraphics[width=\linewidth,page=1]{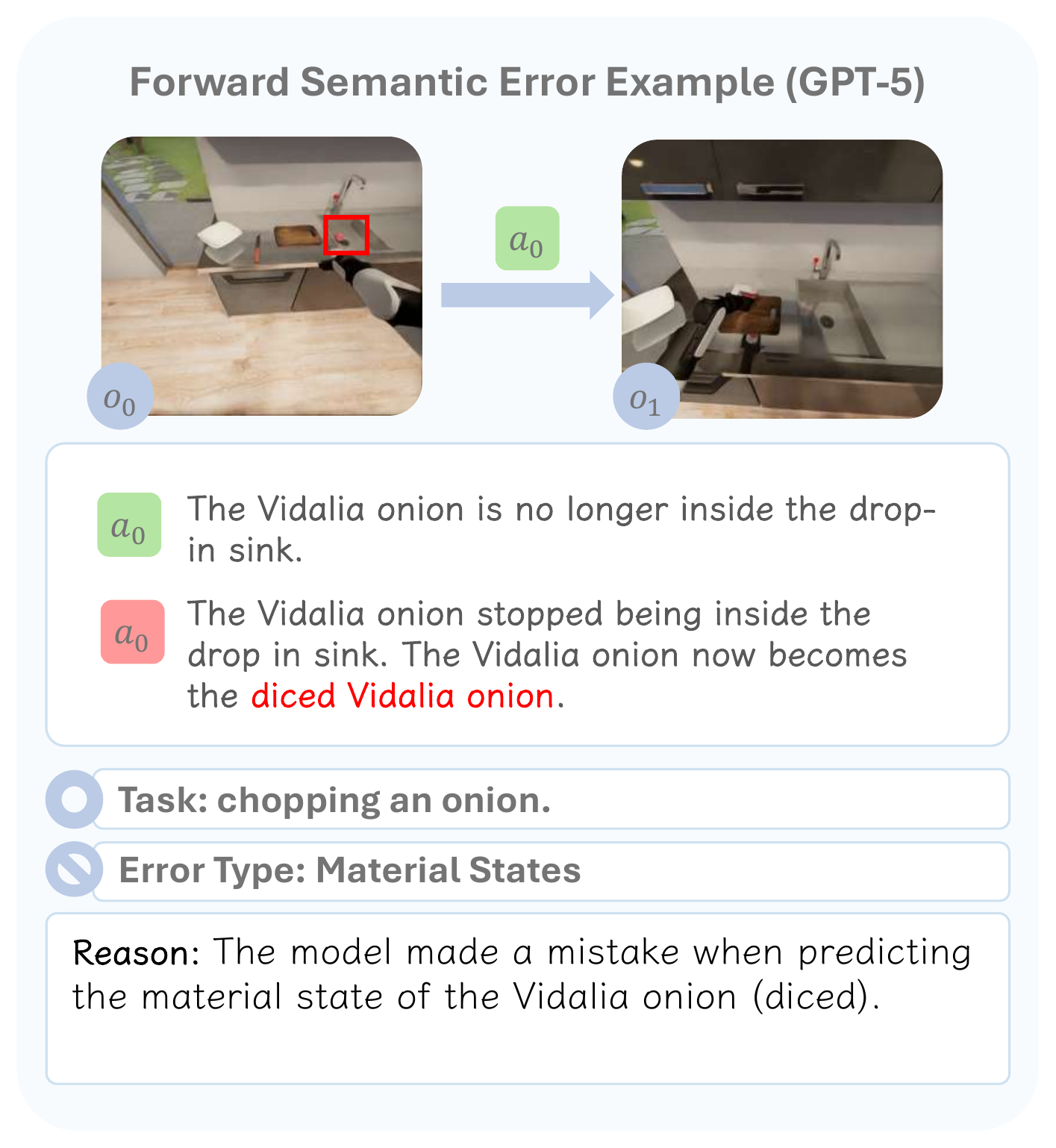}
    \label{fig:f_sem_MS}
  \end{subfigure}\hfill
  \begin{subfigure}[t]{0.5\linewidth}
    \centering
    \includegraphics[width=\linewidth]{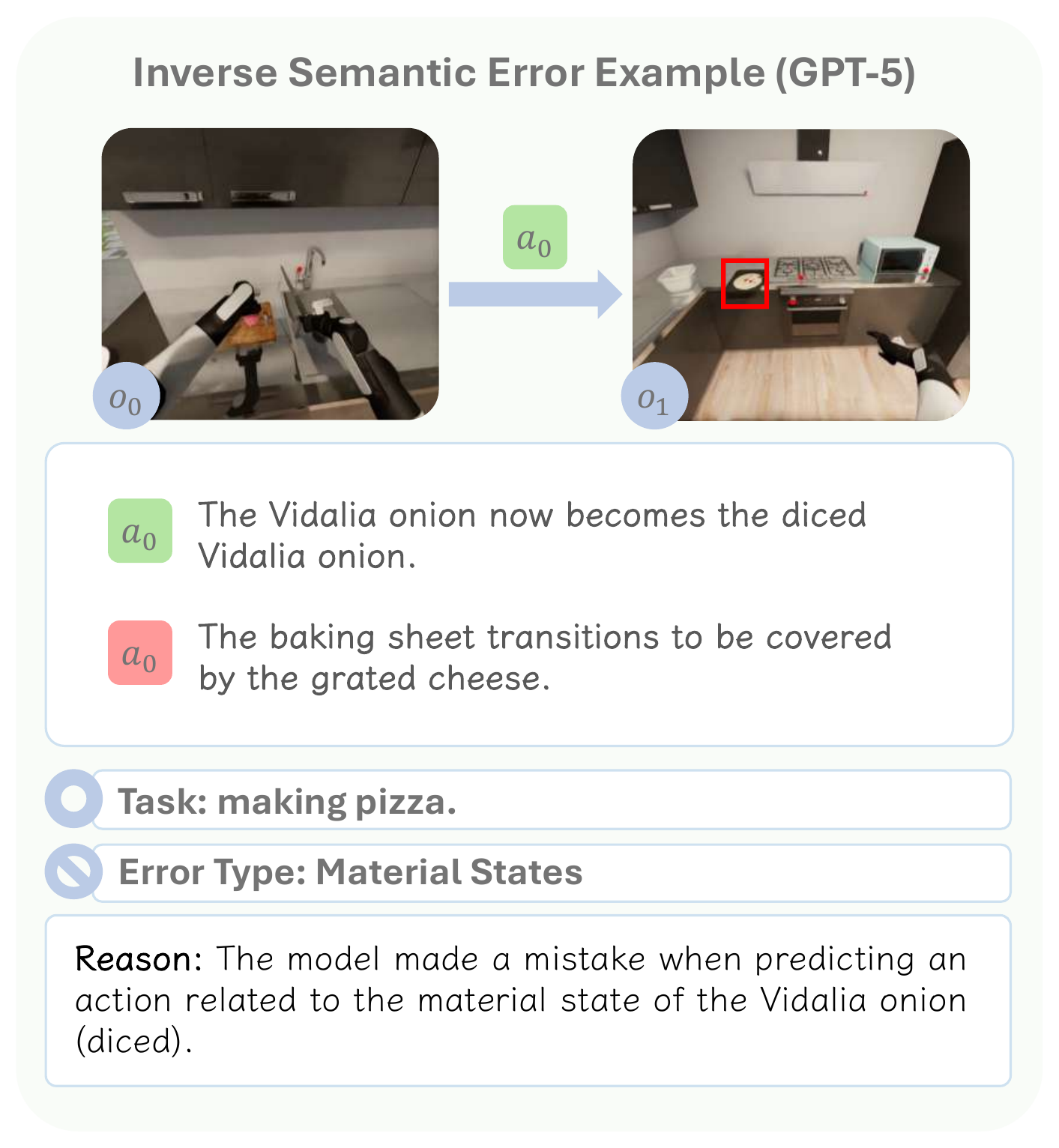}
    \label{i_sem_MS}
  \end{subfigure}
  
  \caption{Example of semantic error \textbf{Material States} by GPT-5 under forward and inverse tasks.}
  \label{fig:fi_sem_MS}
\end{figure}

\begin{figure}[t]
  \centering
  
  \begin{subfigure}[t]{0.5\linewidth}
    \centering
    \includegraphics[width=\linewidth,page=1]{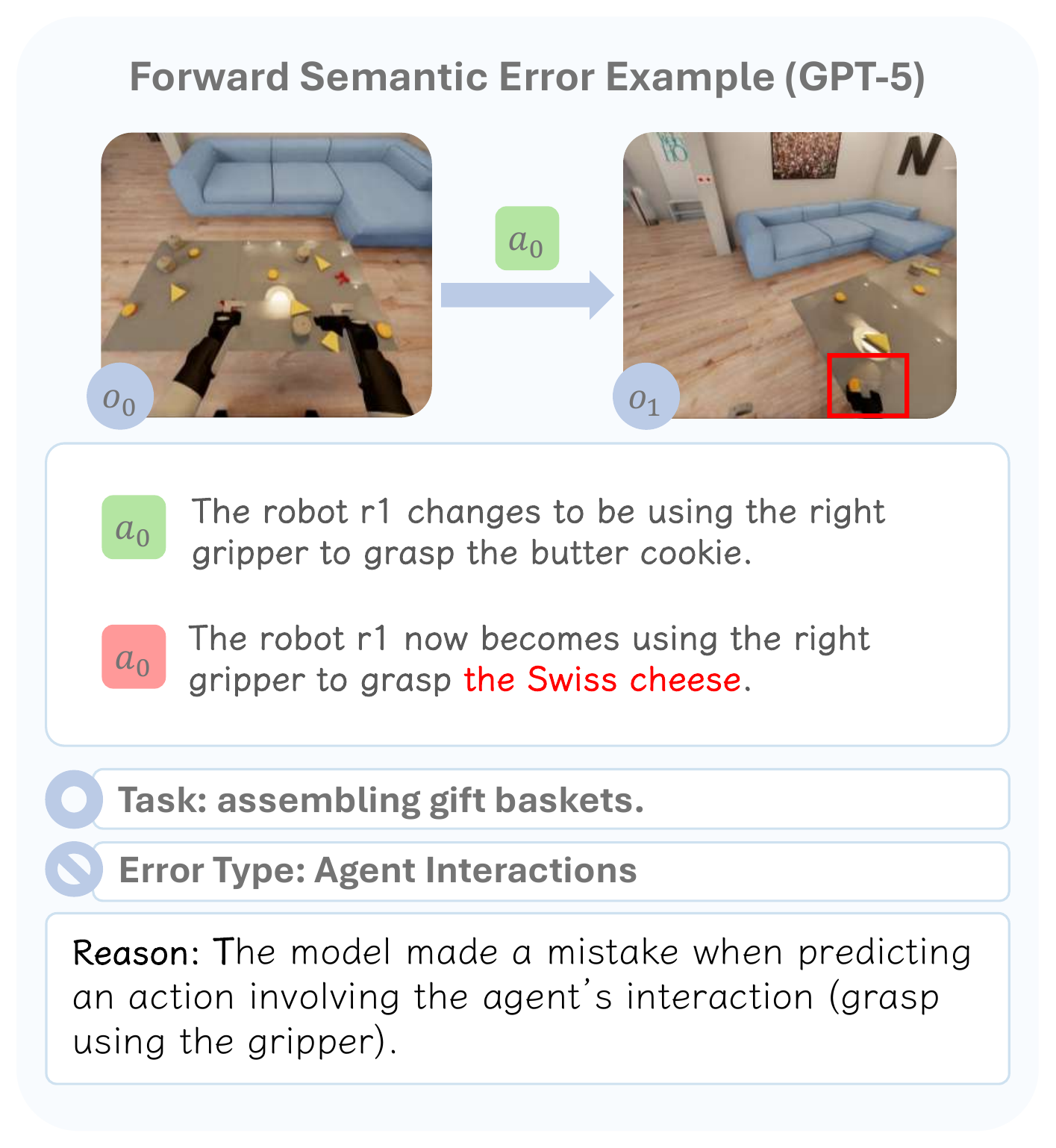}
    \label{f_sem_AI}
  \end{subfigure}\hfill
  \begin{subfigure}[t]{0.5\linewidth}
    \centering
    \includegraphics[width=\linewidth]{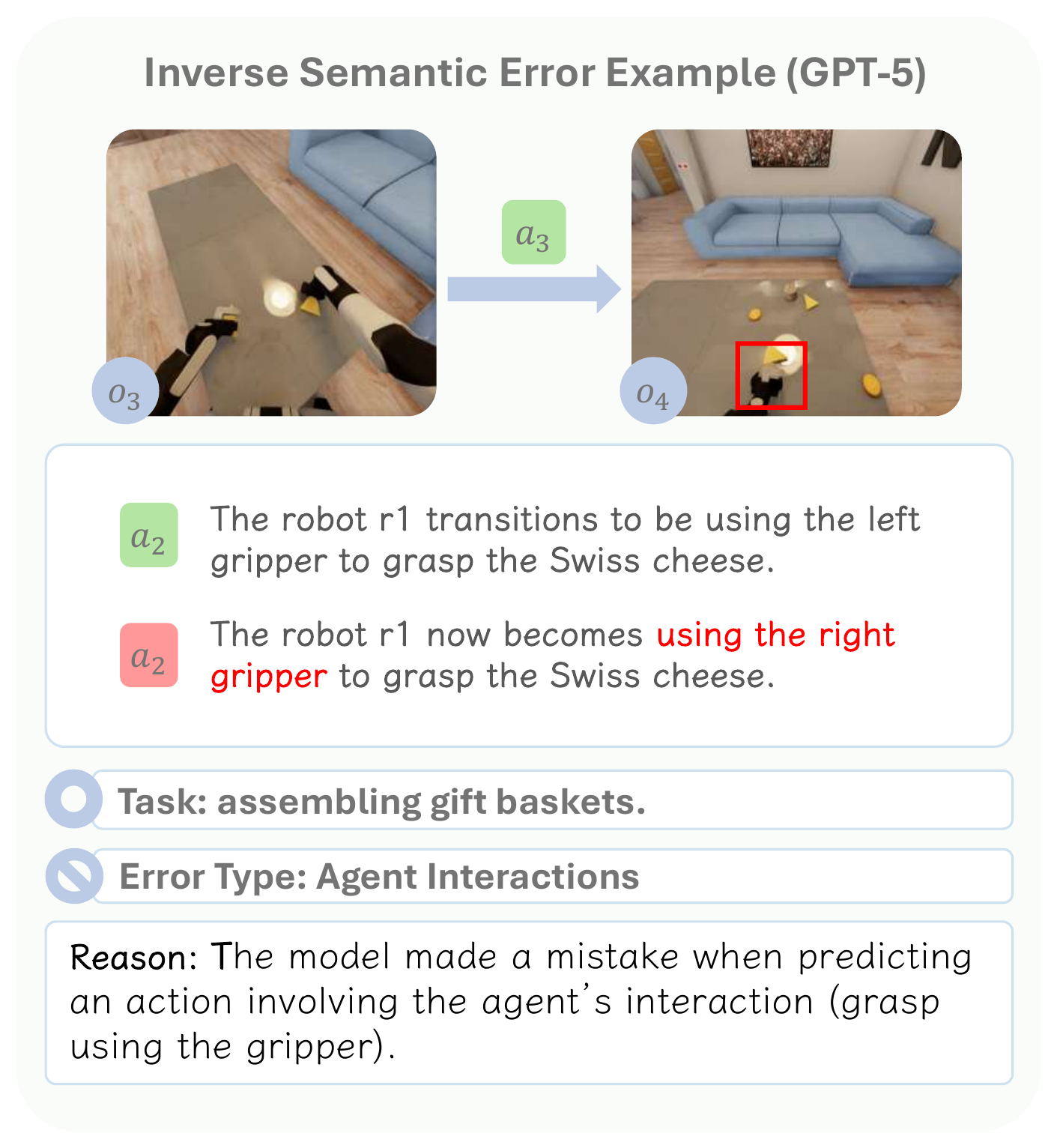}
    \label{i_sem_AI}
  \end{subfigure}
  
  \caption{Example of semantic error \textbf{Agent Interactions} by GPT-5 under forward and inverse tasks.}
  \label{fig:fi_sem_AI}
\end{figure}

In our semantic error analysis (Figure~\ref{fig:semantic_errors_statistics}), all systems—Gemini-2.5 Pro, GPT-5, GPT-5 mini, InternVL-3.5-241B-A28B, and humans—show a similar pattern: errors are concentrated in Spatial Relations and Agent Interaction, reflecting difficulties in reasoning about object positions and agent actions (e.g., left/right-hand grasping). A task-dependent asymmetry also appears: spatial-relations errors are more common in forward tasks, while agent-interaction errors are higher in inverse tasks. For illustration, we sample representative GPT-5 cases for each semantic category under both settings (Figures~\ref{fig:fi_sem_SR}, \ref{fig:fi_sem_FS}, \ref{fig:fi_sem_MS}, \ref{fig:fi_sem_AI}).

\SetKw{KwDownTo}{down to}
\SetKw{KwBreak}{break}

\begin{algorithm}[t]
\caption{Action-level Parsing of Signatures Data}
\label{alg:error-calc-parsing}
\DontPrintSemicolon
\KwIn{{Dataset of signatures $\mathcal{D}_{sig}$, each with ground-truth signatures $a^{sig}_{gt}$ and predicted signatures $a^{sig}_{p}$}}
\KwOut{Data of signatures $\mathcal{D}_{sig}'$ with missing, matched and hallucinated components}

\textbf{Signatures filtering}: 
$\mathcal{D}_{sig}' \leftarrow \emptyset$\; 
\ForEach{$(a^{sig}_{gt}, a^{sig}_p) \in \mathcal{D}_{sig}$}{
  \If{$|a^{sig}_{gt}| = |a^{sig}_{p}|$}{
    add $(a^{sig}_{gt}, a^{sig}_p)$ to $\mathcal{D}_{sig}'$\;
  }
  \Else{
    discard $(a^{sig}_{gt}, a^{sig}_p)$\;
  }
}

\textbf{Action-pairwise Comparison}: 
\ForEach{$(a^{sig}_{gt}, a^{sig}_p) \in \mathcal{D}_{sig}'$}{
  \tcc{$c_{mi}:$ missing components, $c_{ma}:$ matched components, $c_h:$ hallucinated components}
  $c_{mi} \leftarrow \emptyset$, 
  $c_{ma} \leftarrow \emptyset$, 
  $c_h \leftarrow \emptyset$\;
    \ForEach{$(c_{gt}, c_p) \in (a^{sig}_{gt}, a^{sig}_p)$}{
      \If{$c_{gt} = c_p$}{
        add $c_{gt}$ to $c_{ma}$\;
      }
      \Else{
        add $c_{gt}$ to $c_{mi}$\;
        add $c_{p}$ to $c_{h}$
      }
    }
    add $(c_{mi}, c_{ma}, c_h)$ to $\mathcal{D}_{sig}'$, 
    discard $(a^{sig}_{gt}, a^{sig}_p)$
}

\Return $\mathcal{D}_{sig}'$
\end{algorithm}

\SetKw{KwDownTo}{down to}
\SetKw{KwBreak}{break}

\begin{algorithm}[t]
\caption{Action-level Structural and Semantic Error Categorization}
\label{alg:error-calc-categorization}
\DontPrintSemicolon
\KwIn{Parsed signatures dataset $\mathcal{D}_{sig}'$, predicates $preds$}
\KwOut{Categorized errors dataset $\mathcal{D}_{err}$}

\textbf{Structural errors categorization}: 
$PI \leftarrow \emptyset$,\quad
$PS \leftarrow \emptyset$,\quad
$ES \leftarrow \emptyset$,\quad
$OM \leftarrow \emptyset$,\quad
$HA \leftarrow \emptyset$\;

\ForEach{$a^{sig} \in \mathcal{D}_{sig}'$}{
  $(C_{mi}, C_h) \leftarrow (c_{mi}(a^{sig}),\ c_{h}(a^{sig}))$\;  

  $(c_{mi}, c_h) \leftarrow$ \texttt{FindPairwiseErrors}
  $(C_{mi},C_h, \mathrm{polarity \ inversion})$;\\
  \If{$(c_{mi}, c_h) \neq \emptyset$}
    {add $(c_{mi}, c_h)$ to $PI$ \\
    remove $(c_{mi}, c_h)$ from $a^{sig}$}

  $(c_{mi}, c_h) \leftarrow$ \texttt{FindPairwiseErrors}
  $(C_{mi},C_h, \mathrm{predicate \ substitution})$;\\
  \If{$(c_{mi}, c_h) \neq \emptyset$}
    {add $(c_{mi}, c_h)$ to $PS$ \\
    remove $(c_{mi}, c_h)$ from $a^{sig}$ }

  $(c_{mi}, c_h) \leftarrow$ \texttt{FindPairwiseErrors}
  $(C_{mi},C_h, \mathrm{entity \ substitution})$;\\
  \If{$(c_{mi}, c_h) \neq \emptyset$}
    {add $(c_{mi}, c_h)$ to $ES$ \\
    remove $(c_{mi}, c_h)$ from $a^{sig}$ }
    
  \ForEach{$c_{mi}\in C_{mi}$}{ add $mi$ to $OM$ }

  \ForEach{$c_h\in C_h$}{ add $h$ to $HA$ }
}
\tcc{PI: Polarity Inversion, PS: Predicate Substitution, ES: Entity Substitution, OM: Omission, HA: Hallucination}
$\mathcal{D}_{err} \leftarrow (PI, PS, ES, OM, HA)$

\textbf{Semantic errors labeling}: 
\ForEach{$c$ in $\mathcal{D}_{err}$}
    {\ForEach{$pred$ in $preds$}
      {\If{$pred \in c$} {label $c$ with \texttt{SemanticError}($pred$)}}}

\Return $\mathcal{D}_{err}$
\end{algorithm}

\begin{algorithm}[t]
\caption{Dataset-Level Detection of Left--Right Hand Confusion}
\label{alg:hand-confusion-dataset}
\DontPrintSemicolon
\KwIn{Dataset of signature-level differences $\mathcal{D}_{diff} = \{(c_{mi}, c_{ma}, c_{h})\}$}
\KwOut{Confusion dataset $\mathcal{D}_{hand} = \{(\mathcal{D}_{l2r}, \mathcal{D}_{r2l})\}$}

\textbf{Left to right hand confusion}: 
 $\mathcal{D}_{l2r} \leftarrow \emptyset$\; 
\ForEach{$(c_{mi}, c_{ma}, c_{h}) \in \mathcal{D}_{diff}$}{
  \If{$\exists m \in c_{mi}$ that involves \text{left hand}}{
    \If{$\exists h \in c_{h}$ that involves \text{left hand}}{
      \textbf{continue},
    }
    \ElseIf{$\exists h \in c_{h}$ that involves \text{right hand}}{
      \ForEach{$m \in c_{mi}$}{
        \If{$m$ involves \text{left hand}}{ add $m$ to $\mathcal{D}_{l2r}$ }
      }
    }
  }
}

\textbf{Right to left hand confusion}: 
$\mathcal{D}_{r2l} \leftarrow \emptyset$\;
\ForEach{$(c_{mi}, c_{ma}, c_{h}) \in \mathcal{D}_{diff}$}{
  \If{$\exists m \in c_{mi}$ that involves \text{right hand}}{
    \If{$\exists h \in c_{h}$ that involves \text{right hand}}{
      \textbf{continue},
    }
    \ElseIf{$\exists h \in c_{h}$ that involves \text{left hand}}{
      \ForEach{$m \in c_{mi}$}{
        \If{$m$ involves \text{right hand}}{ add $m$ to $\mathcal{D}_{r2l}$ }
      }
    }
  }
}

$\mathcal{D}_{hand} \leftarrow (\mathcal{D}_{l2r}, \mathcal{D}_{r2l})$\;

\Return $\mathcal{D}_{hand}$
\end{algorithm}

\clearpage
\section{The Use of Large Language Models}
\label{06_llm_usage_justification}
We used large language models (LLMs), including Google's Gemini~2.5~Pro and OpenAI's GPT-5, as auxiliary tools to assist with writing, editing, and conducting the literature review for this manuscript. All content was critically revised and fact-checked by the human authors to ensure its scientific validity and originality. The authors are fully responsible for all statements and conclusions presented in this paper. Specifically, we use LLMs for polishing our wording and writing, and we use LLMs to retrieve several related works.

\end{document}